\pdfoutput=1

\documentclass[11pt]{article}

\usepackage{acl}

\usepackage{times}
\usepackage{latexsym}

\usepackage[T1]{fontenc}

\usepackage[utf8]{inputenc}

\usepackage{hyperref}       
\usepackage{url}            
\usepackage{booktabs}       
\usepackage{nicefrac}       
\usepackage{microtype}      
\usepackage{xcolor}         
\usepackage{graphicx}
\usepackage{amsmath}
\usepackage{amssymb}
\usepackage{mathtools}
\usepackage{amsthm}
\usepackage{array}
\usepackage{longtable}
\usepackage[most]{tcolorbox}
\usepackage{bibentry}
\usepackage{bm} 
\usepackage{tabularx}
\usepackage{multirow} 
\usepackage{algorithm}
\usepackage{algorithmic}
\usepackage{arydshln}
\usepackage{appendix}
\makeatletter
\def\blankfootnote{\xdef\@thefnmark{}\@footnotetext}
\makeatother

\definecolor{royalblue(web)}{rgb}{0.25, 0.41, 0.88}
\definecolor{blue-violet}{rgb}{0.54, 0.17, 0.89}
\definecolor{brightmaroon}{rgb}{0.76, 0.13, 0.28}
\definecolor{darkmagenta}{rgb}{0.55, 0.0, 0.55}
\definecolor{bleudefrance}{rgb}{0.19, 0.55, 0.91}
\definecolor{palatinateblue}{rgb}{0.15, 0.23, 0.89}
\definecolor{royalblue(web)}{rgb}{0.25, 0.41, 0.88}
\definecolor{whitesmoke}{rgb}{0.96, 0.96, 0.96}
\definecolor{thulianpink}{rgb}{0.87, 0.44, 0.63}
\definecolor{amber(sae/ece)}{rgb}{1.0, 0.49, 0.0}
\definecolor{darkblue}{rgb}{0.0, 0.0, 0.55}
\definecolor{alizarin}{rgb}{0.82, 0.1, 0.26}
\definecolor{asparagus}{rgb}{0.53, 0.66, 0.42}
\definecolor{darkspringgreen}{rgb}{0.09, 0.45, 0.27}
\definecolor{columbiablue}{rgb}{0.61, 0.87, 1.0}
\definecolor{wildblueyonder}{rgb}{0.64, 0.68, 0.82}
\definecolor{trolleygrey}{rgb}{0.5, 0.5, 0.5}
\definecolor{paleaqua}{rgb}{0.74, 0.83, 0.9}
\definecolor{bubblegum}{rgb}{0.99, 0.76, 0.8}
\definecolor{coralred}{rgb}{1.0, 0.25, 0.25}
\definecolor{green(ryb)}{rgb}{0.4, 0.69, 0.2}
\definecolor{flame}{rgb}{0.89, 0.35, 0.13}
\definecolor{bittersweet}{rgb}{1.0, 0.44, 0.37}
\definecolor{darksalmon}{rgb}{0.91, 0.59, 0.48}
\definecolor{emerald}{rgb}{0.31, 0.78, 0.47}
\definecolor{green(pigment)}{rgb}{0.0, 0.65, 0.31}
\definecolor{codegreen}{rgb}{0,0.6,0}
\definecolor{codegray}{rgb}{0.5,0.5,0.5}
\definecolor{codepurple}{rgb}{0.58,0,0.82}
\definecolor{backcolour}{rgb}{0.96,0.96,0.94}
\definecolor{bluegray}{rgb}{0.3, 0.38, 0.47}
\definecolor{whitesmoke}{rgb}{0.96, 0.96, 0.96}
\definecolor{codegreen}{rgb}{0,0.6,0}
\definecolor{codegray}{rgb}{0.5,0.5,0.5}
\definecolor{codepurple}{rgb}{0.58,0,0.82}
\definecolor{backcolour}{rgb}{0.96,0.96,0.94}

\tcbuselibrary{breakable}
\usepackage{listings}
\lstdefinestyle{mystyle}{
  basicstyle=\scriptsize\ttfamily,
  frame=single, 
  columns=fixed, 
}

\newtheorem{theorem}{Theorem}

\newtheorem{proposition}{Proposition}
\newtheorem{definition}{Definition}

\newcommand{\ours}{{\fontfamily{qpl}\selectfont Flexora}}
\newcommand{\norm}[1]{\left\|#1\right\|}

\usepackage{amsmath,amsfonts,bm}

\def\1{\bm{1}}

\DeclareMathAlphabet{\mathsfit}{\encodingdefault}{\sfdefault}{m}{sl}
\SetMathAlphabet{\mathsfit}{bold}{\encodingdefault}{\sfdefault}{bx}{n}

\def\gD{{\mathcal{D}}}

\def\gR{{\mathcal{R}}}

\def\sR{{\mathbb{R}}}

\newcommand{\E}{\mathbb{E}}

\DeclareMathOperator*{\argmin}{arg\,min}

\title{\ours{}: Flexible Low-Rank Adaptation for Large Language Models}

\author{
  Chenxing Wei$^{* \dagger \S}$, Yao Shu$^{* \wr}$, Ying Tiffany He$^{\dagger}$, Fei Yu$^{\# \dagger \S \ddagger}$\\
$^\dagger$College of Computer Science and Software Engineering, Shenzhen University, China \\
$^{\S}$Guangdong Laboratory of Artificial Intelligence and Digital Economy (SZ), China \\
$^{\wr}$Hong Kong University of Science and Technology (Guangzhou), China \\
$^{\ddagger}$School of Information Technology, Carleton University, Canada \\
\texttt{weichenxing2023@email.szu.edu.cn}, \texttt{yaoshu@hkust-gz.edu.cn}\\ \texttt{heying@szu.edu.cn}, \texttt{yufei@gml.ac.cn}
}  

\begin{document}
\maketitle
\begin{abstract}
Large language models (LLMs) have revolutionized artificial intelligence, but their performance on specific tasks is often limited by knowledge boundaries. While fine-tuning techniques like low-rank adaptation (LoRA) aim to address this, they can suffer from overfitting. We propose \textit{\underline{flex}ible l\underline{o}w-\underline{r}ank \underline{a}daptation} (\ours{}), a novel method that automatically selects the most critical layers for fine-tuning to optimize performance across diverse downstream tasks. \ours{} formulates layer selection as a hyperparameter optimization problem, employs unrolled differentiation for efficient solving, and identifies the most impactful layers based on optimized hyperparameters. Extensive experiments across various pre-trained models and natural language tasks demonstrate that \ours{} consistently outperforms existing baselines. We provide theoretical insights and comprehensive ablation studies to elucidate the effectiveness of \ours{}. Therefore, \ours{} offers a robust solution to enhance LoRA fine-tuning for LLMs, potentially advancing the field of adaptive language model optimization.
\end{abstract}
\blankfootnote{$*$ Equal contribution, ${\#}$ corresponding author.}
\section{Introduction}

The advent of large language models (LLMs) \citep{c:23,c:24} has revolutionized artificial intelligence, offering unprecedented capabilities across various domains. However, this progress comes at a significant cost: LLMs demand substantial computational resources due to their vast parameter sets and complex functionalities \citep{c:25,c:29,wei2025paftpromptagnosticfinetuning}. This challenge has spurred the development of parameter-efficient fine-tuning (PEFT) methods \citep{c:38,c:39}, with low-rank adaptation (LoRA) \citep{c:36} emerging as a particularly promising approach. The innovation of LoRA lies in its ability to freeze pre-trained parameters while introducing trainable auxiliary parameters ($\Delta \bm{W}$) at each layer, dramatically reducing training costs while maintaining impressive performance. However, despite its widespread adoption, LoRA is not without limitations. It can underperform on certain tasks, likely due to overfitting issues, as evidenced in benchmarks like GLUE \citep{c:60}, summary tasks \citep{c:115}, and complex reasoning tasks \citep{c:66}. Existing techniques to combat overfitting, such as dropout \citep{c:110} and novel regularization strategies \citep{c:83}, often yield performance comparable to or lower than vanilla LoRA and lack the flexibility to adapt across different tasks. Moreover, current methods typically require manual hyperparameter tuning, limiting their practical applicability in diverse scenarios. These challenges therefore underscore the urgent need for an algorithm that delivers superior performance, enables automatic hyperparameter tuning, and supports flexible training across various tasks.

\begin{figure*}[t]
\centering
\includegraphics[width=1.0\textwidth]{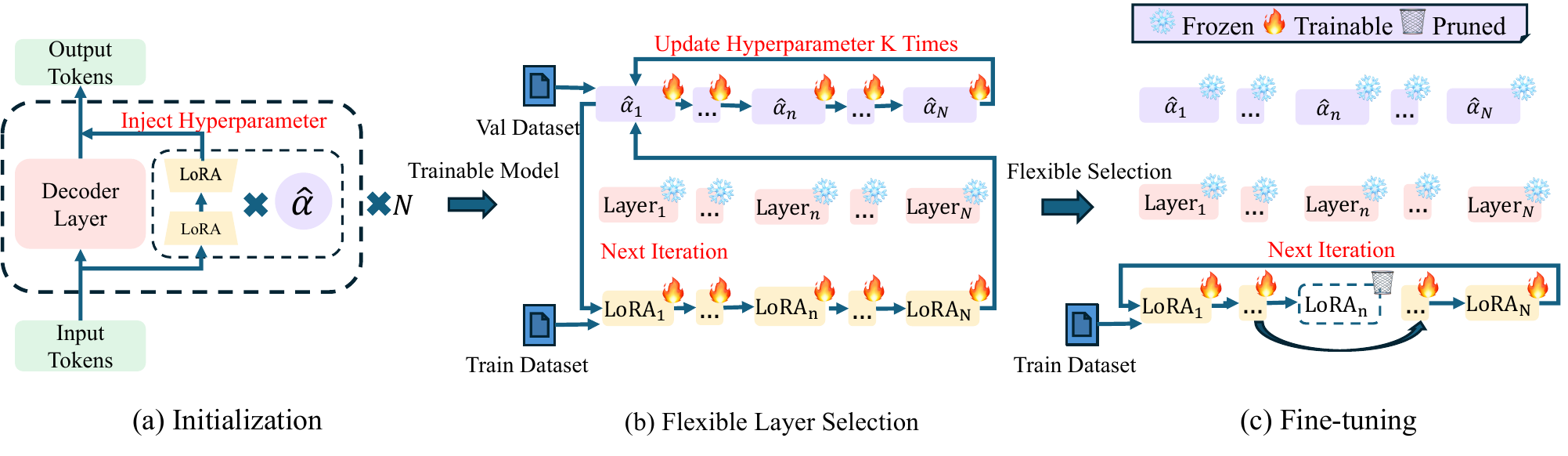}
\vspace{-5mm}
\caption{An overview of \ours{}: (a) Initialization of hyperparameters $\widehat{\alpha}$ and their integration with LoRA parameters to produce the Trainable Model. (b) Simultaneous training of LoRA parameters and hyperparameters $\widehat{\alpha}$ using different datasets, minimizing empirical risk for both validation and training datasets. The hyperparameter vector $\widehat{\alpha}$ is then ranked based on magnitude. (c) Flexible selection of layers to be trained, where higher-ranked layers are activated for training while others remain frozen.}
\label{fig1}
\vspace{-3mm}
\end{figure*}

To address these limitations, we introduce \textit{\underline{flex}ible l\underline{o}w-\underline{r}ank \underline{a}daptation} (\ours{}), a novel framework designed to flexibly fine-tune LLMs using an automated layer-level policy. Our approach is inspired by hyperparameter optimization (HPO) and offers several key innovations. First, we demonstrate that fine-tuning only the most critical layers can significantly reduce overfitting and enhance performance. Second, we frame the layer selection problem as an HPO task and employ unrolled differentiation (UD) to solve it efficiently. Third, we develop a three-stage process that automatically identifies and focuses on the most important layers for downstream tasks. As illustrated in Figure~\ref{fig1}, \ours{} operates through an initialization stage (Sec.~\ref{sec:initial}) that injects defined hyperparameters into LoRA parameters, a flexible layer selection stage (Sec.~\ref{sec:select}) that optimizes these hyperparameters using UD, and a fine-tuning stage (Sec.~\ref{sec:fine-tune}) that selectively updates only the most crucial layers, significantly reducing computational overhead. Our extensive empirical results (Sec.~\ref{sec:results}) demonstrate that \ours{} effectively reduces unimportant LoRA parameters, mitigates overfitting, and enhances overall performance across a variety of tasks and model architectures.

In summary, our key contributions consist of: (a) the introduction of \ours{}, a novel framework for automatic layer selection in LoRA fine-tuning; (b) a formulation of layer selection as an HPO task, efficiently solved using unrolled differentiation; (c) comprehensive validation through extensive experiments on various LLMs and downstream tasks; and (d) theoretical insights into the performance improvements achieved by \ours{}, providing a deeper understanding of its effectiveness.

\section{Related Work}
\label{related work}
\paragraph{Low-Rank Adaptation (LoRA)}
Low-Rank Adaptation (LoRA) methods are widely used to reduce training parameters when fine-tuning large language models (LLMs) for specific applications. However, LoRA often suffers from overfitting, which can degrade performance on downstream tasks. To mitigate this, various strategies have been proposed: LoRA-SP \citep{c:60} randomly freezes half of the LoRA parameters during fine-tuning to alleviate overfitting; LoRA-FA \citep{c:61} freezes down-projection weights while updating only up-projection weights; VeRA \citep{kopiczko2024vera} introduces vector-based random matrix adaptation, significantly reducing trainable parameters compared to LoRA; LoRA-drop \citep{c:65} prunes less important parameters based on layer output analysis; AdaLoRA \citep{c:41} dynamically allocates the parameter budget across weight matrices based on importance scores; LoRAPrune \citep{c:66} jointly prunes parts of the LoRA matrix and LLM parameters based on gradients; and LoRAShear \citep{c:71} employs knowledge-based structured pruning to reduce costs while enhancing generalization. Despite their benefits, these methods often \textit{(a)} require significant design effort, \textit{(b)} struggle to adapt across different tasks, and \textit{(c)} can be overly complex for practical application. In contrast, we introduce \ours{}, a framework designed for flexible LoRA fine-tuning across various tasks using a simple, automated layer-level policy.

\paragraph{Hyperparameter Optimization (HPO)}
HPO is widely applied across various domains. Specifically, in the domain of neural architecture search, DARTS \citep{c:81} conceptualizes the coefficients defining the network architecture as hyperparameters. In the domains of feature learning, $\rm DS^3L$ \citep{c:49} considers feature extractors as hyperparameters. In the field of data science, TPOT \citep{c:100} employs hyperparameters as weights to measure the importance of data. By minimizing the validation loss over these hyperparameters, the optimal variables, e.g., the architectures in \citet{c:81}, the features in \citet{c:49}, and the data in \citet{c:100}, are identified, leading to superior performance in their respective domains. Drawing inspiration from these works, we initially formulated the layer selection in the LoRA method as an HPO problem. This involves optimizing hyperparameters to quantify the contributions of different layers, aiming to achieve optimal performance on downstream tasks and thereby select the most crucial layers for fine-tuning. This formulation subsequently led to the development of our \ours{}.

\begin{figure*}[t]
\centering
\includegraphics[width=1.0\textwidth]{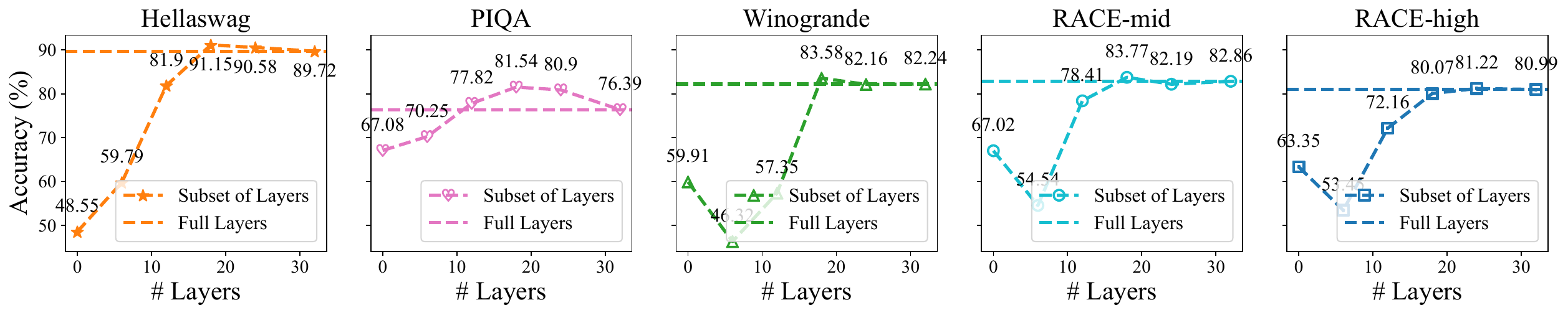} 
\vspace{-5mm}
\caption{This figure depicts the relationship between the number of LoRA fine-tuning layers and model accuracy across four distinct datasets: Hellaswag, PIQA, Winogrande, and RACE, with the latter including two separate tasks, RACE-mid and RACE-high, which vary in difficulty. Results for LoRA rank 8 are shown here.
The $x$-axis represents the number of fine-tuned layers, ranging from 0 to 32, where 0 corresponds to the base model without fine-tuning. Selected configurations include 6, 12, 18, 24, and 32 randomly fine-tuned layers. The full 32-layer configuration, representing the vanilla LoRA setup, is shown as a horizontal dashed line in the plots. The $y$-axis indicates model accuracy as a percentage.}
\label{fig2}
\vspace{-3mm}
\end{figure*}

\section{Preliminaries}
In this section, we first provide empirical insights showing that layer selection is crucial for improving the performance of LLMs in Sec.~\ref{sec:emp-insights}, and then frame the layer selection problem as a well-defined HPO problem in Sec.~\ref{sec:problem formulation}. 

\subsection{Empirical Insights}\label{sec:emp-insights}
To study the impact of the number of LoRA fine-tuning layers on overall performance, we conducted a preliminary study using Llama3-8B \cite{c:80} across a range of downstream tasks. Here, we randomly selected different subsets of layers, different ranks (e.g., 4, 8, 16, 32) for LoRA fine-tuning, and evaluated their performance on these tasks.  The findings, shown in Figure~\ref{fig2} and Appendix~\ref{sec:random layer rank}, reveal a clear trend: while increasing the number of fine-tuned layers generally improves model performance, there is a critical point beyond which fine-tuning more layers leads to potential overfitting and subsequent performance decline. This hence suggests that selecting an optimal subset of layers for LoRA fine-tuning is crucial for maximizing performance, which interestingly aligns with the previous empirical studies\cite{c:111, c:65, c:71, wei2025redit}.

\subsection{Problem Formulation}\label{sec:problem formulation}
Inspired by the empirical insights above, we aim to identify the most critical layers in LoRA fine-tuning to improve generalization performance across a variety of downstream tasks. Formally, we consider an $N$-layer LLM with LoRA fine-tuning parameters $\theta \in \sR^d$, and let the hyperparameter $\alpha \in \{0,1\}^N$ denote the selection of fine-tuning layers, where a value of 1 indicates that a layer is selected for fine-tuning. Given the test data distribution $\gD_{\text{test}}$ and the training dataset $S_{\text{train}}$, we then define the expected test and training error as $\gR^{\text{test}}(\theta, \alpha) \triangleq \E_{x \sim \gD_{\text{test}}} \left[\ell(x, \theta; \alpha)\right]$ and $\gR^{\text{train}}(\theta, \alpha) \triangleq \E_{x \sim S_{\text{train}}} \left[\ell(x, \theta; \alpha)\right]$, respectively.

Hence, to select the optimal LoRA fine-tuning layers for maximized performance on downstream tasks, we aim to solve the following bilevel optimization problem:
\begin{equation}\label{eq:csee}
\begin{aligned}
    &\min_{\alpha \in \{0,1\}^N} \gR^{\text{test}}(\theta^*(\alpha), \alpha) \\
    &\text{s.t.} \quad \theta^*(\alpha) = \argmin_{\theta \in \mathbb{R}^d} \gR^{\text{train}}(\theta, \alpha) \ .
\end{aligned}
\end{equation}
This formulation follows a standard hyperparameter optimization (HPO) approach as demonstrated in \citet{c:69}, where $\alpha$ serves as the hyperparameter. Thus, the layer selection problem for LoRA fine-tuning in LLMs is framed as a well-defined HPO problem.

Unfortunately, it is typically infeasible to access the full test distributions, denoted by $\gD_{\text{test}}$, for this optimization. This expected test error can typically be approximated by the empirical validation error based on the validation dataset $S_{\text{val}}$, which is defined as 
$\widehat{\gR}^{\text{val}}(\theta, \alpha) \triangleq \E_{x \sim S_{\text{val}}} \left[\ell(x, \theta; \alpha)\right]$
Therefore, \eqref{eq:csee} can be simplified as:
\begin{equation}\label{equ: main loss}
\begin{aligned}
    &\min_{\alpha \in \{0,1\}^N}\widehat{\gR}^{\text{val}}(\theta^*(\alpha), \alpha) \\
    &\text{s.t.} \quad \theta^*(\alpha) = \argmin_{\theta \in \mathbb{R}^d} \gR^{\text{train}}(\theta, \alpha) \ .
\end{aligned}
\end{equation}

\section{The \ours{} Framework}
To address the layer selection problem defined above, we propose our \textit{flexible low-rank adaptation for LLMs} (\ours{}) framework in Figure~\ref{fig1}. As illustrated in Figure~\ref{fig1}, the \ours{} framework consists of three key stages: an initial stage (detailed in Sec.~\ref{sec:initial}), a flexible layer selection stage (detailed in Sec.~\ref{sec:select}), and a fine-tuning stage for the selected LoRA layers (detailed in Sec.~\ref{sec:fine-tune}).

\subsection{Initial Stage}\label{sec:initial}
We begin by introducing a special formulation of LoRA, which incorporates the layer selection hyperparameter $\alpha = (\alpha^{(1)}, \cdots, \alpha^{(N)}) \in \{0, 1\}^N$, as follows:
\begin{equation}\label{eq:csv}
h^{(i)} = W z^{(i)} + \alpha^{(i)} B^{(i)} A^{(i)} z^{(i)}, \quad \text{s.t.}\quad \alpha_i \in \{0, 1\} \ .
\end{equation}
Here, $h^{(i)}$ is the output of the $i$-th layer, where $W$ is the original weight matrix, $z^{(i)}$ is the input, and $B^{(i)}$ and $A^{(i)}$ are the low-rank adaptation matrices of LoRA. The hyperparameter $\alpha^{(i)}$ determines whether LoRA is applied for layer $i$. Specifically, if $\alpha^{(i)} = 0$, the equation simplifies to $h^{(i)} = W z^{(i)}$, meaning the $i$-th layer reverts to standard computation without LoRA, implying that the additional complexity of LoRA is unnecessary for layer $i$. Conversely, when $\alpha^{(i)} = 1$, the equation becomes the standard LoRA form, $h^{(i)} = W z^{(i)} + B^{(i)} A^{(i)} z^{(i)}$, indicating that LoRA significantly enhances the performance of layer $i$ by allowing the low-rank matrices to better capture complex patterns. So, this dynamic adjustment allows the model to selectively apply LoRA when a specific layer is most beneficial, thereby optimizing the fine-tuning process and mitigating overfitting. 

However, due to the inherent difficulty of directly optimizing the discrete layer selection hyperparameter $\alpha$, we adopt a continuous relaxation approach by replacing the $\alpha$ in \eqref{eq:csv} with its continuous counterpart, $\widehat{\alpha} = (\widehat{\alpha}^{(1)}, \cdots, \widehat{\alpha}^{(N)})$: 
\begin{equation}\label{eq:cshf}
\begin{aligned}
    &h^{(i)} = W_0 z^{(i)} + \widehat{\alpha}^{(i)} B^{(i)} A^{(i)} z^{(i)}, \\
    &\text{s.t.}\quad \widehat{\alpha}^{(i)} = \frac{\exp{(\alpha^{(i)})}}{\sum_{i \in [N]} \exp{(\alpha^{(i)})}} N \ .
\end{aligned}
\end{equation}
Notably, $\alpha \in \sR^N$ now and $\alpha$ are typically initialized to zeros, providing a neutral starting point where no layer is initially excluded from LoRA fine-tuning. Meanwhile, the constant scale $N$ ensures that when all layers are selected for fine-tuning, the scale of each selected layer for LoRA fine-tuning is preserved, resulting in $\widehat{\alpha}^{(i)} = 1$ for all layers, aligning with the vanilla LoRA scale as shown above.

\subsection{Flexible Layer Selection Stage}\label{sec:select}
\paragraph{Optimization Strategy.}

Given the continuous relaxation $\widehat{\alpha}$ defined above, we propose to solve the well-defined HPO problem in Equation~\ref{equ: main loss} using the widely applied unrolled differentiation (UD) method \citep{c:116,c:117,c:118,c:119,c:120}. The UD method typically involves two alternating optimization processes: (a) the inner-level and (b) the outer-level optimization. In this paper, the outer-level optimization is defined as $\argmin_{\theta \in \mathbb{R}^d} \gR^{\text{train}}(\theta, \alpha)$, in which the layer selection hyperparameter $\alpha$ is fixed, and the LoRA parameters \(\theta\) are updated using stochastic gradient methods (e.g., SGD~\citep{c:121}) on the training dataset $S_{\text{train}}$. This step focuses on optimizing model performance by adjusting the parameters associated with the selected layers (line 3-4 in Algorithm~\ref{algo:ud}). Meanwhile, the inner-level optimization is $\argmin_{\alpha \in \sR^N}\widehat{\gR}^{\text{val}}(\theta, \alpha)$, in which the layer selection hyperparameter \(\alpha\) is updated using stochastic gradient methods (e.g., SGD) based on the validation performance of the optimized LoRA parameters $\theta$ from the inner-level process (lines 6–9 in Algorithm~\ref{algo:ud}). This step intends to maximize the validation performance of LoRA fine-tuning based on a subset of selected layers. These two alternating processes therefore iteratively refine both the model parameters and the layer selection criteria, making LoRA layer selection more computationally efficient in practice. After $T$ iterations of these alternating processes, the converged $\alpha_T$ is output as the optimal layer selection denoted as $\alpha^*$ (line 12 in Algorithm~\ref{algo:ud}). 

\begin{algorithm}[t]
\caption{The \ours{} Framework}
\label{algo:ud}
\small
\begin{algorithmic}[1] 
    \STATE {\bfseries Input:} Number of steps $T$ and $K$; Initialized LoRA parameters $\theta_0$ and hyperparameter  $\alpha_0=0$; Learning rate $\eta_{\alpha}$ and $\eta_{\theta}$
    \FOR{$t=0$ {\bfseries to} $T-1$}
      \STATE Sample a mini-batch $B_{\text{train}} \sim S_{\text{train}}$
      \STATE $\theta_{t + 1} \gets \theta_{t} - \eta_{\theta} \nabla_{\theta} \left( \frac{1}{|B_{\text{train}}|} \sum_{x \in B_{\text{train}}} \ell(x, \theta; \alpha_t) \right) \big|_{\theta = \theta_{t}}$
      \STATE $\alpha_{t+1}^0 \gets \alpha_{t}$
      \FOR{$k=0$ {\bfseries to} $K-1$}
      \STATE Sample a mini-batch $B_{\text{val}} \sim S_{\text{val}}$
      \STATE $\alpha_{t+1}^{k+1} \gets \alpha_{t+1}^k - \eta_{\alpha} \nabla_{\alpha} \left( \frac{1}{|B_{\text{val}}|} \sum_{x \in B_{\text{val}}} \ell(x, \theta_{t+1}; \alpha) \right) \big|_{\alpha = \alpha_{t+1}^k}$
      \ENDFOR
    \STATE $\alpha_{t+1} \gets \alpha_{t+1}^K$
    \ENDFOR
    \STATE {\bf return} $\alpha^* = \alpha_{T}$
\end{algorithmic}
\end{algorithm}

\begin{table*}[t!]
    \centering
    \scriptsize
    \caption{Comparison of accuracy across various common sense reasoning tasks using Llama3-8B. The baseline experimental configuration is detailed in Appendix~\ref{app:setup}. Here, "Pre-trained" refers to using the base model for reasoning, "Full FT" indicates full parameter fine-tuning, and "Random (Greedy)" represents the best result from randomly selected layers. Unless otherwise specified, the results are based on the default LoRA Rank of 8.
}
\label{tab:accuracy_comparison}
\resizebox{\textwidth}{!}{
    \begin{tabular}{l*{6}{c}}
        \toprule
        \textbf{Methods} & \textbf{Hellaswag} & \textbf{PIQA} & \textbf{Winogrande} & \textbf{RACE-mid}& \textbf{RACE-high} & \textbf{Average} \\
        \midrule
        Pre-trained & 48.55 & 67.08 & 59.91 & 67.02 & 63.35 & 61.18 \\
        Full FT & 90.53 & 79.32 & 81.16 & 81.92 & 79.36 & 82.46 \\
        \midrule
        LoRA($r=8$) & 89.72 & 76.39 & 82.24 & 82.86 & 80.99 & 83.04 \\
        LoRA($r=16$) & 89.99 & 78.47 & 82.77 & 81.63 & 79.68 & 82.51\\
        LoRA($r=32$) & 90.01 & 79.56 & 84.36 & 82.36 & 80.99 & 83.46 \\
        LoRA-SP & 89.37 & 78.97 & 83.67 & 83.27 & 79.01 & 82.86 \\
        LoRA-FA & 89.16 & 75.97 & 82.16 & 82.79 & 79.03 & 81.83 \\
        VeRA & 90.98 & 78.63 & 83.64 & 83.55 & 78.84 & 83.13 \\
        LoRAPrune (Ratio = 0.5) & 88.42 & 77.12 & 81.23 & 82.96 & 80.42 & 82.03 \\
        AdaLoRA ($r_0=4$)& 90.17 & 80.20 & 77.19 & 83.15 & 77.93 & 81.73 \\
        LoRA-drop & 91.86 & 77.91 & 76.46 & 77.30 & 75.24 & 79.75 \\
        \midrule
        Random (Greedy) & 91.15 & 81.54 & 83.58 & 83.77 & 81.22 & 84.25 \\
        \textbf{\ours{}($r=8$)} & 93.62 & 85.91 & \textbf{85.79} & 84.61 & 82.36 & 86.46 \\
        \textbf{\ours{}($r=16$)} & 93.71 & 85.26 & 84.99 & \textbf{85.62} & \textbf{83.03} & \textbf{86.52}\\
        \textbf{\ours{}($r=32$)} & \textbf{93.87} & \textbf{86.02} & 85.01 & 84.27 & 81.97 & 86.23\\
        \bottomrule
    \end{tabular}
}
\vspace{-3mm}
\end{table*}

\paragraph{Selection Strategy.} To begin with, we introduce the following proposition:
\begin{proposition} \label{proposition1}
If $\alpha$ is initialized to zeros, then for any $T \geq 0$ and $K \geq 0$ in Alg.~\ref{algo:ud},
$\sum_{i=1}^{N} \alpha^{(i)} = 0$.
\end{proposition}
The proof of this proposition is provided in Appendix~\ref{proof_pro1}. This result highlights that the mean value of the hyperparameter \(\alpha\) remains 0, indicating that after the layer selection stage, the elements in the optimized hyperparameter \(\alpha^*\) can take on both positive and negative values. Therefore, we propose to determine the layers for LoRA fine-tuning by selecting layers with $\alpha^{(i)} > 0$ , as these layers are expected to make positive contributions. In contrast, layers with $\alpha^{(i)} \leq 0$ are believed to be less beneficial or even harmful to LoRA fine-tuning. As a result, this method not only facilitates automatic layer selection but also provides flexibility in adjusting the number and specific layers for LoRA fine-tuning, helping to mitigate the potential overfitting and improve overall performance.

\subsection{Fine-Tuning Stage}\label{sec:fine-tune}
During the fine-tuning stage, as illustrated in Figure~\ref{fig1}c, we adopt a selective activation strategy. In this phase, we freeze the layers not selected for fine-tuning, keeping their parameters unchanged, and focus on retraining only the selected layers to enhance performance. This targeted approach concentrates computational resources on the most critical layers for the downstream task. By retraining the LoRA parameters from scratch in these layers, the model adaptively learns optimal representations, reducing the risk of overfitting and improving performance, especially for simpler tasks. We will validate this approach with the empirical results presented below.

\section{Empirical Results}\label{sec:results}
In this section, we present comprehensive experiments to support the effectiveness of our \ours{} framework with datasets and experimental setup detailed in Sec.~\ref{sec:setup}, main results detailed in Sec.~\ref{sec:main result}, and ablation studies detailed in Sec.~\ref{ablation study}.
\begin{figure*}[t]
\centering
\includegraphics[width=1.0\textwidth]{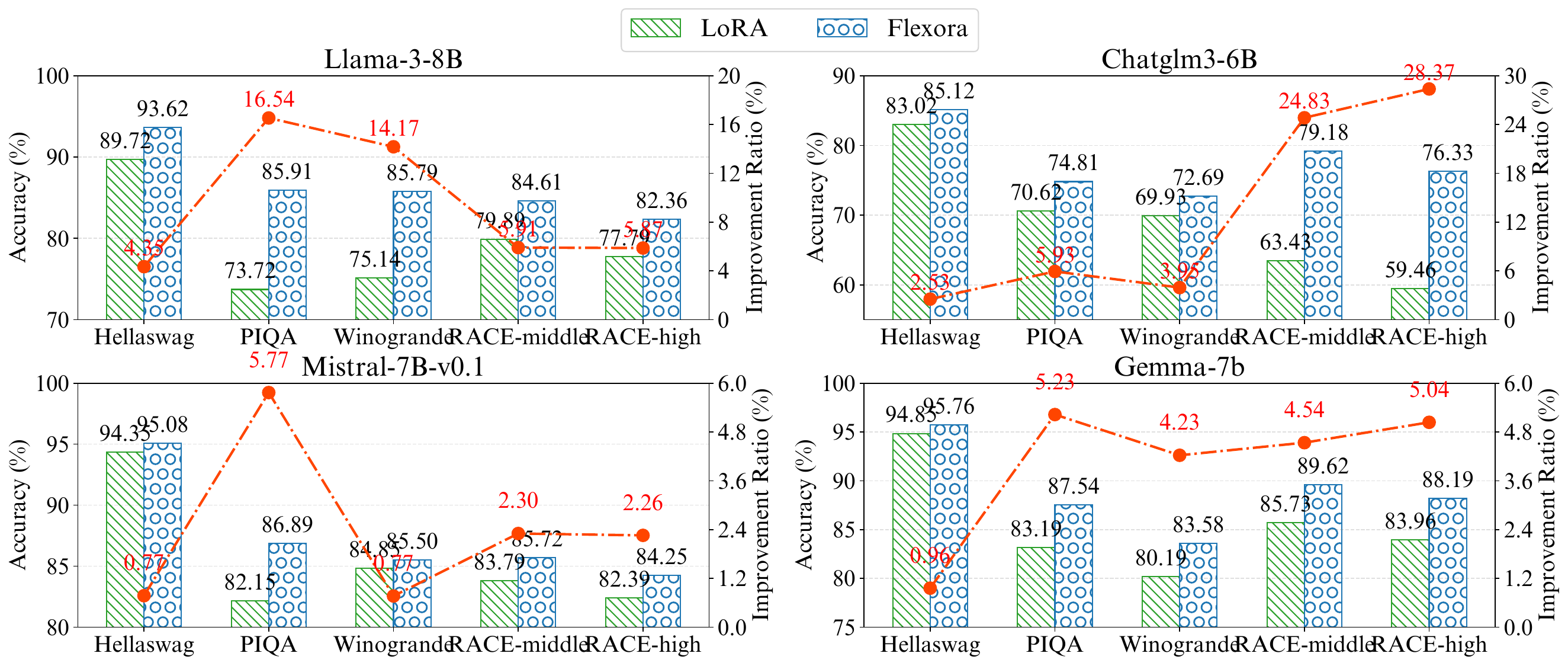} 
\caption{Comparison of the accuracy of various models (Llama-3-8B, ChatGLM3-6B, Mistral-7B-v0.1, and Gemma-7B) across different tasks. Bars with green diagonal stripes represent LoRA accuracy, while blue circles indicate \ours{} accuracy, and the red dotted line represents the improvement ratio of \ours{} over LoRA. Notably, \ours{} generally outperforms LoRA in most tasks and models, demonstrating its effectiveness.}
\label{fig3}
\vspace{-3mm}
\end{figure*}

\subsection{Datasets and Setup}\label{sec:setup}
To evaluate the performance of our proposed \ours{} method, we primarily focus on reasoning and reading comprehension tasks. Since \ours{} is the first algorithm to select layers based on specific downstream tasks, we refer to and modify the dataset selection process of \citet{hu-etal-2023-llm}. We selected the Winogrande\cite{c:74}, PIQA\cite{c:73}, and Hellaswag\cite{c:72} reasoning benchmarks as recommended by \citet{hu-etal-2023-llm}, and additionally included the reading comprehension benchmark RACE\cite{c:75}. Each of these datasets, measured by accuracy, has independent training, validation, and test sets. In all experiments, we use the training set to train the LoRA parameters, the validation set to tune the hyperparameters introduced by \ours{}, and finally the test set for evaluation. It is important to emphasize that the test set remains unseen during the training phase. Our experimental setup includes 11 mainstream large-scale language models (LLMs), such as Llama3-8B \citep{c:80} and others. Our \ours{} method is implemented on the Llama-factory framework\citep{c:98} and evaluated using the Opencompass framework\citep{c:99}. The benchmarks for comparison include pre-trained models, Full FT, LoRA, and various LoRA enhancement methods that reduce trainable parameters, such as LoRAPrune, AdaLoRA, LoRA-drop, and others.  Detailed descriptions of the experimental setup are provided in Appendix \ref{app:setup}. All experiments are conducted on a single NVIDIA A100 GPU.

\begin{table*}[t]
    \renewcommand\arraystretch{1.2}
    \scriptsize
    \centering
    \caption{Comparison of the accuracy of different randomly selected fine-tuning layers with the same number of fine-tuning layers. We fixed the number of fine-tuning layers to match the number selected by \ours{}, ensuring that the number of fine-tuning parameters remained constant while the layers were randomly selected for fine-tuning. 
    }
    \label{tab:ablation_studies_1}
    \begin{tabular}{l*{6}{c}}
        \toprule
        \textbf{Methods} & \textbf{Hellaswag} & \textbf{PIQA} & \textbf{Winogrande} & \textbf{RACE-mid}& \textbf{RACE-high} & \textbf{Average} \\
        \midrule
        Random 1 & 92.97 & 82.91 & 80.98  & 83.98 & 81.10 & 84.39\\
        Random 2 & 93.11 & 80.79 & 76.09  & \textbf{85.45} & 81.16 & 83.32\\
        Random 3 & 92.52 & 80.47 & 83.50  & 84.54 & 81.93 & 84.59\\
        \cdashline{0-6}[4pt/1pt]
        Random (Avg.) & 92.87 & 81.39 & 80.19  & 84.66 & 81.40 & 84.10 \\
        \ours{}  & \textbf{93.62} & \textbf{85.91} & \textbf{85.79} & 84.61 & \textbf{82.36} & \textbf{86.46} \\
        \bottomrule
    \end{tabular}
\end{table*}
\begin{table*}
    \scriptsize
    \centering
    \caption{Comparison of the performance of models with and without a fine-tuning phase on various tasks.
}
    \label{tab:fine-tuning phase}
    \begin{tabular}{lcccccc}
        \toprule
        \textbf{Methods} & \textbf{Hellaswag} & \textbf{PIQA} & \textbf{Winogrande} & \textbf{RACE-mid} & \textbf{RACE-high} & \textbf{Average} \\
        \midrule
        \ours{} (w/o Fine-Tuning Stage)& 48.93 & 80.20 & 66.38 & 62.72 & 60.76 & 63.80 \\
        \ours{} (w/ Fine-Tuning Stage) & \textbf{93.62} & \textbf{85.91} & \textbf{85.79} & \textbf{84.61} & \textbf{82.36} & \textbf{86.46}\\
        \bottomrule
    \end{tabular}
\end{table*}
\subsection{Main Results}\label{sec:main result}
In this section, we evaluate the performance improvement of \ours{} on Llama3-8B, and the results are listed in Table~\ref{tab:accuracy_comparison}. The loss metrics are discussed in Appendix~\ref{main loss}. The results show that \ours{} outperforms all baseline methods. Specifically, compared with full fine-tuning and LoRA, \ours{} fine-tunes 0.02\% and 50\% of its parameters, respectively, to achieve superior performance. This demonstrates that fine-tuning too many parameters can lead to overfitting, which not only fails to improve the performance of the model on downstream tasks but may also reduce the generalization ability of the model due to the overfitting effect. Therefore, it is crucial to select the layers most relevant to the downstream tasks for optimization. The flexible layer selection stage of \ours{} is able to consider the relationship between the pre-trained parameters of each LLM layer and the downstream task. This stage effectively identifies the most critical layers for various downstream tasks and minimizes the risk of model overfitting by focusing on training these layers, resulting in excellent performance. In Table~\ref{tab:accuracy_comparison}, we also compare \ours{} with other methods that attempt to enhance the model by reducing model parameters. Particularly, the experimental results using LoRAShear are detailed in Appendix~\ref{Comparison_lorashear} and the experimental results using \ours{} on full-parameter fine-tuning and on the instruction model are detailed in Appendix~\ref{full-parameter} and~\ref{instruction} respectively. The results show that \ours{} can most accurately identify the most important parameters to achieve the largest performance improvement. \ours{} introduces a flexible layer selection stage, but incurs no additional computational overhead (see Appendix~\ref{time} for details). \ours{} is a vertical method, and the discussion of integration with many LoRA enhancement methods is detailed in Appendix~\ref{scalability}. We also evaluate \ours{} at different LoRA ranks, and the results show that changing the rank has a negligible impact on the performance of \ours{}. The specific layers selected are listed in Table~\ref{tab:layer selection1}  in the Appendix. It is worth noting that the layers selected by \ours{} are roughly consistent under different level conditions, which shows that \ours{} effectively identifies the layers that are most suitable for downstream tasks. We also discuss the impact of different search samples on the training time and final performance of the flexible layer selection stage, as detailed in Appendix~\ref{search sample}. In addition, \ours{} shows strong generalization and scalability across different LLMs. As shown in Figure~\ref{fig3} and explained in detail in Appendix~\ref{Other_LLMs}, almost all LLMs can significantly improve performance with fewer fine-tuning parameters by leveraging \ours{}. In Appendix~\ref{sec:Special cases}, we compare \ours{} with LoRA using specific cases. The model fine-tuned with \ours{} outperforms LoRA on challenging cases and provides correct explanations for answers not seen in the training set, demonstrating its strong learning and generalization capabilities. Finally, we discuss the impact of two hyperparameters \(K\) and \(T\) introduced by Algorithm~\ref{algo:ud} on the results. The results show that changes in \(K\) and \(T\) have little effect on the layer selection results and model performance. For a more detailed discussion, see Appendix~\ref{sec:hyperparameters}.


\subsection{Ablation Studies} \label{ablation study}
\paragraph{Effective Layer Selection in \ours{}.} 
In the first ablation experiment, we maintained the number of layers selected by \ours{} unchanged but chose different layers for fine-tuning, aiming to verify whether \ours{} selected the right layers. The experimental results are shown in Table~\ref{tab:ablation_studies_1}. The result underscores two key points: First, \ours{} can precisely determine the number of layers for fine-tuning. Even when the specific fine-tuning layers are chosen at random, the results continue to outperform LoRA. The theoretical explanation for this result can be found in Sec.~\ref{theoretical^{(i)}nsights}. Secondly, \ours{} enables adaptive layer selection for fine-tuning, optimizing performance and generalization by focusing on crucial layers while mitigating local optima-induced performance degradation (see Appendix~\ref{sec:local} for details). Appendix~\ref{layer_analyis} provides an analysis of the characteristics of the selected layers in \ours{}, revealing a distinct layer selection pattern. The loss metrics are discussed in Appendix~\ref{ablation study 1 loss}.

\paragraph{Flexible Layer Selection in \ours{}.} 
In the second ablation experiment, we manually determine the number of fine-tuning layers and compare \ours{} with random selection, highlighting the flexibility of \ours{}. The results in Table~\ref{tab:ablation_studies_2} show that it can achieve the best performance regardless of the number of fine-tuning layers. The specific layers selected are shown in Table~\ref{tab:layer selection2}. The loss metrics are discussed in Appendix~\ref{ablation study 2 loss}. A noteworthy observation is that \ours{} usually chooses the initial and final layers. An intuitive explanation is that the initial and final layers of the model have a significant impact on the data. The initial layers directly contact the original input, while the final layers are related to the model output, rendering them crucial. In addition, for the same downstream task, the input of the initial layer is consistent and closely coupled to the task, and the output of the final layer is also consistent. Focusing on optimizing these layers can improve learning efficiency. This conclusion has also been confirmed by other studies. LoRAShear\cite{c:71} observed that the knowledge distribution in LLM is mainly concentrated in the initial and final layers. LASER\cite{c:45} revealed steep loss gradients in both initial and final layers, enhancing model training efficacy. LISA~\cite{c:113} found much higher weight norms in initial and final layers, indicating their increased importance.

\begin{table*}[t]
    \scriptsize
    \renewcommand\arraystretch{1.2}
    \centering
    \caption{Comparison of the accuracy of fine-tuning a subset of layers. We standardized the number of layers to be fine-tuned and compared the performance of layers selected by \ours{} against those selected randomly. 
    }
    \label{tab:ablation_studies_2}
    \begin{tabular}{l*{6}{c}}
        \toprule
        \textbf{Methods} & \textbf{Hellaswag} & \textbf{PIQA} & \textbf{Winogrande} & \textbf{RACE-mid}& \textbf{RACE-high} & \textbf{Average} \\
        \midrule
        Random (6 Layers) & 59.79 & 70.25 & 46.32 & 54.54 & 53.45 & 56.87 \\
        \ours{} (First 6 Layers) & \textbf{60.04} \textcolor{red}{(+0.25)} & \textbf{77.20} \textcolor{red}{(+6.95)} & \textbf{57.54} \textcolor{red}{(+11.22)} & \textbf{69.71} \textcolor{red}{(+15.17)} & \textbf{58.35} \textcolor{red}{(+4.90)} & \textbf{64.57} \textcolor{red}{(+7.70)} \\
        \cdashline{0-6}[4pt/1pt]
        Random (12 Layers) & 81.90 & 77.82 & 57.35 & 78.41 & 72.16 & 73.53 \\
        \ours{} (First 12 Layers) & \textbf{88.85} \textcolor{red}{(+6.95)} & \textbf{79.71 } \textcolor{red}{(+1.89)} & \textbf{65.82} \textcolor{red}{(+8.47)} & \textbf{79.42} \textcolor{red}{(+1.01)} & \textbf{72.33} \textcolor{red}{(+0.17)} & \textbf{77.23} \textcolor{red}{(+3.70)} \\
        \cdashline{0-6}[4pt/1pt]
        Random (18 Layers) & 91.15 & 81.54 & 83.58 & 83.77 & 81.22 & 84.25 \\
        \ours{} (First 18 Layers) & \textbf{91.31} \textcolor{red}{(+0.16)} & \textbf{82.21} \textcolor{red}{(+0.67)} & \textbf{84.69} \textcolor{red}{(+1.11)} & \textbf{84.07} \textcolor{red}{(+0.30)} & \textbf{81.53} \textcolor{red}{(+0.31)} & \textbf{84.76} \textcolor{red}{(+0.51)} \\
        \cdashline{0-6}[4pt/1pt]
        Random (24 Layers) & 90.58 & 80.90 & 82.16 & 82.19 & 79.22 & 83.01 \\
        \ours{} (First 24 Layers) & \textbf{91.01} \textcolor{red}{(+0.43)} & \textbf{81.21} \textcolor{red}{(+0.31)} & \textbf{82.87} \textcolor{red}{(+0.71)} & \textbf{83.53} \textcolor{red}{(+1.34)} & \textbf{80.22} \textcolor{red}{(+1.00)} & \textbf{83.77} \textcolor{red}{(+0.76)} \\
        \bottomrule
    \end{tabular}
\end{table*}

\begin{table*}
    \scriptsize
    \centering
    \caption{Performance comparison under different training configurations}
    \label{tab:epoch_ablation}
    \begin{tabular}{lcccccc}
        \toprule
        \textbf{Methods(Epoch)} & \textbf{Hellaswag} & \textbf{PIQA} & \textbf{Winogrande} & \textbf{RACE-mid} & \textbf{RACE-high} & \textbf{Average} \\
        \midrule
        LoRA (3) & 89.72 & 76.39 & 82.24 & 82.86 & 80.99 & 82.32 \\
        LoRA (4) & 89.74 & 76.27 & 82.47 & 82.73 & 81.04 & 82.45 \\
        \ours{} (1+3) & 93.62 \textcolor{red}{(+3.90)} & 85.91 \textcolor{red}{(+9.52)}  & 85.79 \textcolor{red}{(+3.55)} & 84.61 \textcolor{red}{(+1.75)} & 82.36 \textcolor{red}{(+1.37)} & 86.46 \textcolor{red}{(+4.14)} \\
        \bottomrule
    \end{tabular}
\end{table*}

\paragraph{Importance of the Fine-Tuning Stage} In the third ablation experiment, we investigated the significance of the Fine-Tuning Stage in the \ours{} method by comparing model performance from the Flexible Layer Selection Stage and the Fine-Tuning Stage on the test set. Results in Table~\ref{tab:fine-tuning phase} show that omitting the Fine-Tuning Stage significantly degrades performance. This is because the layer selection stage outputs continuous values $\widehat{\alpha}^{(i)} \in [0, 1]^N$ , while we need discrete $\alpha \in \{0, 1\}^N$ values. The discrepancy between continuous and discrete $\alpha$ values leads to a performance gap. The Fine-Tuning Stage is crucial as it addresses this gap by refining the model to better approximate the discrete $\alpha$ values, thereby mitigating the performance loss.

\paragraph{Performance of \ours{} from training framework}
To determine whether the performance of \ours{} enhancement primarily originates from its layer selection mechanism rather than an extended effective training duration (due to its two-stage process), we conducted a meticulous ablation study. A central question was whether exposing the model to the data set twice, once during layer selection (1 epoch) and again during fine-tuning (3 epochs) - artificially inflates the observed performance. We compared three distinct configurations:
a) \textbf{LoRA (3 epochs)}: The standard LoRA baseline, trained for 3 epochs.
b) \textbf{LoRA (4 epochs)}: An extended training control, designed to evaluate if merely increasing training epochs of LoRA to match \ours{}'s total fine-tuning-equivalent duration yields comparable gains.
c) \textbf{\ours{} (1+3 epochs)}: The proposed \ours{}  method, which dedicates 1 epoch to layer selection followed by 3 epochs of fine-tuning on the identified layers.

The results, detailed in Table~\ref{tab:epoch_ablation}, provide clear insights.
Extending LoRA training from 3 to 4 epochs (LoRA (4 epochs)) yielded a marginal average improvement of +0.13\%, suggesting that LoRA's performance largely converges by the third epoch for these tasks.
In contrast, \ours{} (1+3 epochs) significantly outperformed both LoRA configurations, achieving an average score of 86.46\%. This represents a substantial +4.14\% improvement over the LoRA (3 epochs) baseline and +4.01\% over LoRA (4 epochs). These findings robustly confirm that the primary driver of \ours{}'s superior performance is its layer selection mechanism, not simply the cumulative number of training iterations. The framework's capacity to identify and concentrate optimization efforts on task-critical layers results in significant performance enhancements that surpass those achievable by merely scaling up the training duration of standard LoRA.

\section{Theoretical Insights} \label{theoretical^{(i)}nsights}
In this section, we provide theoretical explanations for \textit{why \ours{} (using only a subset of LoRA layers) can achieve excellent results}. We first introduce Theorem \ref{theorem1} below, and then derive our general Proposition \ref{proposition2}, aiming to offer theoretical insights. 

\begin{theorem}[Theorem 3.8 in \cite{c:84}]\label{theorem1}
Assume that $f(\cdot ; z) \in[0, 1]$ is an $L$-Lipschitz and $\beta$-smooth loss function for every sample z. Suppose that we run stochastic gradient method (e.g., SGD) for $T$ steps with monotonically non-increasing step sizes $\eta_t \leq c / t \ (t \in [T])$, and the number of samples is $m$. In particular, omitting constant factors that depend on $\beta$, $c$, and $L$, we have
$
\gR^{\normalfont \text{test}}(\theta, \eta) \leq \gR^{\normalfont \text{train}}(\theta, \eta) + \frac{T^{1 - 1/(\beta c + 1)}}{m}
$.
\end{theorem}

Theorem \ref{theorem1} reveals that if all the conditions except for $\beta$ in Theorem \ref{theorem1} remain the same, a smaller smoothness $\beta$ will typically result in a smaller test error \(\gR^{\text{test}}(\theta, \eta)\), indicating a better generalization performance in practice. The specific definition of smoothness $\beta$ can be found in Appendix \ref{bate definition}. To show how the number of LoRA layers is related to this $\beta$, we then follow the practice in \citep{c:88} to prove our Proposition \ref{proposition2} below.
\begin{proposition} \label{proposition2}
For an \(N\)-layer linear multi-layer perceptron (MLP): \(y^{(N)} \triangleq \prod_{j=1}^{N} W^{(j)} \boldsymbol{x}\) with MSE function $\ell \triangleq (y^{(N)} - y)^2 / 2$ where $y$ denotes the true label, let \(\lambda^{(i)} = \left\| W^{(i)} \right\|\) for any $i \in [N]$, we then have
$\left\| \frac{\partial \ell}{\partial W^{(i)}_1} - \frac{\partial \ell}{\partial W^{(i)}_2} \right\| \leq \left( \prod_{j=1, j\neq i}^{N} \lambda^{(j)} \right)^2\left\|\boldsymbol{x}\right\|^2 \left\|  W^{(i)}_1 - W^{(i)}_2  \right\|$.
\end{proposition}
The proof of Proposition~\ref{proposition2} is in Appendix~\ref{proof_pro2}. Given Proposition~\ref{proposition2}, the block-wise smoothness $\beta_i^{(N)}$ on layer $i \in [N]$ of an \(N\)-th layer MLP can be bounded by:
$\beta_i^{(N)} \ \leq \left( \prod_{j=1, j\neq i}^{N} \lambda^{(j)} \right)^2\left\|\boldsymbol{x}\right\|^2 $.
From this bound, we can see that as the number of layers \( N \) increases, the upper bound of \( \beta_i^{(N)} \) will also be increasing as $\lambda^{(i)} > 1$ for \( i \in [N] \). Thus, shallow MLP of fewer layers are more likely to have smaller overall smoothness \( \beta \). 
Thanks to this smaller overall smoothness \( \beta \), shallow MLP of fewer layers are more likely to achieve a smaller generalization gap (i.e., the second term on the right-hand side of Theorem \ref{theorem1}) than deep MLP with more layers. When the training error \( \gR^{\text{train}}(\theta, \eta) \) is the same, that is, both shallow and deep MLPs are fully trained to converge, the shallower MLP may have a lower test error \( \gR^{\text{test}}(\theta, \eta) \) and thus may exhibit better performance on downstream tasks.  To demonstrate that Proposition~\ref{proposition2} is also applicable to the Transformer model and LoRA method, we present theoretical insights and experiments in Appendix~\ref{sec:numerical experiments}. These experiments demonstrate that, the smoothness of the Transformer model also increases exponentially with the number of layers.

We can now answer the question posed earlier. \ours{} employs LoRA adapters to a subset of LLM layers, effectively reducing the smoothness of the network. When sufficiently trained to convergence, the aforementioned theory suggests that networks with less smoothness are more likely to better generalization and performance on downstream tasks. In summary, the reason \ours{} achieves excellent results is that it makes the model more suitable for downstream tasks.

\section{Conclusion}
We introduce \ours{}, a method to improve the efficiency and effectiveness of fine-tuning in LLMs by automatically selecting key layers, by formulating layer selection as an HPO problem, and using UD. Experiments show \ours{} decreases parameters , mitigates overfitting, is scalable and outperforms baselines.

\newpage
\section*{Limitations}
In this section, we aim to highlight some potential considerations that may lead to suboptimal performance of \ours{}. The layer selection strategy in \ours{} is primarily based on the magnitude of the optimized hyperparameters. If the validation set used for optimizing these hyperparameters is too small, especially when the downstream task is complex, it may result in the optimization process converging to a hyperparameter gap that is too narrow. In such cases, the layer selection strategy may fail, leading to the incorrect choice of layers for subsequent optimization stages, ultimately resulting in poor performance. To address the issue of having a minimal validation set for different datasets, we conducted additional experiments on search samples, as detailed in Appendix~\ref{search sample}. These experiments demonstrate that an insufficient number of samples can indeed lead to poor performance. However, this issue can be mitigated by increasing the number of search samples. Furthermore, although \ours{} is a vertical method and can theoretically be combined with all LoRA methods, there are certain methods for which fine-tuning only specific layers significantly impacts the model's fine-tuning effectiveness. In such cases, these methods may not be compatible with \ours{}.

\section*{Ethics Statement}
We have manually reevaluated the dataset we created to ensure it is free of any potential for discrimination, human rights violations, bias, exploitation, and any other ethical concerns.
\newpage
\medskip
\bibliography{anthology}

\begin{thebibliography}{48}
\expandafter\ifx\csname natexlab\endcsname\relax\def\natexlab#1{#1}\fi

\bibitem[{Bao et~al.(2021)Bao, Wu, Li, Zhu, and Zhang}]{c:69}
Fan Bao, Guoqiang Wu, Chongxuan Li, Jun Zhu, and Bo~Zhang. 2021.
\newblock \href {http://arxiv.org/abs/2106.04188} {Stability and generalization of bilevel programming in hyperparameter optimization}.

\bibitem[{Bisk et~al.(2019)Bisk, Zellers, Bras, Gao, and Choi}]{c:73}
Yonatan Bisk, Rowan Zellers, Ronan~Le Bras, Jianfeng Gao, and Yejin Choi. 2019.
\newblock \href {http://arxiv.org/abs/1911.11641} {Piqa: Reasoning about physical commonsense in natural language}.

\bibitem[{Blair(1985)}]{c:85}
Charles Blair. 1985.
\newblock Problem complexity and method efficiency in optimization (as nemirovsky and db yudin).
\newblock \emph{Siam Review}, 27(2):264.

\bibitem[{Chen et~al.(2023)Chen, Ding, Yadav, Zharkov, and Liang}]{c:71}
Tianyi Chen, Tianyu Ding, Badal Yadav, Ilya Zharkov, and Luming Liang. 2023.
\newblock \href {http://arxiv.org/abs/2310.18356} {Lorashear: Efficient large language model structured pruning and knowledge recovery}.

\bibitem[{Contributors(2023)}]{c:99}
OpenCompass Contributors. 2023.
\newblock Opencompass: A universal evaluation platform for foundation models.
\newblock \url{https://github.com/open-compass/opencompass}.

\bibitem[{Franceschi et~al.(2017)Franceschi, Donini, Frasconi, and Pontil}]{c:116}
Luca Franceschi, Michele Donini, Paolo Frasconi, and Massimiliano Pontil. 2017.
\newblock \href {http://arxiv.org/abs/1703.01785} {Forward and reverse gradient-based hyperparameter optimization}.

\bibitem[{Franceschi et~al.(2018)Franceschi, Frasconi, Salzo, Grazzi, and Pontil}]{c:117}
Luca Franceschi, Paolo Frasconi, Saverio Salzo, Riccardo Grazzi, and Massimilano Pontil. 2018.
\newblock \href {http://arxiv.org/abs/1806.04910} {Bilevel programming for hyperparameter optimization and meta-learning}.

\bibitem[{Fu et~al.(2016)Fu, Luo, Feng, Low, and Chua}]{c:118}
Jie Fu, Hongyin Luo, Jiashi Feng, Kian~Hsiang Low, and Tat-Seng Chua. 2016.
\newblock \href {http://arxiv.org/abs/1601.00917} {Drmad: Distilling reverse-mode automatic differentiation for optimizing hyperparameters of deep neural networks}.

\bibitem[{Guo et~al.(2020)Guo, Zhang, Jiang, Li, and Zhou}]{c:49}
Lan-Zhe Guo, Zhen-Yu Zhang, Yuan Jiang, Yu-Feng Li, and Zhi-Hua Zhou. 2020.
\newblock Safe deep semi-supervised learning for unseen-class unlabeled data.
\newblock In \emph{International conference on machine learning}, pages 3897--3906. PMLR.

\bibitem[{Hardt et~al.(2016)Hardt, Recht, and Singer}]{c:84}
Moritz Hardt, Ben Recht, and Yoram Singer. 2016.
\newblock Train faster, generalize better: Stability of stochastic gradient descent.
\newblock In \emph{International conference on machine learning}, pages 1225--1234. PMLR.

\bibitem[{Hu et~al.(2021)Hu, Shen, Wallis, Allen-Zhu, Li, Wang, Wang, and Chen}]{c:36}
Edward~J. Hu, Yelong Shen, Phillip Wallis, Zeyuan Allen-Zhu, Yuanzhi Li, Shean Wang, Lu~Wang, and Weizhu Chen. 2021.
\newblock \href {http://arxiv.org/abs/2106.09685} {Lora: Low-rank adaptation of large language models}.

\bibitem[{Hu et~al.(2023)Hu, Wang, Lan, Xu, Lim, Bing, Xu, Poria, and Lee}]{hu-etal-2023-llm}
Zhiqiang Hu, Lei Wang, Yihuai Lan, Wanyu Xu, Ee-Peng Lim, Lidong Bing, Xing Xu, Soujanya Poria, and Roy Lee. 2023.
\newblock \href {https://doi.org/10.18653/v1/2023.emnlp-main.319} {{LLM}-adapters: An adapter family for parameter-efficient fine-tuning of large language models}.
\newblock In \emph{Proceedings of the 2023 Conference on Empirical Methods in Natural Language Processing}, pages 5254--5276, Singapore. Association for Computational Linguistics.

\bibitem[{Kalajdzievski(2023)}]{kalajdzievski2023rankstabilizationscalingfactor}
Damjan Kalajdzievski. 2023.
\newblock \href {http://arxiv.org/abs/2312.03732} {A rank stabilization scaling factor for fine-tuning with lora}.

\bibitem[{Kopiczko et~al.(2024)Kopiczko, Blankevoort, and Asano}]{kopiczko2024vera}
Dawid~Jan Kopiczko, Tijmen Blankevoort, and Yuki~M Asano. 2024.
\newblock \href {https://openreview.net/forum?id=NjNfLdxr3A} {Ve{RA}: Vector-based random matrix adaptation}.
\newblock In \emph{The Twelfth International Conference on Learning Representations}.

\bibitem[{Lai et~al.(2017)Lai, Xie, Liu, Yang, and Hovy}]{c:75}
Guokun Lai, Qizhe Xie, Hanxiao Liu, Yiming Yang, and Eduard Hovy. 2017.
\newblock \href {http://arxiv.org/abs/1704.04683} {Race: Large-scale reading comprehension dataset from examinations}.

\bibitem[{Lester et~al.(2021)Lester, Al-Rfou, and Constant}]{c:39}
Brian Lester, Rami Al-Rfou, and Noah Constant. 2021.
\newblock \href {http://arxiv.org/abs/2104.08691} {The power of scale for parameter-efficient prompt tuning}.

\bibitem[{Li and Liang(2021)}]{c:38}
Xiang~Lisa Li and Percy Liang. 2021.
\newblock \href {http://arxiv.org/abs/2101.00190} {Prefix-tuning: Optimizing continuous prompts for generation}.

\bibitem[{Lin et~al.(2024)Lin, Ma, Chu, Jin, Yang, Wang, and Mei}]{c:110}
Yang Lin, Xinyu Ma, Xu~Chu, Yujie Jin, Zhibang Yang, Yasha Wang, and Hong Mei. 2024.
\newblock \href {http://arxiv.org/abs/2404.09610} {Lora dropout as a sparsity regularizer for overfitting control}.

\bibitem[{Liu et~al.(2019)Liu, Simonyan, and Yang}]{c:81}
Hanxiao Liu, Karen Simonyan, and Yiming Yang. 2019.
\newblock \href {http://arxiv.org/abs/1806.09055} {Darts: Differentiable architecture search}.

\bibitem[{Liu et~al.(2024)Liu, Kundu, Li, Wan, Jiang, and Beerel}]{c:115}
Zeyu Liu, Souvik Kundu, Anni Li, Junrui Wan, Lianghao Jiang, and Peter~Anthony Beerel. 2024.
\newblock \href {http://arxiv.org/abs/2403.13269} {Aflora: Adaptive freezing of low rank adaptation in parameter efficient fine-tuning of large models}.

\bibitem[{Maclaurin et~al.(2015)Maclaurin, Duvenaud, and Adams}]{c:119}
Dougal Maclaurin, David Duvenaud, and Ryan~P. Adams. 2015.
\newblock \href {http://arxiv.org/abs/1502.03492} {Gradient-based hyperparameter optimization through reversible learning}.

\bibitem[{Mao et~al.(2024{\natexlab{a}})Mao, Huang, Guan, Bao, Mo, and Xu}]{c:43}
Yulong Mao, Kaiyu Huang, Changhao Guan, Ganglin Bao, Fengran Mo, and Jinan Xu. 2024{\natexlab{a}}.
\newblock \href {http://arxiv.org/abs/2405.17357} {Dora: Enhancing parameter-efficient fine-tuning with dynamic rank distribution}.

\bibitem[{Mao et~al.(2024{\natexlab{b}})Mao, Ping, Zhao, Liu, and Ding}]{c:83}
Yuzhu Mao, Siqi Ping, Zihao Zhao, Yang Liu, and Wenbo Ding. 2024{\natexlab{b}}.
\newblock \href {http://arxiv.org/abs/2407.12074} {Enhancing parameter efficiency and generalization in large-scale models: A regularized and masked low-rank adaptation approach}.

\bibitem[{Meta(2024)}]{c:80}
Meta. 2024.
\newblock Introducing meta llama 3: The most capable openly available {LLM} to date.
\newblock \emph{Meta Blog}.

\bibitem[{Olson et~al.(2016)Olson, Bartley, Urbanowicz, and Moore}]{c:100}
Randal~S. Olson, Nathan Bartley, Ryan~J. Urbanowicz, and Jason~H. Moore. 2016.
\newblock \href {http://arxiv.org/abs/1603.06212} {Evaluation of a tree-based pipeline optimization tool for automating data science}.

\bibitem[{Pan et~al.(2024)Pan, Liu, Diao, Pi, Zhang, Han, and Zhang}]{c:113}
Rui Pan, Xiang Liu, Shizhe Diao, Renjie Pi, Jipeng Zhang, Chi Han, and Tong Zhang. 2024.
\newblock \href {http://arxiv.org/abs/2403.17919} {Lisa: Layerwise importance sampling for memory-efficient large language model fine-tuning}.

\bibitem[{Sakaguchi et~al.(2019)Sakaguchi, Bras, Bhagavatula, and Choi}]{c:74}
Keisuke Sakaguchi, Ronan~Le Bras, Chandra Bhagavatula, and Yejin Choi. 2019.
\newblock \href {http://arxiv.org/abs/1907.10641} {Winogrande: An adversarial winograd schema challenge at scale}.

\bibitem[{Shaban et~al.(2019)Shaban, Cheng, Hatch, and Boots}]{c:120}
Amirreza Shaban, Ching-An Cheng, Nathan Hatch, and Byron Boots. 2019.
\newblock \href {http://arxiv.org/abs/1810.10667} {Truncated back-propagation for bilevel optimization}.

\bibitem[{Sharma et~al.(2023)Sharma, Ash, and Misra}]{c:45}
Pratyusha Sharma, Jordan~T. Ash, and Dipendra Misra. 2023.
\newblock \href {http://arxiv.org/abs/2312.13558} {The truth is in there: Improving reasoning in language models with layer-selective rank reduction}.

\bibitem[{Shu et~al.(2020)Shu, Wang, and Cai}]{c:88}
Yao Shu, Wei Wang, and Shaofeng Cai. 2020.
\newblock \href {http://arxiv.org/abs/1909.09569} {Understanding architectures learnt by cell-based neural architecture search}.

\bibitem[{Sra et~al.(2011)Sra, Nowozin, and Wright}]{c:121}
Suvrit Sra, Sebastian Nowozin, and Stephen~J Wright. 2011.
\newblock \emph{Optimization for machine learning}, page 351–368. Mit Press.

\bibitem[{Touvron et~al.(2023)Touvron, Lavril, Izacard, Martinet, Lachaux, Lacroix, Rozière, Goyal, Hambro, Azhar, Rodriguez, Joulin, Grave, and Lample}]{c:29}
Hugo Touvron, Thibaut Lavril, Gautier Izacard, Xavier Martinet, Marie-Anne Lachaux, Timothée Lacroix, Baptiste Rozière, Naman Goyal, Eric Hambro, Faisal Azhar, Aurelien Rodriguez, Armand Joulin, Edouard Grave, and Guillaume Lample. 2023.
\newblock \href {http://arxiv.org/abs/2302.13971} {Llama: Open and efficient foundation language models}.

\bibitem[{Wei et~al.(2025{\natexlab{a}})Wei, Shu, Ou, He, and Yu}]{wei2025paftpromptagnosticfinetuning}
Chenxing Wei, Yao Shu, Mingwen Ou, Ying~Tiffany He, and Fei~Richard Yu. 2025{\natexlab{a}}.
\newblock \href {http://arxiv.org/abs/2502.12859} {Paft: Prompt-agnostic fine-tuning}.

\bibitem[{Wei et~al.(2025{\natexlab{b}})Wei, Wang, He, Shu, and Yu}]{wei2025tmpo}
Chenxing Wei, Hong Wang, Ying~Tiffany He, Yao Shu, and Fei Yu. 2025{\natexlab{b}}.
\newblock \href {https://openreview.net/forum?id=6eN5FZJuLI} {{TMPO}: Test-time multi-turn policy optimization}.
\newblock In \emph{First Workshop on Multi-Turn Interactions in Large Language Models}.

\bibitem[{Wei et~al.(2025{\natexlab{c}})Wei, Yu, He, Dong, Shu, and Yu}]{wei2025redit}
Chenxing Wei, Jiarui Yu, Ying~Tiffany He, Hande Dong, Yao Shu, and Fei Yu. 2025{\natexlab{c}}.
\newblock \href {https://openreview.net/forum?id=nDzBpkbxk7} {Redit: Reward dithering for improved {LLM} policy optimization}.
\newblock In \emph{2nd Workshop on Models of Human Feedback for AI Alignment}.

\bibitem[{Wei et~al.(2022)Wei, Tay, Bommasani, Raffel, Zoph, Borgeaud, Yogatama, Bosma, Zhou, Metzler, Chi, Hashimoto, Vinyals, Liang, Dean, and Fedus}]{c:25}
Jason Wei, Yi~Tay, Rishi Bommasani, Colin Raffel, Barret Zoph, Sebastian Borgeaud, Dani Yogatama, Maarten Bosma, Denny Zhou, Donald Metzler, Ed~H. Chi, Tatsunori Hashimoto, Oriol Vinyals, Percy Liang, Jeff Dean, and William Fedus. 2022.
\newblock \href {http://arxiv.org/abs/2206.07682} {Emergent abilities of large language models}.

\bibitem[{Wu et~al.(2024{\natexlab{a}})Wu, Wang, Zhao, and Wong}]{wu2024mixtureofsubspaceslowrankadaptation}
Taiqiang Wu, Jiahao Wang, Zhe Zhao, and Ngai Wong. 2024{\natexlab{a}}.
\newblock \href {http://arxiv.org/abs/2406.11909} {Mixture-of-subspaces in low-rank adaptation}.

\bibitem[{Wu et~al.(2024{\natexlab{b}})Wu, Xiang, Huo, Gong, and Liang}]{c:60}
Yichao Wu, Yafei Xiang, Shuning Huo, Yulu Gong, and Penghao Liang. 2024{\natexlab{b}}.
\newblock \href {http://arxiv.org/abs/2403.08822} {Lora-sp: Streamlined partial parameter adaptation for resource-efficient fine-tuning of large language models}.

\bibitem[{Wu et~al.(2024{\natexlab{c}})Wu, Arora, Wang, Geiger, Jurafsky, Manning, and Potts}]{wu2024reftrepresentationfinetuninglanguage}
Zhengxuan Wu, Aryaman Arora, Zheng Wang, Atticus Geiger, Dan Jurafsky, Christopher~D. Manning, and Christopher Potts. 2024{\natexlab{c}}.
\newblock \href {http://arxiv.org/abs/2404.03592} {Reft: Representation finetuning for language models}.

\bibitem[{Xu et~al.(2023)Xu, Xie, Qin, Tao, and Wang}]{c:24}
Lingling Xu, Haoran Xie, Si-Zhao~Joe Qin, Xiaohui Tao, and Fu~Lee Wang. 2023.
\newblock \href {http://arxiv.org/abs/2312.12148} {Parameter-efficient fine-tuning methods for pretrained language models: A critical review and assessment}.

\bibitem[{Zellers et~al.(2019)Zellers, Holtzman, Bisk, Farhadi, and Choi}]{c:72}
Rowan Zellers, Ari Holtzman, Yonatan Bisk, Ali Farhadi, and Yejin Choi. 2019.
\newblock \href {http://arxiv.org/abs/1905.07830} {Hellaswag: Can a machine really finish your sentence?}

\bibitem[{Zhang et~al.(2023{\natexlab{a}})Zhang, Zhang, Shi, Chu, and Li}]{c:61}
Longteng Zhang, Lin Zhang, Shaohuai Shi, Xiaowen Chu, and Bo~Li. 2023{\natexlab{a}}.
\newblock \href {http://arxiv.org/abs/2308.03303} {Lora-fa: Memory-efficient low-rank adaptation for large language models fine-tuning}.

\bibitem[{Zhang et~al.(2024)Zhang, Chen, Shen, Yang, Ou, Yu, and Zhuang}]{c:66}
Mingyang Zhang, Hao Chen, Chunhua Shen, Zhen Yang, Linlin Ou, Xinyi Yu, and Bohan Zhuang. 2024.
\newblock \href {http://arxiv.org/abs/2305.18403} {Loraprune: Pruning meets low-rank parameter-efficient fine-tuning}.

\bibitem[{Zhang et~al.(2023{\natexlab{b}})Zhang, Chen, Bukharin, Karampatziakis, He, Cheng, Chen, and Zhao}]{c:41}
Qingru Zhang, Minshuo Chen, Alexander Bukharin, Nikos Karampatziakis, Pengcheng He, Yu~Cheng, Weizhu Chen, and Tuo Zhao. 2023{\natexlab{b}}.
\newblock \href {http://arxiv.org/abs/2303.10512} {Adalora: Adaptive budget allocation for parameter-efficient fine-tuning}.

\bibitem[{Zhao et~al.(2023)Zhao, Zhou, Li, Tang, Wang, Hou, Min, Zhang, Zhang, Dong, Du, Yang, Chen, Chen, Jiang, Ren, Li, Tang, Liu, Liu, Nie, and Wen}]{c:23}
Wayne~Xin Zhao, Kun Zhou, Junyi Li, Tianyi Tang, Xiaolei Wang, Yupeng Hou, Yingqian Min, Beichen Zhang, Junjie Zhang, Zican Dong, Yifan Du, Chen Yang, Yushuo Chen, Zhipeng Chen, Jinhao Jiang, Ruiyang Ren, Yifan Li, Xinyu Tang, Zikang Liu, Peiyu Liu, Jian-Yun Nie, and Ji-Rong Wen. 2023.
\newblock \href {http://arxiv.org/abs/2303.18223} {A survey of large language models}.

\bibitem[{Zheng et~al.(2024)Zheng, Zhang, Zhang, Ye, Luo, Feng, and Ma}]{c:98}
Yaowei Zheng, Richong Zhang, Junhao Zhang, Yanhan Ye, Zheyan Luo, Zhangchi Feng, and Yongqiang Ma. 2024.
\newblock \href {http://arxiv.org/abs/2403.13372} {Llamafactory: Unified efficient fine-tuning of 100+ language models}.
\newblock In \emph{Proceedings of the 62nd Annual Meeting of the Association for Computational Linguistics (Volume 3: System Demonstrations)}, Bangkok, Thailand. Association for Computational Linguistics.

\bibitem[{Zhou et~al.(2024)Zhou, Lu, Xu, Zhu, Zhao, and Yang}]{c:65}
Hongyun Zhou, Xiangyu Lu, Wang Xu, Conghui Zhu, Tiejun Zhao, and Muyun Yang. 2024.
\newblock \href {http://arxiv.org/abs/2402.07721} {Lora-drop: Efficient lora parameter pruning based on output evaluation}.

\bibitem[{Zhu et~al.(2023)Zhu, Yang, Wu, and Zhang}]{c:111}
Yunqi Zhu, Xuebing Yang, Yuanyuan Wu, and Wensheng Zhang. 2023.
\newblock \href {http://arxiv.org/abs/2305.08285} {Parameter-efficient fine-tuning with layer pruning on free-text sequence-to-sequence modeling}.

\end{thebibliography}

\newpage
\onecolumn
\begin{appendices}

\section{Theorems and proofs} \label{sec:math}
We first prove Proposition~\ref{proposition1}, then introduce the theorems proposed by \cite{c:85} and \cite{c:84}, which reveal the properties of $\beta$-smooth, a necessary theoretical basis for proving Proposition~\ref{proposition2}. Finally, we prove Proposition~\ref{proposition2}.
\subsection{Proof of proposition~\ref{proposition1}}\label{proof_pro1}
The proof of Proposition~\ref{proposition1} is expressed as follows:
\begin{proof}\label{proof_pos1}
It is easy to verify that 
\begin{equation*}
    \frac{\partial \widehat{\alpha}^{(j)}}{\partial \alpha^{(i)}} = \begin{cases}
    \widehat{\alpha}^{(j)}(1 - \frac{1}{n}\widehat{\alpha}^{(j)}) ,& \text{if } j=i\\
    -\frac{1}{n}\widehat{\alpha}^{(j)}\widehat{\alpha}^{(i)},              & \text{if } j \neq i
    \end{cases}.
\end{equation*}

Therefore, given that $\sum_{i=1}^n \widehat{\alpha}^{(i)} = n$
\begin{equation*}
\begin{aligned}
    \sum_{i=1}^n\frac{\partial \widehat{\gR}^{\text{val}}}{\partial{\widehat{\alpha}^{(j)}}}\frac{\partial \widehat{\alpha}^{(j)}}{\partial \alpha^{(i)}} &= \frac{\partial \widehat{\gR}^{\text{val}}}{\partial{\widehat{\alpha}^{(j)}}}\left(\widehat{\alpha}^{(j)} - \frac{1}{n}\left(\widehat{\alpha}^{(j)}\right)^2 - \frac{1}{n}\sum_{i=1, i\neq j}^n\widehat{\alpha}^{(j)}\widehat{\alpha}^{(i)}\right) \\
    &= \frac{\partial \widehat{\gR}^{\text{val}}}{\partial{\widehat{\alpha}^{(j)}}}\left(\widehat{\alpha}^{(j)} - \frac{\widehat{\alpha}^{(j)}}{n}\sum_{i=1}^n \widehat{\alpha}^{(i)}\right) \\[8pt]
    &= 0 \ .
\end{aligned}
\end{equation*}

When applying SGD to update $\alpha$, we have
\begin{equation*}
    \sum_{i=1}^{n} \alpha^{(i)} - \eta \sum_{i=1}^n\sum_{j=1}^n\frac{\partial \widehat{\gR}^{\text{val}}}{\partial{\widehat{\alpha}^{(j)}}}\frac{\partial \widehat{\alpha}^{(j)}}{\partial \alpha^{(i)}} = \sum_{i=1}^{n} \alpha^{(i)} \ .
\end{equation*}

That is, the updated $\alpha$ shares the same summation as the one before the updates, which therefore concludes our proof.
\end{proof}

\subsection{\texorpdfstring{Definition of $\beta$-Smooth}{beta-smooth}} \label{bate definition}
\begin{definition}\label{definition1}
$\beta$-smooth refers to the Lipschitz continuity of the gradient of the loss function, that is, for all \(w\) and \(w'\):
$$
\|\nabla f(w; z) - \nabla f(w'; z)\| \leq \beta \|w - w'\|
$$
where \( \|\cdot\| \) denotes the norm of the vector, and \( f(w; z) \) is the loss function with parameter \( w \) for sample \( z \).
\end{definition}

Let \(f_{\text{deep}}(w) \) and \(f_{\text{shallow}}(w) \) be the loss functions for deep and shallow architectures, respectively. According to Definition~\ref{definition1}, the relationship between \(\beta_{\text{deep}}\) and \(\beta_{\text{shallow}}\) illustrates the relationship between the generalization and performance of deep and shallow networks.

\subsection{Proof of proposition~\ref{proposition2}}\label{proof_pro2}
\begin{figure}[t]
\vspace{-3mm}
\centering
\includegraphics[width=0.22\textwidth]{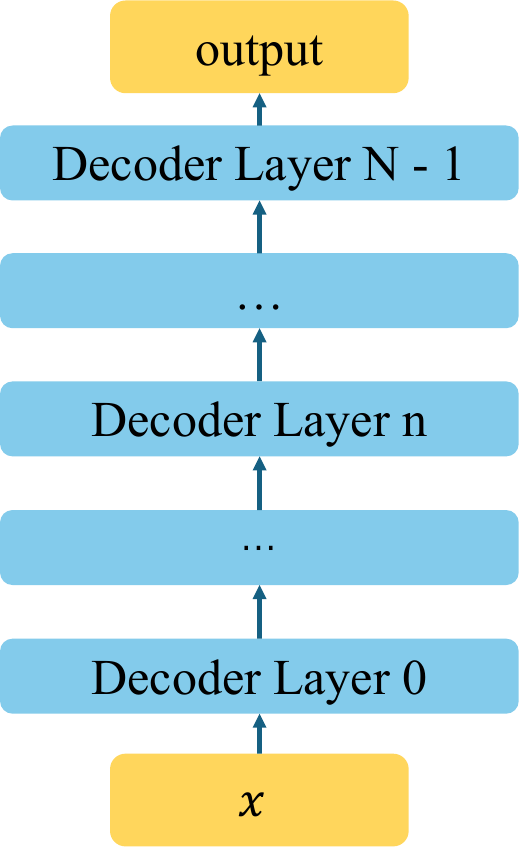} 
\caption{We present LLM as a hierarchical network. In this context, all parameters of a Decoder layer are represented as a weight matrix $W$ for subsequent analysis.}
\label{fig4}
\vspace{-4mm}
\end{figure}
\textbf{Abstract LLM into a layered network:}\cite{c:88} As shown in Figure~\ref{fig4}, we abstract LLM into a hierarchical network, and the weight of each layer is represented by \(W^{(i)} \). Figure~\ref{fig4} represents the general case. The output of the \(i\)-th layer network is: 
\begin{equation}\label{equ: networkb}
\boldsymbol{y} = \prod_{j=1}^{n} W^{(j)} \boldsymbol{x}.
\end{equation}

\textbf{Gradient analysis:} For the abstract network, represented in Equation~\ref{equ: networkb}. The gradient of the loss function \(\ell\) with respect to the weight \(W^{(i)}\) is:
\begin{align}
\frac{\partial \ell}{\partial W^{(i)}} &= \left( \prod_{j=i+1}^{n} W^{(j)} \right) \frac{\partial \ell}{\partial \boldsymbol{y}^{(i)}} \boldsymbol{x} \left( \prod_{j=1}^{i-1} W^{(j)} \right).
\end{align}

The proof of Proposition~\ref{proposition2} is expressed as follows:
\begin{proof}\label{proof_bate}
For the abstract network, we begin with Definition~\ref{definition1}: 
\begin{equation}\label{equ:11}
\begin{aligned}
&\left\| \frac{\partial \ell}{\partial W^{(i)}_1} - \frac{\partial \ell}{\partial W^{(i)}_2} \right\|\\
& = \left\| \left( \prod_{j=i+1}^{n} W^{(j)} \right) \left( \frac{\partial \ell}{\partial \boldsymbol{y}^{(i)}_1} - \frac{\partial \ell}{\partial \boldsymbol{y}^{(i)}_2} \right)\boldsymbol{x} \left( \prod_{j=1}^{i-1} W^{(j)} \right) \right\| .
\end{aligned}
\end{equation}

Taking MSE Loss as an example, for one predictions $y^{(N)}$ and their corresponding true values $y$: 
\begin{equation} 
\ell \triangleq (\boldsymbol{y}^{(N)} - \boldsymbol{y})^2 / 2,
\end{equation} 
therefore: 
\begin{equation} \label{equ:13}
\frac{\partial \ell}{\partial \boldsymbol{y}^{(i)}} = \left( \boldsymbol{y}^{(N)} - \boldsymbol{y} \right) \prod_{j=i+1}^{N} W^{(j)}.
\end{equation} 
We select the MSE loss function and calculat the \(i\)-th layer of N layers network, Substituting Equation~\ref{equ:13} into Equation~\ref{equ:11}:
 
\begin{equation}
\begin{aligned}
\left\| \frac{\partial \ell}{\partial W^{(i)}_1} - \frac{\partial \ell}{\partial W^{(i)}_2} \right\| &= \left\| \left( \prod_{j=i+1}^{N} W^{(j)} \right)^2  \left(\boldsymbol{y}^{(N)}_1 - \boldsymbol{y} - \boldsymbol{y}^{(N)}_2 + \boldsymbol{y} \right) \boldsymbol{x}\left( \prod_{j=1}^{i-1} W^{(j)} \right) \right\| \\
& \leq  \left( \frac{1}{\lambda^{(i)}} \left( \prod_{j=1}^{N} \lambda^{(j)} \right) \right) \left( \prod_{j=i + 1}^{N} \lambda^{(j)} \right) \left\| \prod_{j=1}^{i-1} W^{(j)} \left(W^{(i)}_1 - W^{(i)}_2 \right)  {\boldsymbol{x}^2}  \right\|\\
& \leq   \left( \prod_{j=1, j\neq i}^{N} \lambda^{(j)} \right) \left\|{\boldsymbol{x}^2}\right\|  \left( \prod_{j=i + 1}^{N} \lambda^{(j)} \right) \left( \prod_{j=1}^{i - 1} \lambda^{(j)} \right) \left\| \left(W^{(i)}_1  - W^{(i)}_2 \right) \right\|\\
& \leq  \left( \prod_{j=1, j\neq i}^{N} \lambda^{(j)} \right)^2  \left\|{\boldsymbol{x}^2}\right\| \left\|  W^{(i)}_1 - W^{(i)}_2  \right\| \ ,
\end{aligned}
\end{equation}
which therefore concludes our proof.
\end{proof}

\clearpage
\section{Experimental setting}\label{app:setup}
In the main experiment, we compared \ours{} with the baseline. The datasets and experimental parameters are as follows:
\subsection{Dataset}

In this section, we introduce the statistics of the dataset and the additional processing performed on the dataset. The statistics of the dataset are shown in Table~\ref{tab:Statistics For Data}. In addition, We added new templates to the original dataset to ensure the model could complete the required tasks and output formats. It is important to note that the added templates did not alter the original dataset, and special processing was performed for different LLMs. The specific examples are as follows:
\begin{tcolorbox}[title={\textbf{\small Dataset Format of Hellaswag}},
colback=whitesmoke, colframe=darkblue, boxrule=2pt, arc=0mm]
{\scriptsize
\begin{lstlisting}[style=mystyle]
dataset: Hellaswag
    dataset format:
    {
    "instruction": "{Article}\n 
    Question: {Question}\n  
    A. {Option A}\n
    B. {Option B}\n
    C. {Option C}\n
    D. {Option D}\n 
    You may choose from 'A', 'B', 'C', 'D'.\n Answer:",
    "output": "{Answer}"
    }
    example:
    {
    "instruction": "A man is sitting on a roof. He\n
    Question: Which ending makes the most sense?\n 
    A. is using wrap to wrap a pair of skis.\n
    B. is ripping level tiles off.\n
    C. is holding a Rubik's cube.\n
    D. starts pulling up roofing on a roof.\n 
    You may choose from 'A', 'B', 'C', 'D'.\n Answer:",
    "output": "D"
    }
\end{lstlisting}
}
\end{tcolorbox}
\begin{tcolorbox}[title={\textbf{\small Dataset Format of PIQA}},
colback=whitesmoke, colframe=darkblue, , boxrule=2pt, arc=0mm]
{\scriptsize
\begin{lstlisting}[style=mystyle]
dataset: PIQA
    dataset format:
    {
    "instruction": "There is a single choice question.
    Answer the question by replying A or B.'\n 
    Question: {Question}\n
    A. {Option A}\n
    B. {Option B}\n
    Answer:",
    "output": "{Answer}"
    }
    example:
    {
    "instruction": "There is a single choice question.
    Answer the question by replying A or B.'\n 
    Question: When boiling butter, when it's ready, you can\n
    A. Pour it onto a plate\n
    B. Pour it into a jar\n
    Answer:",
    "output": "B"
    }
\end{lstlisting}
}
\end{tcolorbox}
\begin{tcolorbox}[title={\textbf{\small Dataset Format of Winogrande}},
colback=whitesmoke, colframe=darkblue, boxrule=2pt, arc=0mm]
{\scriptsize
\begin{lstlisting}[style=mystyle]
dataset: Winogrande
    dataset format:
    {
    "instruction": "There is a single choice question, 
    you need to choose the correct option to fill in the blank. 
    Answer the question by replying A or B.'\n 
    Question: {Question}\n
    A. {Option A}\n
    B. {Option B}\n
    Answer:",
    "output": "{Answer}"
    }
    example:
    {
    "instruction": "There is a single choice question, 
    you need to choose the correct option to fill in the blank. 
    Answer the question by replying A or B.'\n 
    Question: Sarah was a much better surgeon than Maria so _ always got the 
    easier cases.\n
    A. Sarah\n
    B. Maria\n
    Answer:",
    "output": "B"
    }
\end{lstlisting}
}
\end{tcolorbox}
\begin{tcolorbox}[title={\textbf{\small Dataset Format of RACE}},
colback=whitesmoke, colframe=darkblue,  boxrule=2pt, arc=0mm]
{\scriptsize
\begin{lstlisting}[style=mystyle]
dataset: RACE
    dataset format:
    {
    "instruction": "{Article}  
    {Question}\n  
    [ {Option A}, {Option B}\, {Option C},  {Option D}]",
    "output": "{Answer}"
    }
    example:
    {
    "instruction": "I am a psychologist. I first met Timothy, a quiet, 
    overweight eleven-year-old boy, when his mother brought him to me to discuss 
    his declining grades. A few minutes with Timothy were enough to confirm that 
    his self-esteem  and general happiness were falling right along with _ . 
    I asked about Timothy's typical day. He awoke every morning at six thirty
    so he could reach his school by eight and arrived home around four thirty each
    afternoon. He then had a quick snack, followed by either a piano lesson
    or a lesson with his math tutor. He finished dinner at 7 pm, and then he sat
    down to do homework for two to three hours. Quickly doing the math in my
    head, I found that Timothy spent an average of thirteen hours a day
    at a writing desk.\n
    What if Timothy spent thirteen hours a day at a sewing machine instead of 
    a desk? We would immediately be shocked, because that would be called 
    children being horribly mistreated. Timothy was far from being mistreated,
    but the mountain of homework he faced daily resulted in a similar consequence
    --he was being robbed of his childhood. In fact, Timothy had no time
    to do anything he truly enjoyed, such as playing video games, watching
    movies, or playing board games with his friends.\n
    Play, however, is a crucial part of healthy child development.
    It affects children's creativity, their social skills, and even their brain
    development. The absence of play, physical exercise, and freefrom social
    interaction takes a serious toll on many children. It can also cause
    significant health problems like childhood obesity, sleep problems
    and depression.\nExperts in the field recommend the minutes children
    spend on their homework should be no more than ten times the number
    of their grade level./nWhat did the writer think of Timothy after
    learning about his typical day?/n
    ['Timothy was very hardworking.', 
    'Timothy was being mistreated.',
    'Timothy had a heavy burden.', 
    'Timothy was enjoying his childhood.']",
    "output": "C"
    }
\end{lstlisting}
}
\end{tcolorbox}

\begin{table*}[t]
\centering
\caption{Number of samples in the train, validation, and test datasets for various dateset.}
\label{tab:Statistics For Data}
\begin{tabular}{l*{4}{c}}
\toprule
\textbf{Number of samples} &\textbf{train dataset} & \textbf{validation dataset} & \textbf{test dataset} \\ 
\midrule
Hellaswag & 39900 & 10000 & 10000  \\
 PIQA & 16000 & 2000 & 3000  \\
Winogrande & 40398 & 1267 & 1767  \\
 RACE & 87866 & 4887 & 4934  \\
\bottomrule
\end{tabular}
\end{table*}
\subsection{Specific experimental parameters} 
Based on the Llama3-8B model configuration, several adjustments were made to optimize model performance. In the baseline model experiment, generation parameters were adjusted to ensure the correct output. In the LoRA experiment, adjustments to the generation parameters were retained, and LoRA-related parameters were adjusted. In the \ours{} experiment, the size of the validation set was adjusted to control the time required to search for the optimal layer. In the AdaLoRA experiment, the initial rank size was modified to ensure that the fine-tuning parameters are consistent with \ours{}. In the LoRA-drop experiment, the number of fine-tuning layers was set to be consistent with \ours{} to ensure that the fine-tuning parameters are consistent. In the LoRAShear experiment, the pruning ratio was modified, where the parameter amount with a pruning ratio of 50\% is consistent with \ours{}. For specific experimental parameters, see the table~\ref{tab:parameters}.

\begin{table*}[t]
\centering
\scriptsize
\caption{Detailed experimental parameters. This table lists the specific parameters we used in the experiments for various methods. These parameters include the target module of LoRA (Lora Target), the maximum sequence length (Max Length), the number of samples for supervised fine-tuning (SFT Samples), the learning rate (LR), the number of search samples (Search Samples), the initial rank (Init Rank), the target rank (Target Rank), and the ratio of pruning (Ratio). Epoch represents the epoch of training. In particular, the epochs of \ours{} in the \ours{} Layer Selection stage and the Fine-tuning stage are different. In the table, the former is the epoch of the Flexible Layer Selection stage and the latter is the epoch of the Fine-tuning stage. All other parameters not listed here remain consistent across all experiments. }
\label{tab:parameters}
\begin{tabular}{l*{9}{c}}
\toprule
\textbf{Methods} &\textbf{LoRA Target} & \textbf{Max Length} & \textbf{SFT Samples} & \textbf{LR} & \textbf{Search Samples}& \textbf{Init Rank} & \textbf{Target Rank}& \textbf{Ratio} & \textbf{Epoch} \\ 
\midrule
LoRA & q \& v Proj & 1024 & 20000 & 0.0001 & - & -& - & - & 3 \\
\midrule
\ours{} & q \& v Proj & 1024 & 20000 & 0.0001 & 20000 & -& - & - & 1/3\\
\midrule
AdaLoRA & q \& v Proj & 1024 & 20000 & 0.0001 & - & 4 & 8 & - & 3 \\
\midrule
LoRA-drop& q \& v Proj & 1024 & 20000 & 0.0001 & 20000 & -& -& - & 3 \\
\midrule
LoRAShear & q \& v Proj & 1024 & 20000 & 0.0001 & 20000 & -& - & 0.5 & 3 \\
\midrule
Dora & q \& v Proj & 1024 & 20000 & 0.0001 & 20000 & -& -& - & 3 \\
\midrule
rsLoRA & q \& v Proj & 1024 & 20000 & 0.0001 & 20000 & -& -& - & 3 \\
\midrule
LoRAPrune & q \& v Proj & 1024 & 20000 & 0.0001 & 20000 & -& -& 0.5 & 3 \\
\bottomrule
\end{tabular}
\end{table*}

\subsection{Other LLMs experimental parameters}
In order to explore the versatility and scalability of \ours{}, we conducted experiments on multiple different LLMs. The specific training parameters are shown in Table~\ref{tab:other_parameters}.

\begin{table*}[t]
\scriptsize
\centering
\caption{Detailed LLM experiment parameters. This table provides a comprehensive overview of the specific parameters used for different large language models (LLMs) in our experiments. These parameters include the LoRA alpha value (LoRA Alpha), the dropout rate of LoRA (LoRA Dropout), the rank used in LoRA (LoRA Rank), and the target module of LoRA (LoRA Target). In addition, the table lists the specific templates used for each LLM, which are derived from Llama-factory (Template). For experiments involving different downstream tasks using the same model, all other parameters are kept consistent to ensure fair comparison and best performance.}
\label{tab:other_parameters}
\begin{tabular}{l*{5}{c}}
\toprule
\textbf{LLM} &\textbf{LoRA Alpha} & \textbf{LoRA Dropout} & \textbf{LoRA Rank} & \textbf{LoRA Target} & \textbf{Tamplate (From Llama-factory)} \\ 
\midrule
Llama3 & 16 & 0 & 8 & q \& v Proj & llama3 \\
\midrule
Llama & 16 & 0 & 8 & q \& v Proj & defult \\
\midrule
Llama2 & 16 & 0 & 8 & q \& v Proj & llama2 \\
\midrule
chatglm3 & 16 & 0 & 8 & query\_key\_value & chatglm3 \\
\midrule
Mistral-v0.1 & 16 & 0 & 8 & q \& v Proj & mistral \\
\midrule
gemma & 16 & 0 & 8 & q \& v Proj & gemma \\
\midrule
zephyr & 16 & 0 & 8 & q \& v Proj & zephyr \\
\midrule
vicuna & 16 & 0 & 8 & q \& v Proj & vicuna \\
\midrule
xuanyuan & 16 & 0 & 8 & q \& v Proj & xuanyuan \\
\midrule
qwen1.5 & 16 & 0 & 8 & q \& v Proj & qwen \\
\midrule
yi & 16 & 0 & 8 & q \& v Proj & yi \\
\bottomrule
\end{tabular}
\end{table*}
\section{More results}\label{more_results}
\subsection{Computational Overhead of \ours{}}\label{time}

This section analyzes the computational overhead of \ours{}, focusing on the flexible layer selection stage and comparing the overall cost with LoRA.

\paragraph{Flexible Layer Selection Cost}

\ours{} operates in two phases: (1) flexible layer selection and (2) fine-tuning.  The layer selection phase identifies the optimal layer combination, and its computational cost scales with the number of samples used for the search.  Table~\ref{tab:different sample} shows the average search time increases with the number of samples, but remains manageable. For example, searching with 1,000 samples takes 0.08 hours, while searching with 10,000 samples takes 0.8 hours. This demonstrates the efficiency of the search process, even for larger datasets. It is important to note that the cost associated with the flexible layer selection phase is relatively low, especially when considering the significant improvements in accuracy it yields. While the search time does increase with the number of samples, the overall cost remains acceptable, particularly in tasks where accuracy is of paramount importance. In such scenarios, the benefits of identifying the optimal layer combination far outweigh the modest computational expense incurred during the search process. Moreover, the efficiency of this search process allows for exploration of a wider range of potential layer combinations, increasing the likelihood of discovering highly effective architectures that contribute to improved model performance and generalization. This efficient layer selection strategy is a crucial component of \ours{}, enabling it to achieve state-of-the-art results without incurring prohibitive computational costs.
\paragraph{Comparison with LoRA}

During fine-tuning, \ours{} significantly reduces both training time and the number of trainable parameters compared to LoRA, as shown in Table~\ref{tab:time_parameter_comparison}. \ours{} reduces training time by 4.0\% to 22.6\% and the number of trainable parameters by 41.2\% to 50.0\% across various datasets.  Importantly, the total computational cost of \ours{} (search plus fine-tuning) is comparable to LoRA's fine-tuning cost alone.  For instance, on Hellaswag, LoRA fine-tuning requires 5.30 hours, while \ours{} takes 4.71 + 0.08 = 4.79 hours.  On Winogrande, LoRA requires 4.96 hours, and \ours{} takes 3.84 + 0.08 = 3.92 hours.  This shows \ours{} doesn't introduce significant additional overhead compared to LoRA, while achieving better performance and efficiency.

\paragraph{Resource-Constrained Scenarios}

\ours{}'s efficiency is particularly beneficial in resource-constrained settings. The layer search can use a small number of samples (e.g., 1,000), requiring minimal resources (e.g., 0.08 hours). The reduction in trainable parameters and training time further makes \ours{} suitable for deployment in resource-limited environments.  In conclusion, \ours{} exhibits minimal computational overhead, comparable to LoRA, while substantially improving efficiency and performance, making it a practical approach for fine-tuning large language models. 

\begin{table}[t]
\centering
\caption{Comparison of training time and parameters, with the green font indicating the reduction ratio, is conducted on a single NVIDIA A100 GPU using Llama3-8B. The time metric reflects the wallclock time for the fine-tuning phase of LoRA and \ours{}, excluding the layer selection phase.
}
\label{tab:time_parameter_comparison}
\begin{tabular}{lccccc}
\toprule
\textbf{Metrics} &\textbf{Method} & \textbf{Hellaswag} & \textbf{PIQA} & \textbf{Winogrande} & \textbf{RACE} \\ 
\midrule
\multirow{2}{*}{Time (h)} & LoRA & 5.30 & 4.03 & 4.96 & 8.37\\
& \ours{} & \textbf{4.71 \textcolor{green(pigment)}{(11.1\%)}} & \textbf{3.87 \textcolor{green(pigment)}{(4.0\%)}} & \textbf{3.84 \textcolor{green(pigment)}{(22.6\%)}} & \textbf{7.46 \textcolor{green(pigment)}{(10.9\%)}} \\
\midrule
\multirow{2}{*}{\# Params (M)} & LoRA & 3.4 & 3.4 & 3.4 & 3.4 \\
& \ours{} & \textbf{2.00 \textcolor{green(pigment)}{(41.2\%)}} & \textbf{1.70 \textcolor{green(pigment)}{(50.0\%)}} & \textbf{1.70 \textcolor{green(pigment)}{(50.0\%)}} & \textbf{1.70 \textcolor{green(pigment)}{(50.0\%)}} \\
\bottomrule
\end{tabular}
\vspace{-3mm}
\end{table}

\subsection{The results of other LLMs experiment}\label{Other_LLMs}
\paragraph{Wide Applicability of \ours{}.} According to the parameter settings in Table~\ref{tab:other_parameters}, the verification results for various LLMs are presented in Table~\ref{tab:llms_accuracy_comparison}. The selected LLMs include Llama3-8B, Llama-7B, Llama2-7B, ChatGLM3-6B, Mistral-7B-v0.1, Gemma-7B, Zephyr-7B-beta, Vicuna-7B-v1.5, XuanYuan-6B, Qwen1.5-7B, and Yi-6B. These models demonstrate unique characteristics in terms of training data, architecture design, and optimized training. First, the models utilize varied training data, leading to differences in data distribution. Additionally, some models have enhanced attention mechanisms: Mistral-7B-v0.1 employs grouped query attention (GQA) and sliding window attention (SWA), while ChatGLM3-6B features a special attention design to support tool calling and code execution capabilities. Activation functions vary across these models. Llama3-8B uses the SwiGLU activation function, inspired by the PaLM model, to improve performance and convergence speed, while ChatGLM3-6B uses the Swish activation function. Furthermore, differences in reasoning optimization and multilingual capabilities contribute to varied reasoning abilities across fields. The experimental result of each model is shown in Table~\ref{tab:llms_accuracy_comparison}, which presents the scores of each model on different downstream tasks after LoRA and \ours{} fine-tuning. It should be noted that all models fine-tuned using LoRA will have a certain degree of overfitting, while \ours{} can effectively identify and analyze unnecessary layers in specific downstream tasks and prune them to reduce model overfitting. After optimization by \ours{}, these LLMs showed significant performance improvements on downstream tasks. In particular, models that originally performed poorly on some tasks, such as ChatGLM3-6B, experienced significant improvements, achieving more than a 15\% increase on the RACE-mid and RACE-high tasks. This improvement is attributable to the key layer selection by \ours{} and efficient model learning. In summary, \ours{} is applicable across Transformer models of various structures, excels in diverse tasks, and effectively enhances areas where model capabilities are lacking.
\begin{table*}
    \centering
    \footnotesize
    \caption{Detailed comparison of the accuracy of different LLMs. This table presents a comprehensive comparison of the accuracy results obtained by fine-tuning various mainstream Large Language Models (LLMs) using \ours{} and LoRA methods. The accuracy metrics are reported across multiple benchmark datasets, including HellaSwag, PIQA, Winogrande, RACE-mid, and RACE-high. The average accuracy across all datasets is also provided. The exact values of accuracy improvements for each method, highlighted in red, indicate the performance gains achieved. }
    \label{tab:llms_accuracy_comparison}
    \begin{tabularx}{\textwidth}{l*{6}{X}}
        \toprule
        \textbf{Methods} & \textbf{Hellaswag} & \textbf{PIQA} & \textbf{Winogrande} & \textbf{RACE-mid} & \textbf{RACE-high} & \textbf{Average} \\
        \midrule
        Llama3-8B-LoRA & 89.72 & 76.39 & 82.24 & 82.86 & 80.99 & 83.04 \\
        Llama3-8B-\ours{} & 93.62 \textcolor{red}{(+3.90)} & 85.91 \textcolor{red}{(+9.52)} & 85.79 \textcolor{red}{(+3.55)} & 84.61 \textcolor{red}{(+1.75)} & 82.36 \textcolor{red}{(+1.37)} & 86.46 \textcolor{red}{(+3.42)} \\

        \midrule
        Llama-7B-LoRA & 76.10 & 69.80 & 67.01 & 75.69 & 70.81 & 71.88 \\
        Llama-7B-\ours{} & 85.28 \textcolor{red}{(+9.18)} & 71.93 \textcolor{red}{(+2.13)} & 74.11 \textcolor{red}{(+7.10)} & 81.62 \textcolor{red}{(+5.93)} & 78.62 \textcolor{red}{(+7.81)} & 78.31 \textcolor{red}{(+6.43)} \\
        \midrule
        Llama2-7B-LoRA & 79.60 & 75.90 & 78.60 & 79.32 & 75.07 & 77.70 \\
        Llama2-7B-\ours{} & 90.89 \textcolor{red}{(+11.29)} & 81.72 \textcolor{red}{(+5.82)} & 82.85 \textcolor{red}{(+4.25)} & 84.89 \textcolor{red}{(+5.57)} & 83.19 \textcolor{red}{(+8.12)} & 84.71 \textcolor{red}{(+7.01)}\\
        \midrule
        Chatglm3-6B-LoRA & 83.02 & 70.62 & 69.93 & 63.43 & 59.46 & 69.29 \\
        Chatglm3-6B-\ours{} & 85.12 \textcolor{red}{(+2.10)} & 74.81 \textcolor{red}{(+4.19)} & 72.69 \textcolor{red}{(+2.76)} & 79.18 \textcolor{red}{(+15.75)} & 76.33 \textcolor{red}{(+16.87)} & 77.63 \textcolor{red}{(+8.33)} \\
        \midrule
        Mistral-7B-v0.1-LoRA & 94.35 & 82.15 & 84.85 & 83.79 & 82.39 & 85.51 \\
        Mistral-7B-v0.1-\ours{} & 95.08 \textcolor{red}{(+0.73)} & 86.89 \textcolor{red}{(+4.74)} & 85.50 \textcolor{red}{(+0.65)} & 85.72 \textcolor{red}{(+1.93)} & 84.25 \textcolor{red}{(+1.86)} & 87.49 \textcolor{red}{(+1.98)} \\
        \midrule
        Gemma-7B-LoRA & 94.85 & 83.19 & 80.19 & 85.73 & 83.96 & 85.58 \\
        Gemma-7B-\ours{} & 95.76 \textcolor{red}{(+0.91)} & 87.54 \textcolor{red}{(+4.35)} & 83.58 \textcolor{red}{(+3.39)} & 89.62 \textcolor{red}{(+3.89)} & 88.19 \textcolor{red}{(+4.23)} & 88.94 \textcolor{red}{(+3.35)} \\
        \midrule
        Zephyr-7B-beta-LoRA & 93.77 & 75.03 & 78.37 & 83.45 & 82.25 & 82.57 \\
        Zephyr-7B-beta-\ours{} & 95.05 \textcolor{red}{(+1.28)} & 85.58 \textcolor{red}{(+10.55)} & 84.95 \textcolor{red}{(+6.58)} & 86.19 \textcolor{red}{(+2.74)} & 84.30 \textcolor{red}{(+2.05)} & 87.21 \textcolor{red}{(+4.64)} \\
        \midrule
        Vicuna-7B-v1.5-LoRA & 87.64 & 69.48 & 63.85 & 67.30 & 73.90 & 72.43 \\
        Vicuna-7B-v1.5-\ours{} & 90.43 \textcolor{red}{(+2.79)} & 79.49 \textcolor{red}{(+10.01)} & 76.06 \textcolor{red}{(+12.21)} & 82.94 \textcolor{red}{(+15.64)} & 81.90 \textcolor{red}{(+8.00)} & 82.16 \textcolor{red}{(+9.73)} \\
        \midrule
        XuanYuan-6B-LoRA & 82.38 & 74.16 & 65.27 & 78.04 & 72.11 & 74.39 \\
        XuanYuan-6B-\ours{} & 88.41 \textcolor{red}{(+6.03)} & 79.43 \textcolor{red}{(+5.27)} & 73.40 \textcolor{red}{(+8.13)} & 84.89 \textcolor{red}{(+6.85)} & 80.70 \textcolor{red}{(+8.59)} & 81.37 \textcolor{red}{(+6.97)} \\
        \midrule
        Qwen1.5-7B-LoRA & 91.75 & 75.03 & 78.14 & 87.59 & 81.36 & 82.77 \\
        Qwen1.5-7B-\ours{} & 91.96 \textcolor{red}{(+0.21)} & 84.33 \textcolor{red}{(+9.30)} & 80.69 \textcolor{red}{(+2.55)} & 89.90 \textcolor{red}{(+2.31)} & 87.08 \textcolor{red}{(+5.72)} & 86.79 \textcolor{red}{(+4.02)} \\
        \midrule
        Yi-6B-LoRA & 89.46 & 78.29 & 76.01 & 80.02 & 85.13 & 81.78 \\
        Yi-6B-\ours{} & 92.24 \textcolor{red}{(+2.78)} & 84.82 \textcolor{red}{(+6.53)} & 84.96 \textcolor{red}{(+8.95)} & 88.72 \textcolor{red}{(+8.70)} & 86.91 \textcolor{red}{(+1.78)} & 87.53 \textcolor{red}{(+5.75)} \\
        \midrule
        Qwen2.5-32B-LoRA & 90.35 & 78.94 & 84.59 & 84.17 & 81.03 & 83.82 \\
        Qwen2.5-32B-\ours{} & 94.01 \textcolor{red}{(+3.66)} & 86.72 \textcolor{red}{(+7.78)} & 87.36 \textcolor{red}{(+2.77)} & 87.16 \textcolor{red}{(+2.99)} & 84.27 \textcolor{red}{(+3.24)} & 87.90 \textcolor{red}{(+4.08)} \\
        \bottomrule
    \end{tabularx}
\end{table*}
\subsection{The results of other LoRAs experiment } \label{scalability}
\paragraph{Strong Scalability of \ours{}.} Recently, as highlighted in the introduction, many LoRA improvement methods have been proposed and have achieved excellent performance in specific fine-tuning tasks. In this section, we explore the potential of combining our algorithm with other emerging LoRA algorithms. Four promising LoRA variants are selected from different methods, each demonstrating impressive performance. Specifically, DoRA (Decomposed Low Rank Adaptation by Weight) \cite{c:43} achieves low-rank adaptation through weight decomposition, and rsLoRA (Rank-Stabilized LoRA) \cite{kalajdzievski2023rankstabilizationscalingfactor} addresses the slow training speed of traditional LoRA by introducing a rank-stable scaling factor when increasing the rank. These methods primarily solve the parameter overfitting problem within the LoRA parameters but overlook the overall overfitting problem. By innovatively combining these methods with our algorithm, we first address the overall overfitting problem and then tackle the overfitting issue of the remaining LoRA parameters, thereby significantly improving performance. Additionally, we attempt to integrate with other methods to enhance the representation ability of LoRA. For instance, MoSLoRA (Mixture-of-Subspaces in Low-Rank Adaptation) \cite{wu2024mixtureofsubspaceslowrankadaptation}decomposes LoRA into subspaces via structural re-parameterization, employing a learnable mixer to fuse more subspaces more flexibly. LoReFT (Low-rank Linear Subspace ReFT) \cite{wu2024reftrepresentationfinetuninglanguage}is a parameter-efficient finetuning method that operates on a frozen base model, learning task-specific interventions on hidden representations. The specific experimental results are shown in Table~\ref{tab:different LoRA}. The results indicate that \ours{} can be effectively integrated with DoRA and rsLoRA, alleviating the overfitting problem of LLM and improving performance with less than half of the parameters. Notably, the integration of \ours{} and LoReFT can further enhance performance. \ours{} helps LoReFT identify the most suitable layer for fine-tuning, avoiding performance loss caused by manually selecting the fine-tuning layer. However, MoSLoRA is not suitable for integration with \ours{} because MoSLoRA combines the A and B matrices of all LoRA layers. Deleting a layer would cause significant changes and degrade performance. The specific implementation requires replacing LoRA with DoRA, rsLoRA, MoSLoRA, or LoReFT for inner layer optimization during the flexible layer selection stage, while the outer layer optimization remains unchanged. These adjustments can be achieved through direct modification. The results demonstrate that \ours{} exhibits strong scalability when combined with algorithms for enhancing LoRA parameters, highlighting its great potential.

\begin{table*}
    \footnotesize
    \centering
    \caption{Detailed comparison of the accuracy of the combination of \ours{} and different LoRA algorithms on Llama3-8B. This table presents a detailed comparison of the accuracy results obtained by integrating \ours{} with various improved LoRA algorithms, including DoRA, rsLoRA, MoSLoRA and LoReFT, while maintaining other experimental settings constant. The accuracy metrics are reported across multiple benchmark datasets, including HellaSwag, PIQA, Winogrande, RACE-mid, and RACE-high, with the average accuracy across all datasets also provided. The results are compared against those obtained from direct fine-tuning without \ours{}. The experimental findings indicate that the application of \ours{} can significantly reduce model overfitting and enhance overall performance.
}
    \label{tab:different LoRA}
    \begin{tabular}{lcccccc}
        \toprule
        \textbf{Methods} & \textbf{Hellaswag} & \textbf{PIQA} & \textbf{Winogrande} & \textbf{RACE-mid} & \textbf{RACE-high} & \textbf{Average} \\
        \midrule
        LoRA & 89.72 & 76.39 & 82.24 & 85.86 & 80.99 & 83.04 \\
        \ours{} (w/ LoRA) & \textbf{93.62} & \textbf{85.91} & \textbf{85.79} & \textbf{84.61} & \textbf{82.36} & \textbf{86.46}\\
        \midrule
        rsLoRA & 94.33 & 87.21 & 85.32 & 87.60 & 84.36 & 87.76 \\
        \ours{} (w/ rsLoRA) & \textbf{94.83} & \textbf{87.58} & \textbf{86.69} & \textbf{88.21} & \textbf{85.46} & \textbf{88.55} \\
        \midrule
        DoRA & 93.62 & 85.75 & 84.77 & 86.77 & 83.39 & 86.86 \\
        \ours{} (w/ DoRA) & \textbf{94.10} & \textbf{86.05} & \textbf{86.32} & \textbf{87.12} & \textbf{84.45} & \textbf{87.61} \\
        \midrule
        LoReFT & 96.31 & 90.24 & \textbf{87.48} & 88.21 & \textbf{85.33} & 89.51 \\
        \ours{} (w/ LoReFT) & \textbf{96.47} & \textbf{91.06} & 87.23 & \textbf{88.36} & 84.97 & \textbf{89.62} \\
        \midrule
        MoSLoRA & 93.53 & 85.97 & 84.26 & \textbf{86.13} & \textbf{83.75} & \textbf{86.73} \\
        \ours{} (w/ MoSLoRA) & \textbf{93.76} & \textbf{86.43} & \textbf{85.36} & 85.09 & 82.07 & 86.54 \\
        \bottomrule
    \end{tabular}
\end{table*}

\subsection{Comparison with LoRAShear} \label{Comparison_lorashear}
\paragraph{Better Performance of \ours{}.}In this section, the accuracy of \ours{} is compared with that of LoRAShear across various datasets, with specific results presented in Table~\ref{tab:llama_accuracy_comparison}. Since LoRAShear is not open source and poses challenges for direct experimentation, the comparison relies on the experimental configurations and results reported in the LoRAShear paper. Notably, \ours{} can freely adjust the selected layers according to the dataset, achieving an average pruning parameter rate of 50\%. Consequently, under the same pruning rate, \ours{} outperforms by 14\% (Ratio = 0.5). Experiments have shown that under the same pruning rate, \ours{} can achieve better performance. Note that during the test, BoolQ only has training and test sets. We still keep the test set unchanged for testing the model, use 80\% of the training set data to train LoRA parameters, and use the other 20\% of the data to train the hyperparameters introduced by \ours{}. In addition, the reason for the poor performance on the ARC and OBQA datasets is that the number of validation sets is small, and the layer selection may not be accurate enough. For a discussion on the number of validation sets and the accuracy, see section~\ref{search sample}.

\begin{table*}
\centering
\scriptsize
\caption{Detailed comparison of commonsense reasoning task accuracy. This table provides a comprehensive comparison of the accuracy results for various methods applied to common sense reasoning tasks, conducted on the Llama-7B model. The methods compared include the pre-trained model, LoRA, LoRAShear with different pruning ratios (0.5), and \ours{}. The accuracy metrics are reported across multiple benchmark datasets, including BoolQ, PIQA, HellaSwag, Winogrande, ARC-e, ARC-c, and OBQA. The average accuracy across all datasets is also provided. The ``Ratio" column represents the ratio of parameter pruning in LoRAShear.}
\label{tab:llama_accuracy_comparison}
    \begin{tabular}{lccccccccc}
        \toprule
        \textbf{Methods} & \textbf{BoolQ} & \textbf{PIQA} & \textbf{HellaSwag} & \textbf{WinoGrande}& \textbf{ARC-e} & \textbf{ARC-c}&\textbf{OBQA}&\textbf{Average} \\
        \midrule
        Pre-trained & 57.98 & 60.94 & 34.35 & 52.25 & 31.82 & 27.30 & 35.80 & 42.92 \\
        LoRA & 67.76 & 69.80 & 76.10 & 67.01 & 67.21  & 35.23  & 38.60 & 60.24 \\
        LoRAShear (Ratio = 0.5)& 63.40 & \textbf{72.15} & 49.83 & 56.40 & 49.45  & 34.31  & 35.86 & 51.63 \\
        \textbf{\ours{}} & \textbf{73.54} & 71.93 & \textbf{85.28} & \textbf{74.11} & \textbf{71.22} & \textbf{45.64} & \textbf{39.86} & \textbf{65.94} \\
        \bottomrule
    \end{tabular}
\end{table*}
\subsection{\ours{} in full-parameter fine-tuning} \label{full-parameter}
\paragraph{\ours{} can improve the performance of full parameter fine-tuning.} In this section, we evaluate the performance of \ours{} in the context of full-parameter fine-tuning. Specifically, while maintaining the inner loop optimization steps unchanged, we modify the trainable parameters in the outer loop from the LoRA adapter to the full set of parameters of the LLM itself. The experimental results, presented in Table~\ref{tab:full-parameter}, demonstrate that although \ours{} enhances performance in full-parameter fine-tuning compared to baseline methods, it still falls short of the performance achieved by LoRA + \ours{}. This suggests that full-parameter fine-tuning, which involves adjusting all layers of the model, is more susceptible to overfitting, even when employing the layer selection mechanism of \ours{}. These findings underscore the significance of parameter-efficient approaches like LoRA, particularly in scenarios where overfitting is a critical concern, such as in customization or personalization tasks with limited disk storage. The capability of \ours{} to identify and prioritize critical layers proves especially advantageous in these contexts, offering a balanced trade-off between model adaptability and resource efficiency.

\begin{table*}
    \centering
    \footnotesize
    \caption{Performance comparison of Full Fine-Tuning (Full FT), LoRA, and their Flexora-enhanced variants across multiple reasoning and reading comprehension tasks, including Hellaswag, PIQA, Winogrande, RACE-mid, and RACE-high. Results are reported as accuracy, with improvements over baseline methods indicated in parentheses. Flexora significantly enhances both Full FT and LoRA, with the largest gains observed for LoRA, achieving an average accuracy of 86.46\% (+3.42\% over LoRA).}
    \label{tab:full-parameter}
    \begin{tabular}{lcccccc}
        \toprule
        \textbf{Method}& \textbf{Hellaswag} & \textbf{PIQA} & \textbf{Winogrande} & \textbf{RACE-mid} & \textbf{RACE-high} & \textbf{Average}\\
        \midrule
        Full FT & 90.53 & 79.32 & 81.16 & 81.92 & 79.36 &  82.46  \\
        Flexora(w/ Full FT) & 91.32\textcolor{red}{(+0.79)} & 83.21\textcolor{red}{(+3.89)} & 81.73\textcolor{red}{(+0.57)} & 83.13\textcolor{red}{(+1.21)} & 80.37\textcolor{red}{(+1.01)}   & 83.95\textcolor{red}{(+1.49)}\\
        LoRA & 89.72 & 76.39 & 82.24 & 82.86 & 80.99 & 83.04 \\
        Flexora(w/ LoRA) & \textbf{93.62} \textcolor{red}{(+3.90)} & \textbf{85.91} \textcolor{red}{(+9.52)} & \textbf{85.79} \textcolor{red}{(+3.55)} & \textbf{84.61} \textcolor{red}{(+1.75)} & \textbf{82.36} \textcolor{red}{(+1.37)} & \textbf{86.46} \textcolor{red}{(+3.42)} \\
        \bottomrule
    \end{tabular}
\end{table*}
\subsection{\ours{} in instruct model} \label{instruction}
\paragraph{ Effectiveness of \ours{} on instruct models.} To demonstrate the effectiveness of \ours{} on the instruct model, we conducted experiments on Meta-Llama-3-8B-Instruct. The experimental results are shown in Table~\ref{tab:instruction}, which show that \ours{} always maintains excellent performance on both the base model and the instruct model. This confirms that \ours{} is effective in different fine-tuning scenarios (including instruct model adaptation).

\begin{table*}
    \centering
    \scriptsize
    \caption{Comparison of \ours{} with base models and instruction-tuned models on various datasets.  Numbers in parentheses indicate the relative improvement over the corresponding base model or instruction-tuned model.  \ours{} demonstrates consistent improvements across all datasets and settings.}
    \label{tab:instruction}
    \begin{tabular}{lcccccc}
        \toprule
        \textbf{Method}& \textbf{Hellaswag} & \textbf{PIQA} & \textbf{Winogrande} & \textbf{RACE-mid} & \textbf{RACE-high} & \textbf{Average}\\
        \midrule
        Base Model & 48.55 & 67.08 & 59.91 & 67.02 & 63.35 & 61.18  \\
        Flexora(w/ Base Model) & \textbf{93.62} \textcolor{red}{(+45.07)} & \textbf{85.91} \textcolor{red}{(+18.83)} & \textbf{85.79} \textcolor{red}{(+25.88)} & \textbf{84.61} \textcolor{red}{(+17.59)} & \textbf{82.36} \textcolor{red}{(+19.01)} & \textbf{86.46} \textcolor{red}{(+25.28)}\\
        Instruct Model & 89.38 & 80.36 & 81.35 & 81.36 & 79.48 & 82.39 \\
        Flexora(w/ Instruct Model) & 93.53 \textcolor{red}{(+4.15)} & 85.76 \textcolor{red}{(+5.4)} & 85.67 \textcolor{red}{(+4.32)} & 84.58 \textcolor{red}{(+3.22)} & 82.19 \textcolor{red}{(+2.71)} & 86.35 \textcolor{red}{(+3.96)} \\
        \bottomrule
    \end{tabular}
\end{table*}
\subsection{Different search sample} 
\label{search sample}
\paragraph{Flexibility of \ours{} in search sample .}In \ours{}, search time is managed by adjusting the maximum number of search samples (corresponding to the size of the validation dataset) to align with the requirements of the downstream task. In Table~\ref{tab:different sample}, we explore the relationship between different numbers of search samples, downstream task performance, and search time. For simpler datasets like Hellaswag and PIQA, a 10-minute search with 1,000 samples significantly improves performance. For more challenging tasks, at least 1 hour of search time is required for 5,000 samples. In more difficult tasks, using too few samples can prevent validation loss from converging. To optimize performance, it is recommended to dynamically adjust the number of search samples based on the convergence of the validation loss. In summary, for simpler downstream tasks, \ours{} can be rapidly applied to reduce model overfitting significantly and enhance performance. For more challenging downstream tasks, \ours{} balances performance and training resources by adjusting the number of search samples.

\begin{table*}
    \centering
    \footnotesize
    \caption{Detailed analysis of the impact of different numbers of search samples on the \ours{} accuracy of Llama3-8B. This table investigates how varying the number of search samples, i.e., different validation dataset sizes, affects the performance of \ours{}. The accuracy metrics are reported across multiple benchmark datasets, including HellaSwag, PIQA, Winogrande, RACE-mid, and RACE-high, with the average accuracy across all datasets also provided. The number of search samples tested includes 1000, 2000, 5000, 10000, and 200000. All experimental conditions remain unchanged except for the size of the validation set, allowing for a focused analysis on the impact of search sample size on model performance.}
    \label{tab:different sample}
    \begin{tabular}{lcccccc}
        \toprule
        \textbf{\# Samples}& \textbf{Hellaswag} & \textbf{PIQA} & \textbf{Winogrande} & \textbf{RACE-mid} & \textbf{RACE-high} & \textbf{Average Time(h)}\\
        \midrule
        1000 & 93.00 & 80.52 & 83.04 & 76.74 & 72.93 &  0.08  \\
        1267 & - & - & \textbf{85.79} & - & -   & 0.12\\
        2000 & 92.29 & \textbf{85.91} & - & 80.15 & 78.82 & 0.17 \\
        4887 & - & - & - & \textbf{84.82} & \textbf{82.36} & 0.4 \\
        5000 & 93.17 & - & - & - & - & 0.42 \\
        10000 & \textbf{93.62} & - & - & - & -   & 0.8\\
        \bottomrule
    \end{tabular}
\end{table*}

\subsection{Ablation experiments on training settings} 
\label{sec:hyperparameters}
\paragraph{The choice of \( K \) and \( T \) is not important} The \( K \) and \( T \) parameters of \ours{} are inherited from the UD algorithm \cite{c:69}. As discussed in Section~\ref{related work}, some applications of the UD algorithm are highly sensitive to the choice of \( K \) and \( T \)\citep{c:81}. Therefore, we conducted ablation experiments to determine the optimal values for \( K \) and \( T \). The specific experimental results are presented in Table~\ref{tab:hyperparameters}. The results demonstrate that \ours{} is highly robust to variations in \( K \) and \( T \). This robustness may be attributed to the significant variability in the contribution of LLM layers to downstream tasks. For a detailed discussion, see Section~\ref{layer_analyis}. Regardless of the settings for \( K \) and \( T \), \ours{} consistently identifies the layers that contribute the most to downstream tasks.
\begin{table*}
    \centering
    \footnotesize
    \caption{Performance comparison of different hyperparameter settings \( K \) and \( T \) on various datasets. The rows represent different combinations of hyperparameters \( K \) and \( T \). The columns represent the accuracy results on different datasets: HellaSwag, PIQA, Winogrande, RACE-mid, and RACE-high. The last column shows the average accuracy across all datasets.}
    \label{tab:hyperparameters}
    \begin{tabular}{lcccccc}
        \toprule
        \textbf{\# \( K \) and \( T \)}& \textbf{Hellaswag} & \textbf{PIQA} & \textbf{Winogrande} & \textbf{RACE-mid} & \textbf{RACE-high} & \textbf{Average}\\
        \midrule
        \( K \) = 4, \( T \) = 1 & 93.57 & 85.76 & 85.72 & 84.62 & \textbf{82.46} & 86.43  \\
        \( K \) = 8, \( T \) = 1 & 93.62 & 85.91 & \textbf{85.79} & 84.61 & 82.36 & 86.46 \\
        \( K \) = 4, \( T \) = 2 & 93.78 & 85.37 & 84.16 & 83.96 & 83.17 & 86.09 \\
        \( K \) = 8, \( T \) = 2 & 93.07 & 85.16 & 85.01 & 84.57 & 82.06 & 85.97  \\
        \( K \) = 4, \( T \) = 4 & 92.97 & 85.72 & 85.56 & \textbf{85.07} & 82.11 & 86.29  \\
        \( K \) = 8, \( T \) = 4 & \textbf{93.89} & \textbf{86.01} & \textbf{85.79} & 84.99 & \textbf{82.46} & \textbf{86.63}  \\
        \bottomrule
    \end{tabular}
\end{table*}
\subsection{More results for preliminary study} 
\label{sec:random layer rank}
This section provides additional experimental results that are not shown in Section~\ref{sec:emp-insights}. In these experiments, we kept the randomly selected layers unchanged and only varied the LoRA rank. The specific experimental results are shown in Table~\ref{tab:rank}. The results indicate that regardless of the selected rank, the model's performance improves with an increasing number of LoRA fine-tuned layers up to a certain threshold. Beyond this threshold, further increasing the number of fine-tuned layers may lead to a decline in model performance. This intriguing phenomenon motivates our research.
\begin{table*}[bh]
\centering
\tiny
\caption{Performance of the model on various datasets (Hellaswag, PIQA, Winogrande, RACE-mid, RACE-high) under different LoRA ranks and varying numbers of LoRA fine-tuned layers.}
\label{tab:rank}
\begin{tabular*}{\textwidth}{@{\extracolsep{\fill}} lccccccc}
\toprule
\textbf{Rank} & \textbf{Layers} & \textbf{Hellaswag} & \textbf{PIQA} & \textbf{Winogrande} & \textbf{RACE-mid} & \textbf{RACE-high} & \textbf{Average}  \\ 
\midrule
\multirow{5}{*}{$r=4$} 
& 6 layers & 58.36 & 68.23 & 45.71 & 53.35 & 52.99 & 55.73 \\
& 12 layers & 78.23 & 76.53 & 54.78 & 79.04 & 54.99 & 68.71 \\
& 18 layers & \textbf{89.01} & \textbf{80.57} & 82.79 & 82.37 & \textbf{80.96} & \textbf{83.14} \\
& 24 layers & 88.21 & 79.36 & \textbf{82.97} & \textbf{82.39} & 80.12 & 82.61 \\
& 32 layers & 87.68 & 74.36 & 81.74 & 81.10 & 79.63 & 80.90 \\
\midrule
\multirow{5}{*}{$r=8$} 
& 6 layers & 59.79 & 70.25 & 46.32 & 54.54 & 53.45 & 56.87 \\
& 12 layers & 81.9 & 77.82 & 57.35 & 78.41 & 72.16 & 73.53 \\
& 18 layers & \textbf{91.15} & \textbf{81.54} & \textbf{83.58} & \textbf{83.77} & \textbf{81.22} & \textbf{84.25} \\
& 24 layers & 90.58 & 80.9 & 82.16 & 82.19 & 79.22 & 83.01 \\
& 32 layers & 89.72 & 76.39 & 82.24 & 82.86 & 80.99 & 82.44 \\
\midrule
\multirow{5}{*}{$r=16$} 
& 6 layers & 60.98 & 71.36 & 47.12 & 55.78 & 54.26 & 57.90 \\
& 12 layers & 80.23 & 78.01 & 62.69 & 79.55 & 75.62 & 75.22 \\
& 18 layers & \textbf{91.63} & \textbf{81.69} & \textbf{85.06} & \textbf{84.27} & \textbf{83.69} & \textbf{85.27} \\
& 24 layers & 90.11 & 79.60 & 83.57 & 82.13 & 78.39 & 82.76 \\
& 32 layers & 89.99 & 78.47 & 82.77 & 81.63 & 79.68 & 82.51 \\
\midrule
\multirow{5}{*}{$r=32$} 
& 6 layers & 60.45 & 71.46 & 50.36 & 57.36 & 55.13 & 58.95\\
& 12 layers & 82.4 & 79.07 & 63.17 & 80.13 & 78.63 & 76.68 \\
& 18 layers & \textbf{92.08} & \textbf{82.14} & \textbf{86.07} & 85.35 & 83.04 & \textbf{85.74} \\
& 24 layers & 91.55 & 81.37 & 85.13 & \textbf{85.75} & \textbf{83.17} & 82.76 \\
& 32 layers & 90.01 & 79.56 & 84.36 & 82.36 & 80.99 & 83.46 \\
\bottomrule
\end{tabular*}
\end{table*}

\subsection{Selection of layers} \label{layer_analyis}
For different LLMs and datasets, the layers chosen by \ours{} vary due to the different parameters learned in the pre-training stage and the diversity of downstream tasks. In Table~\ref{tab:layer selection1}, Table~\ref{tab:layer selection2}, Table~\ref{tab:layer selection3}, Table~\ref{tab:layer selection4}, and Table~\ref{tab:layer selection5}, we show the layers chosen by \ours{} in all experiments and the corresponding training parameters. In this section, the preferences of the layers chosen by \ours{} are analyzed in detail, providing layer-wise insights for LLMs.

\paragraph{The Effectiveness of \ours{} Comes from Reducing Overfitting.}In Table~\ref{tab:layer selection1}, the layers and parameter amounts selected by different LoRA methods are presented. A comparison between LoRA-drop and \ours{} reveals that \ours{} is more effective. LoRA-drop tends to select the later layers, as these outputs exhibit a larger two-norm, aligning with Proposition~\ref{proposition2}. This result suggests that layers selected during fine-tuning should not concentrate in a specific range but rather be distributed across various ranges, fully utilizing the extensive knowledge system of LLMs. Comparing LoRA with DoRA and rsLoRA shows that LoRA selects more layers, requiring more training parameters but yielding worse performance. This suggests a higher degree of overfitting when \ours{} is applied to LoRA compared to the other two methods. Therefore, using more advanced LoRA improvement algorithms can significantly reduce overfitting and enhance performance, underscoring the importance of the fine-tuning approach. Interestingly, certain layers are consistently fine-tuned in the same downstream task, regardless of whether LoRA, DoRA, or rsLoRA is used. For example, in Hellaswag, layers [0, 1, 2, 4, 14, 15, 19, 20, 21, 23, 26, 27, 28, 29, 31] are consistently selected, suggesting these layers are crucial for this task or represent general knowledge layers (see the next two paragraphs for details), closely related to the LLM itself .

\paragraph{General Knowledge Layers.}In Table~\ref{tab:layer selection2}, the layers and parameters selected in the second ablation study are shown. Observing the "Select first 6 layers by \ours{}" row reveals that certain layers, such as [27, 28], are crucial for any downstream task. These layers may store general knowledge, suggesting that their fine-tuning could enhance the performance across most downstream tasks.

\paragraph{Downstream task-specific layers.} Table~\ref{tab:layer selection3} displays the layers and parameter amounts selected by various LLMs for different downstream tasks. As evident from the table, the same model utilizes the aforementioned general knowledge layers across different tasks. Additionally, unique layers for each downstream task, termed downstream task-specific layers, are predominantly found in the first and last layers. The distinction between general knowledge layers and downstream task-specific layers can be attributed to the self-attention mechanism, which effectively differentiates these layers. In the self-attention mechanism, similar knowledge is aggregated, leading to this layer differentiation. Furthermore, concerning downstream task-specific layers, two conclusions are drawn: (a) Fewer layers are selected for simpler datasets to minimize overfitting. (b) Typically, the initial and final layers are selected for a given dataset. This selection pattern may stem from the initial layer processing the original input and the final layer generating the model's output representation. Given the consistent and predefined input and output, learning these parameters is deemed effective.

\paragraph{Poor Effects with No Critical Layers} Tables~\ref{tab:layer selection4} and~\ref{tab:layer selection5} serve as evidence for the existence of downstream task-specific and general knowledge layers. Failure to select these layers, due to reasons like random selection or lack of convergence, leads to poor performance.

In summary, it is evident that almost all LLMs feature downstream task-specific layers and general knowledge layers. Fine-tuning these layers effectively mitigates model overfitting and enhances both generalization and performance. Fortunately, \ours{} accurately and efficiently identifies both the downstream task-specific layers and the general knowledge layers.
\begin{table*}[bh]
\centering
\tiny
\caption{Comprehensive overview of layer selection strategies in main experiments. This table presents a detailed breakdown of the layer selection strategies used in different experiments involving the Llama3-8B model and its variants (\ours{}, LoRA-drop, DoRA + \ours{}, and rsLoRA + \ours{}). For each model, the specific datasets utilized (HellaSwag, PIQA, RACE, and Winogrande) are listed along with the corresponding layers selected for each dataset. The ``Layer selection" column provides the indices of the layers chosen for each experiment, indicating the specific layers of the model that were fine-tuned or modified. Additionally, the ``Parameter(M)" column indicates the total number of parameters (in millions) used in each configuration. This detailed breakdown allows for a clear understanding of the experimental setup, the layer selection process, and the parameter allocation across different models and datasets, facilitating a deeper analysis of the impact of these strategies on model performance. Unless otherwise specified, the results are based on the default LoRA Rank of 8.}
\label{tab:layer selection1}
\begin{tabular*}{\textwidth}{@{\extracolsep{\fill}} lccc}
\toprule
\textbf{Methods} &\textbf{Dataset} & \textbf{Layer selection} & \textbf{Parameter(M)}  \\ 
\midrule
\multirow{4}{*}{Llama3-8B + \ours{}($r=8$)} & Hellaswag & [0, 1, 2, 3, 4, 5, 6, 14, 15, 19, 20, 21, 23, 24, 26, 27, 28, 29, 31] & 2.0  \\
& PIQA & [1, 2, 3, 4, 5, 7, 8, 9, 14, 20, 25, 26, 27, 28, 29, 30] & 1.7  \\
& RACE & [0, 1, 2, 3, 4, 7, 8, 9, 12, 14, 25, 26, 27, 28, 29, 31] & 1.7 \\
& Winogrande  & [0, 1, 2, 3, 4, 16, 20, 22, 23, 24, 25, 26, 27, 28, 29, 31] & 1.7 \\
\midrule
\multirow{4}{*}{Llama3-8B + \ours{}($r=16$)} & Hellaswag & [0, 1, 2, 3, 4, 5, 10, 14, 18, 19, 20, 21, 23, 24, 26, 27, 28, 29, 31] & 2.0  \\
& PIQA & [1, 2, 3, 4, 5, 7, 8, 10, 14, 20, 25, 26, 27, 28, 29, 30] & 1.7  \\
& RACE & [0, 1, 2, 3, 4, 7, 8, 9, 11, 14, 25, 26, 27, 28, 29, 31] & 1.7 \\
& Winogrande  & [0, 1, 2, 3, 4, 18, 19, 22, 23, 24, 25, 26, 27, 28, 29, 31] & 1.7 \\
\midrule
\multirow{4}{*}{Llama3-8B + \ours{}($r=32$)} & Hellaswag & [0, 1, 2, 3, 4, 5, 6, 11, 15, 18, 20, 21, 23, 24, 26, 27, 28, 29, 31] & 2.0  \\
& PIQA & [1, 2, 3, 4, 5, 7, 10, 12, 14, 20, 25, 26, 27, 28, 29, 30] & 1.7  \\
& RACE & [0, 1, 2, 3, 4, 7, 8, 10, 12, 14, 24, 26, 27, 28, 29, 31] & 1.7 \\
& Winogrande  & [0, 1, 2, 3, 4, 16, 19, 20, 23, 24, 25, 26, 27, 28, 29, 31] & 1.7 \\
\midrule
\multirow{4}{*}{Llama3-8B + LoRA-drop} & Hellaswag & [13, 14, 15, 16, 17, 18, 19, 20, 21, 22, 23, 24, 25, 26, 27, 28, 29, 30, 31] & 2.0  \\
& PIQA & [16, 17, 18, 19, 20, 21, 22, 23, 24, 25, 26, 27, 28, 29, 30, 31] & 1.7  \\
& RACE & [16, 17, 18, 19, 20, 21, 22, 23, 24, 25, 26, 27, 28, 29, 30, 31] & 1.7 \\
& Winogrande  & [16, 17, 18, 19, 20, 21, 22, 23, 24, 25, 26, 27, 28, 29, 30, 31] & 1.7 \\
\midrule
\multirow{4}{*}{Llama3-8B + DoRA + \ours{}} & Hellaswag & [0, 1, 2, 4, 5, 14, 15, 19, 20, 21, 23, 26, 27, 28, 29, 31] & 1.8  \\
& PIQA & [0, 1, 2, 4, 7, 23, 24, 25, 26, 27, 28, 29, 31] & 1.5 \\
& RACE & [1, 3, 4, 7, 9, 12, 14, 23, 25, 27, 28, 29, 31] & 1.3\\
& Winogrande  & [0, 1, 2, 3, 20, 21, 22, 23, 24, 25, 26, 27, 28, 29, 31] & 1.7 \\
\midrule
\multirow{4}{*}{Llama3-8B + rsLoRA + \ours{}} & Hellaswag & [0, 1, 2, 4, 6, 14, 15, 19, 20, 21, 23, 25, 26, 27, 28, 29, 31] & 1.8  \\
& PIQA & [0, 1, 2, 3, 15, 20, 21, 25, 26, 27, 28, 29, 31] & 1.3  \\
& RACE & [0, 1, 2, 3, 7, 8, 12, 13, 25, 26, 27, 28, 29, 31] & 1.5 \\
& Winogrande  & [1, 2, 3, 6, 14, 15, 18, 20, 21, 22, 23, 24, 25, 26, 27, 28, 29, 31] & 1.9 \\
\midrule
\multirow{4}{*}{Llama3-8B + LoReFT + \ours{}} & Hellaswag & [0, 1, 2, 3, 4, 5, 6, 19, 20, 21, 23, 26, 27, 28, 29, 31] & 1.8  \\
& PIQA & [0, 1, 2, 4, 7, 22, 24, 25, 26, 27, 28, 29, 30, 31] & 1.5 \\
& RACE & [0, 1, 3, 4, 7, 9, 14, 23, 25, 27, 28, 29, 31] & 1.3\\
& Winogrande  & [0, 1, 2, 3, 4, 5, 22, 23, 24, 25, 26, 27, 28, 29, 31] & 1.7 \\
\midrule
\multirow{4}{*}{Llama3-8B + MoSLoRA + \ours{}} & Hellaswag & [0, 1, 2, 4, 5, 14, 16, 19, 20, 21, 23, 25, 26, 27, 28, 29, 31] & 1.8  \\
& PIQA & [0, 1, 2, 3, 4, 20, 21, 25, 26, 27, 28, 29, 31] & 1.3  \\
& RACE & [0, 1, 2, 3, 7, 9, 11, 13, 25, 26, 27, 28, 29, 31] & 1.5 \\
& Winogrande  & [1, 2, 3, 8, 10 15, 18, 20, 21, 22, 23, 24, 25, 26, 27, 28, 29, 31] & 1.9 \\
\bottomrule
\end{tabular*}
\end{table*}
\begin{table*}[t]
\centering
\tiny
\caption{Detailed display of selected layers in the second ablation study. In the second ablation experiment, we manually determined the number of fine-tuning layers and contrasted the performance of \ours{} with random layer selection strategies. This table presents the results of this experiment, showcasing different configurations where a specific number of layers (6, 12, 18, and 24) were selected for fine-tuning. For each configuration, the table compares the layers selected by \ours{} with those selected randomly. The datasets used in this experiment include HellaSwag, PIQA, RACE, and Winogrande. The ``Layer selection" column lists the indices of the layers chosen for fine-tuning in each dataset, while the ``Parameter(M)" column indicates the total number of parameters (in millions) used in each configuration. This detailed breakdown provides insights into how different layer selection strategies, with a manually determined number of fine-tuning layers, impact the  performance of model across different datasets, facilitating a comprehensive comparison between \ours{} and random selection methods.}
\label{tab:layer selection2}
\begin{tabular*}{\textwidth}{@{\extracolsep{\fill}} lccc}
\toprule
\textbf{Methods} &\textbf{Dataset} & \textbf{Layer selection} & \textbf{Parameter(M)}  \\ 
\midrule

\multirow{4}{*}{Select first 6 layers by \ours{} } & Hellaswag & [0, 26, 27, 28, 29, 31] & 0.6  \\
& PIQA & [2, 4, 26, 27, 28, 29] & 0.6  \\
& RACE & [0, 7, 12, 27, 28, 29] & 0.6 \\
& Winogrande  & [22, 23, 24, 26, 27, 28] & 0.6 \\
\midrule
\multirow{4}{*}{Random selection 6 layers} & Hellaswag & [2, 4, 11, 19, 23, 25] & 0.6  \\
& PIQA & [2, 4, 11, 19, 23, 25] & 0.6  \\
& RACE & [2, 4, 11, 19, 23, 25] & 0.6 \\
& Winogrande  & [2, 4, 11, 19, 23, 25] & 0.6 \\
\midrule
\multirow{4}{*}{Select first 12 layers by \ours{}} & Hellaswag & [0, 2, 3, 14, 15, 21, 23, 26, 27, 28, 29, 31] & 1.3  \\
& PIQA & [1, 2, 3, 4, 7, 20, 25, 26, 27, 28, 29, 30] & 1.3  \\
& RACE & [0, 1, 3, 7, 8, 12, 13, 25, 27, 28, 29, 31] & 1.3 \\
& Winogrande  & [0, 3, 20, 22, 23, 24, 25, 26, 27, 28, 29, 31] & 1.3 \\
\midrule
\multirow{4}{*}{Random selection 12 layers} & Hellaswag & [1, 3, 4, 12, 14, 18, 20, 21, 22, 27, 29, 31] & 1.3  \\
& PIQA & [1, 3, 4, 12, 14, 18, 20, 21, 22, 27, 29, 31] & 1.3  \\
& RACE & [1, 3, 4, 12, 14, 18, 20, 21, 22, 27, 29, 31] & 1.3 \\
& Winogrande  & [1, 3, 4, 12, 14, 18, 20, 21, 22, 27, 29, 31] & 1.3 \\
\midrule
\multirow{4}{*}{Select first 18 layers by \ours{}} & Hellaswag & [0, 1, 2, 3, 4, 5, 6, 14, 15, 19, 21, 23, 26, 27, 28, 29, 30, 31] & 1.9  \\
& PIQA &  [0, 1, 2, 3, 4, 5, 7, 8, 19, 20, 23, 25, 26, 27, 28, 29, 30, 31] & 1.9  \\
& RACE & [0, 1, 2, 3, 4, 7, 8, 9, 10, 12, 13, 15, 25, 27, 28, 29, 30, 31] & 1.9 \\
& Winogrande  & [0, 1, 3, 5, 7, 9, 15, 20, 21, 22, 23, 24, 25, 26, 27, 28, 29, 31] & 1.9 \\
\midrule
\multirow{4}{*}{Random selection 18 layers} & Hellaswag & [1, 2, 5, 8, 9, 10, 12, 13, 17, 18, 20, 21, 22, 23, 24, 25, 26, 30] & 1.9  \\
& PIQA & [1, 2, 5, 8, 9, 10, 12, 13, 17, 18, 20, 21, 22, 23, 24, 25, 26, 30] & 1.9 \\
& RACE & [1, 2, 5, 8, 9, 10, 12, 13, 17, 18, 20, 21, 22, 23, 24, 25, 26, 30] & 1.9 \\
& Winogrande  & [1, 2, 5, 8, 9, 10, 12, 13, 17, 18, 20, 21, 22, 23, 24, 25, 26, 30] & 1.9 \\
\midrule
\multirow{4}{*}{Select first 24 layers by \ours{} } & Hellaswag & [0, 1, 2, 3, 4, 5, 6, 11, 12, 13, 14, 15, 18, 19, 20, 21, 23, 24, 26, 27, 28, 29, 30, 31] & 2.6  \\
& PIQA & [0, 1, 2, 3, 4, 5, 6, 7, 8, 9, 15, 18, 19, 20, 21, 23, 24, 25, 26, 27, 28, 29, 30, 31] & 2.6  \\
& RACE & [0, 1, 2, 3, 4, 5, 6, 7, 8, 9, 10, 11, 12, 13, 14, 15, 23, 24, 25, 27, 28, 29, 30, 31] & 2.6 \\
& Winogrande  & [0, 1, 2, 3, 4, 5, 7, 8, 9, 10, 15, 19, 20, 21, 22, 23, 24, 25, 26, 27, 28, 29, 30, 31] & 2.6 \\
\midrule
\multirow{4}{*}{Random selection 24 layers} & Hellaswag & [0, 1, 4, 6, 7, 8, 9, 10, 12, 13, 14, 15, 16, 17, 18, 19, 21, 23, 25, 26, 27, 28, 30, 31] & 2.6  \\
& PIQA & [0, 1, 4, 6, 7, 8, 9, 10, 12, 13, 14, 15, 16, 17, 18, 19, 21, 23, 25, 26, 27, 28, 30, 31] & 2.6  \\
& RACE & [0, 1, 4, 6, 7, 8, 9, 10, 12, 13, 14, 15, 16, 17, 18, 19, 21, 23, 25, 26, 27, 28, 30, 31] & 2.6 \\
& Winogrande  & [0, 1, 4, 6, 7, 8, 9, 10, 12, 13, 14, 15, 16, 17, 18, 19, 21, 23, 25, 26, 27, 28, 30, 31] & 2.6 \\
\bottomrule
\end{tabular*}
\end{table*}

\begin{table*}[t]
\centering
\scriptsize
\caption{Comprehensive overview of layer selection strategies and parameter allocation in various experiments. This table provides an in-depth breakdown of the layer selection strategies employed across different models and datasets in the experiments. The models tested include Llama3-8B, Chatglm3-6B, Mistral-7B-v0.1 and others, all combined with \ours{}. For each model, the specific datasets used (HellaSwag, PIQA, RACE, and Winogrande) are listed along with the corresponding layers selected for each dataset. The ``Layer selection" column details the indices of the layers chosen for each experiment, indicating the specific layers of the model that were fine-tuned or modified. Additionally, the ``Parameter(M)" column indicates the total number of parameters (in millions) used in each configuration. This detailed breakdown allows for a clear understanding of the experimental setup, the layer selection process, and the parameter allocation across different models and datasets, facilitating a deeper analysis of the impact of these strategies on model performance.}
\label{tab:layer selection3}
\begin{tabular*}{\textwidth}{@{\extracolsep{\fill}} lccc}
\toprule
\textbf{Methods} &\textbf{Dataset} & \textbf{Layer selection} & \textbf{Parameter(M)}  \\ 
\midrule
\multirow{4}{*}{Llama3-8B + \ours{}} & Hellaswag & [0, 1, 2, 3, 4, 5, 6, 14, 15, 19, 20, 21, 23, 24, 26, 27, 28, 29, 31] & 2.0  \\
& PIQA & [1, 2, 3, 4, 5, 7, 8, 9, 14, 20, 25, 26, 27, 28, 29, 30] & 1.7  \\
& RACE & [0, 1, 2, 3, 4, 7, 8, 9, 12, 14, 25, 26, 27, 28, 29, 31] & 1.7 \\
& Winogrande  & [0, 1, 2, 3, 4, 16, 20, 22, 23, 24, 25, 26, 27, 28, 29, 31] & 1.7 \\
\midrule
\multirow{4}{*}{Chatglm3-6B + \ours{}} & Hellaswag & [1, 2, 3, 4, 5, 6, 7, 10, 12, 13, 16, 18, 20] & 0.9 \\
& PIQA & [0, 1, 2, 3, 5, 6, 7, 8, 9, 19, 21, 23, 25, 27] & 1.0  \\
& RACE & [2, 6, 8, 9, 10, 11, 14, 15, 16, 17, 18, 20, 23, 26] & 1.0 \\
& Winogrande  & [0, 2, 6, 8, 9, 11, 12, 13, 16, 17, 18, 20, 25, 26] & 1.0 \\
\midrule
\multirow{4}{*}{Mistral-7B-v0.1 + \ours{}} & Hellaswag & [0, 1, 2, 3, 4, 5, 6, 7, 14, 22, 26, 27, 30] & 1.5  \\
& PIQA & [6, 8, 14, 17, 18, 22, 23, 24, 25, 26, 27, 28, 29, 30] & 1.7  \\
& RACE & [0, 2, 3, 4, 5, 6, 7, 8, 9, 10, 11, 13, 14, 17, 30, 31]  & 1.7 \\
& Winogrande  & [0, 1, 2, 3, 4, 5, 6, 7, 22, 23, 24, 25, 26, 27, 28, 29, 30, 31] & 1.9 \\
\midrule
\multirow{4}{*}{Gemma-7B + \ours{}} & Hellaswag & [0, 1, 2, 3, 4, 5, 6, 7, 8, 9, 14, 15, 16, 18, 20, 23, 27] & 1.9  \\
& PIQA & [0, 1, 8, 9, 10, 12, 15, 16, 17, 20, 21, 22, 23, 24, 25, 26, 27] & 1.9  \\
& RACE & [0, 1, 2, 3, 4, 5, 6, 7, 8, 9, 15, 16] & 1.4 \\
& Winogrande  & [0, 1, 2, 3, 4, 5, 6, 7, 8, 18, 19, 20, 21, 22, 23, 24, 25, 26, 27] & 2.1 \\
\midrule
\multirow{4}{*}{Vicuna-7B-v1.5 + \ours{}} & Hellaswag & [0, 1, 2, 3, 4, 6, 7, 8, 9, 10, 11, 12] & 1.6  \\
& PIQA & [1, 2, 3, 5, 7, 8, 11, 12, 13, 14, 21, 31] & 1.6  \\
& RACE & [2, 3, 4, 5, 6, 7, 8, 9, 10, 11, 12, 13] & 1.6 \\
& Winogrande  & [0, 2, 3, 4, 6, 8, 9, 12, 20, 21, 22, 23, 24, 25, 26, 27, 28, 29, 30, 31] & 2.6 \\
\midrule
\multirow{4}{*}{Zephyr-7B-beta + \ours{}} & Hellaswag & [1, 13, 15, 17, 18, 22, 23, 24, 25, 26, 27, 28, 30, 31] & 1.5  \\
& PIQA & [2, 3, 6, 7, 14, 15, 16, 17, 22, 26, 27, 28] & 1.4  \\
& RACE & [1, 2, 4, 6, 7, 9, 11, 13, 14, 17, 26, 30, 31] & 1.4 \\
& Winogrande  & [1, 3, 5, 6, 8, 13, 27, 28, 29, 30, 31] & 1.2 \\
\midrule
\multirow{4}{*}{Yi-6B + \ours{}} & Hellaswag & [0, 1, 2, 3, 4, 6, 8, 9, 10, 19, 20, 21, 22] & 1.3  \\
& PIQA & [1, 2, 3, 5, 6, 7, 8, 9, 12, 13, 15, 16, 17, 18, 20, 23] & 1.6 \\
& RACE & [1, 3, 5, 6, 7, 9, 11, 12, 13, 14, 17, 21] & 1.2 \\
& Winogrande  & [0, 1, 2, 3, 5, 6, 7, 11, 23, 26, 27, 30, 31] & 1.3 \\
\midrule
\multirow{4}{*}{Llama-7B + \ours{}} & Hellaswag & [0, 1, 2, 4, 5, 6, 8, 12, 16, 30, 31] & 1.4 \\
& PIQA & [2, 12, 14, 15, 16, 21, 22, 23, 24, 25, 26, 27, 28, 29, 30, 31] & 2.1 \\
& RACE & [4, 5, 6, 7, 8, 10, 11, 23, 30, 31] & 1.3 \\
& Winogrande  & [0, 2, 3, 6, 7, 8, 10, 11, 13, 16, 23, 28, 29, 30, 31] & 2.0 \\
\midrule
\multirow{4}{*}{Llama2-7B + \ours{}} & Hellaswag & [0, 1, 2, 3, 4, 5, 6, 7, 8, 12] & 1.3  \\
& PIQA & [0, 1, 2, 3, 7, 8, 11, 13, 14, 21, 24, 29, 30, 31] & 1.8  \\
& RACE & [0, 1, 2, 4, 5, 6, 7, 8, 9, 10, 11, 12, 14, 16] & 1.8 \\
& Winogrande  & [0, 1, 3, 4, 8, 14, 15, 16, 17, 20, 21, 22, 23, 24, 25, 26, 28, 29, 30] & 2.5 \\
\midrule
\multirow{4}{*}{XuanYuan-6B + \ours{}} & Hellaswag & [1, 2, 3, 4, 5, 6, 7, 8, 9, 12, 13, 14, 17] & 1.7  \\
& PIQA & [3, 4, 7, 8, 12, 14, 16, 17, 19, 21, 23, 25, 28, 29] & 1.8  \\
& RACE & [0, 4, 5, 6, 7, 8, 9, 10, 11, 12, 14, 16, 17, 20, 21, 22, 25, 28, 29]  & 2.5 \\
& Winogrande  & [2, 3, 4, 8, 9, 10, 14, 15, 19, 20, 21, 22, 23, 24, 25, 26, 27, 28, 29] & 2.5 \\
\midrule
\multirow{4}{*}{Qwen1.5-7B + \ours{}} & Hellaswag & [0, 1, 2, 3, 4, 5, 6, 7, 9, 17] & 1.3  \\
& PIQA & [0, 1, 2, 3, 4, 5, 6, 7, 8, 11, 13, 14, 15, 17] & 1.8  \\
& RACE & [0, 1, 2, 3, 4, 5, 6, 7, 8, 9, 10, 11, 12] & 1.7 \\
& Winogrande  & [0, 1, 2, 3, 4, 5, 6, 7, 8, 21, 24, 25, 27, 28, 30] & 2.0 \\

\bottomrule
\end{tabular*}
\end{table*}

\begin{table*}[t]
\centering
\scriptsize
\caption{Detailed display of selected layers in the first ablation study. This table presents the results of the first ablation experiment, where the number of layers selected by \ours{} was kept constant, but different layers were chosen for fine-tuning. The table includes three different random layer selection strategies (Random1, Random2, and Random3) applied to various datasets (HellaSwag, PIQA, RACE, and Winogrande). For each random selection method, the ``Layer selection" column lists the indices of the layers chosen for fine-tuning in each dataset. The ``Parameter(M)" column indicates the total number of parameters (in millions) used in each configuration. This detailed breakdown allows for a clear understanding of how different layer selection strategies impact the performance of model across different datasets while maintaining a consistent number of layers for fine-tuning.}
\label{tab:layer selection4}
\begin{tabular*}{\textwidth}{@{\extracolsep{\fill}} lccc}
\toprule
\textbf{Methods} &\textbf{Dataset} & \textbf{Layer selection} & \textbf{Parameter(M)}  \\ 
\midrule
\multirow{4}{*}{Random1} & Hellaswag & [0, 1, 2, 3, 4, 6, 7, 8, 10, 11, 14, 18, 19, 20, 21, 25, 26, 27, 28] & 2.0  \\
& PIQA & [0, 2, 4, 10, 12, 16, 17, 18, 23, 24, 25, 26, 27, 28, 29, 30] & 1.7 \\
& RACE & [1, 2, 4, 7, 9, 11, 12, 14, 15, 18, 20, 23, 24, 26, 28, 30] & 1.7 \\
& Winogrande  & [1, 2, 4, 5, 9, 10, 11, 13, 15, 17, 20, 21, 24, 26, 30, 31] & 1.7 \\
\midrule
\multirow{4}{*}{Random2} & Hellaswag & [0, 2, 3, 4, 5, 6, 10, 12, 13, 15, 17, 20, 21, 22, 23, 24, 28, 29, 30] & 2.0  \\
& PIQA & [0, 1, 3, 4, 8, 13, 14, 18, 19, 22, 24, 26, 28, 29, 30, 31] & 1.7  \\
& RACE & [5, 6, 7, 8, 9, 11, 12, 13, 15, 19, 20, 21, 25, 27, 28, 30] & 1.7 \\
& Winogrande  & [2, 5, 6, 7, 8, 10, 11, 13, 14, 17, 18, 22, 25, 26, 28, 30] & 1.7 \\
\midrule
\multirow{4}{*}{Random3} & Hellaswag & [0, 1, 3, 4, 6, 9, 12, 13, 17, 18, 19, 20, 21, 22, 25, 26, 27, 28, 29] & 2.0  \\
& PIQA & [0, 3, 4, 9, 12, 13, 14, 15, 16, 24, 25, 26, 27, 28, 30, 31] & 1.7  \\
& RACE & [0, 1, 2, 9, 11, 12, 14, 18, 19, 20, 21, 23, 25, 26, 29, 30] & 1.7 \\
& Winogrande  & [2, 4, 6, 8, 10, 12, 14, 16, 17, 18, 20, 22, 23, 29, 30, 31] & 1.7 \\
\bottomrule
\end{tabular*}
\end{table*}

\begin{table*}[t]
\centering
\scriptsize
\caption{Detailed display of layer selection with varying numbers of searching samples. This table presents the results of an experiment where different numbers of searching samples (1000, 2000, 5000, and 10000) were used to determine the layers for \ours{}. The datasets involved in this experiment include HellaSwag, PIQA, RACE, and Winogrande. For each number of searching samples, the ``Layer selection" column lists the indices of the layers chosen for fine-tuning in each dataset. The ``Parameter(M)" column indicates the total number of parameters (in millions) used in each configuration. This detailed breakdown provides insights into how the number of searching samples impacts the layer selection process and the performance of model across different datasets. }
\label{tab:layer selection5}
\begin{tabular*}{\textwidth}{@{\extracolsep{\fill}} lccc}
\toprule
\textbf{Methods} &\textbf{Dataset} & \textbf{Layer selection} & \textbf{Parameter(M)}  \\ 
\midrule
\multirow{4}{*}{1000 searching samples} & Hellaswag & [0, 2, 4, 5, 6, 8, 10, 16, 21, 26, 27, 28, 30, 31]  & 1.5  \\
& PIQA & [0, 1, 2, 3, 4, 16, 25, 26, 27, 28, 29, 30, 31] & 1.4  \\
& RACE & [0, 1, 2, 3, 4, 16, 21, 28, 29, 30, 31] & 1.2 \\
& Winogrande  & [0, 1, 2, 3, 4, 16, 20, 25, 26, 27, 28, 29, 30, 31] & 1.5 \\
\midrule
\multirow{4}{*}{2000 searching samples} & Hellaswag & [1, 2, 3, 4, 8, 10, 11, 16, 30, 31] & 1.0  \\
& PIQA & [0, 1, 2, 20, 22, 23, 24, 25, 26, 27, 28, 29, 30, 31] & 1.5  \\
& RACE & [0, 1, 2, 3, 4, 10, 20, 23, 27, 28, 29, 30, 31] & 1.4 \\
& Winogrande  & [0, 1, 2, 3, 4, 20, 25, 27, 30, 31] & 1.0 \\
\midrule
\multirow{4}{*}{5000 searching samples} & Hellaswag & [0, 1, 2, 3, 4, 8, 31] & 0.7 \\
& PIQA & [0, 2, 3, 4, 20, 22, 23, 24, 25, 26, 27, 28, 29, 30, 31] & 1.6  \\
& RACE & [1, 3, 4, 6, 9, 10, 11, 12, 14, 27, 28, 29, 30, 31] & 1.5 \\
& Winogrande  & [1, 2, 3, 4, 6, 7, 8, 9, 26, 27, 30, 31] & 1.3 \\
\midrule
\multirow{4}{*}{10000 searching samples} & Hellaswag & [0, 1, 4, 10, 12, 14, 21, 24, 26, 27, 28, 29, 30, 31] & 1.5  \\
& PIQA & [0, 1, 3, 4, 7, 20, 21, 22, 24, 25, 26, 27, 28, 29, 30, 31] & 1.7 \\
& RACE & [1, 2, 7, 13, 14, 23, 25, 26, 27, 28, 29, 31] & 1.3 \\
& Winogrande  & [6, 7, 9, 10, 15, 19, 20, 22, 26, 27, 30, 31] & 1.3 \\
\bottomrule
\end{tabular*}
\end{table*}

\section{Loss} \label{loss}
This section presents the training, evaluation, and validation loss during the \ours{} flexible layer selection and fine-tuning stages, accompanied by intuitive explanations. 
\subsection{Effectiveness of \ours{}.} \label{main loss}
\begin{figure}[t]
\centering
\includegraphics[width=\textwidth]{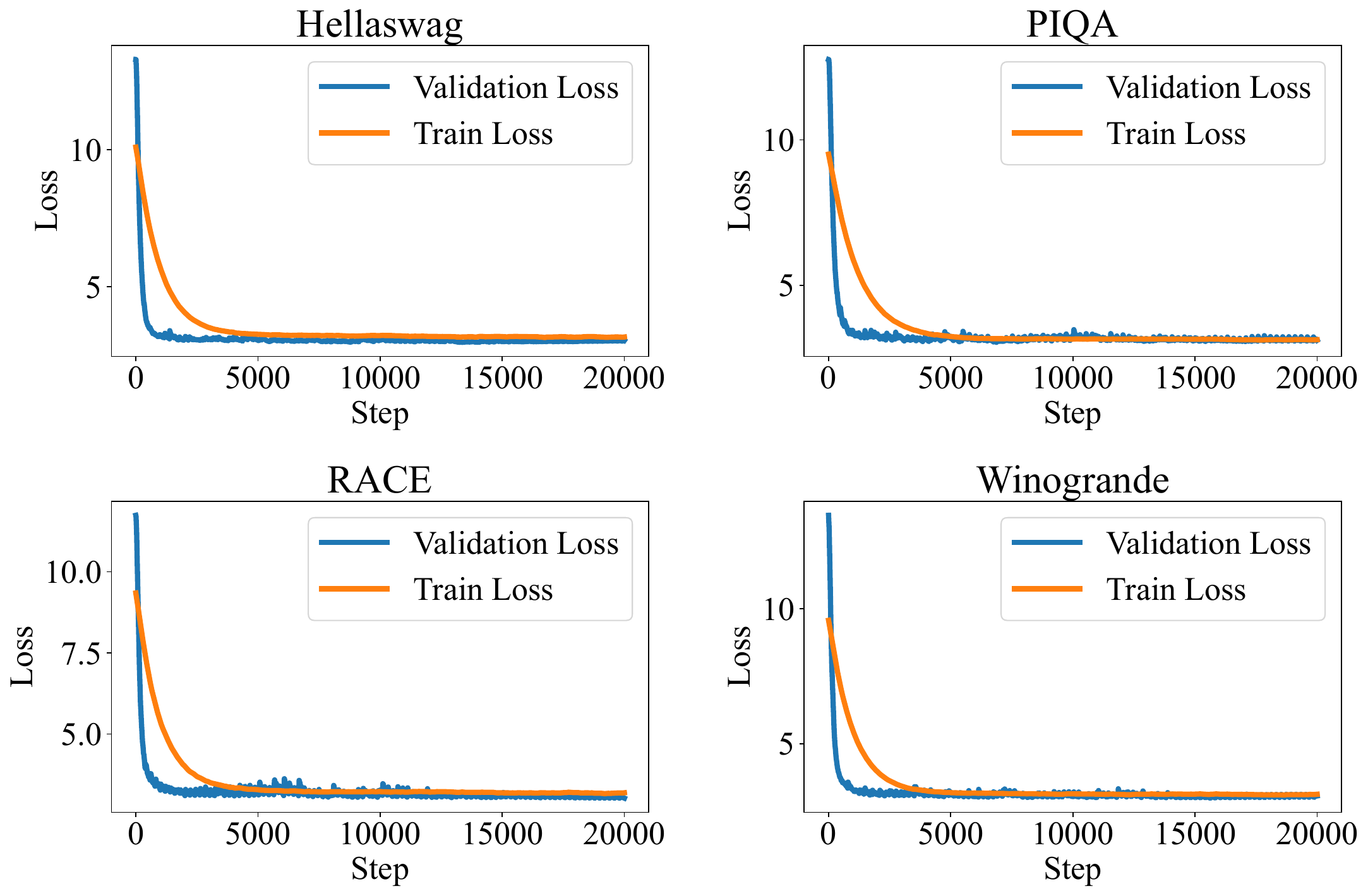} 
\caption{Training and validation loss during the flexible layer selection phase. The figure shows the training and validation loss over 20,000 steps for four different datasets (Hellaswag, PIQA, RACE, and Winogrande), where the batch size at each step is 1. The blue line shows the validation loss and the orange line shows the training loss. These plots visually compare how the performance of the models changes during the flexible layer selection phase, highlighting the convergence behavior.}
\label{fig5}
\end{figure}

Figure~\ref{fig5} plots the training and validation loss curves for Llama-8B during the flexible layer selection stage across four different datasets over one epoch. Both inner and outer layer optimizations are observed to converge well during the flexible layer selection stage, demonstrating the effectiveness of \ours{}.

\subsection{\ours{} can Correctly Identify Critical Layers.}  \label{ablation study 1 loss}
\begin{figure}[t]
\centering
\includegraphics[width=0.47 \textwidth]{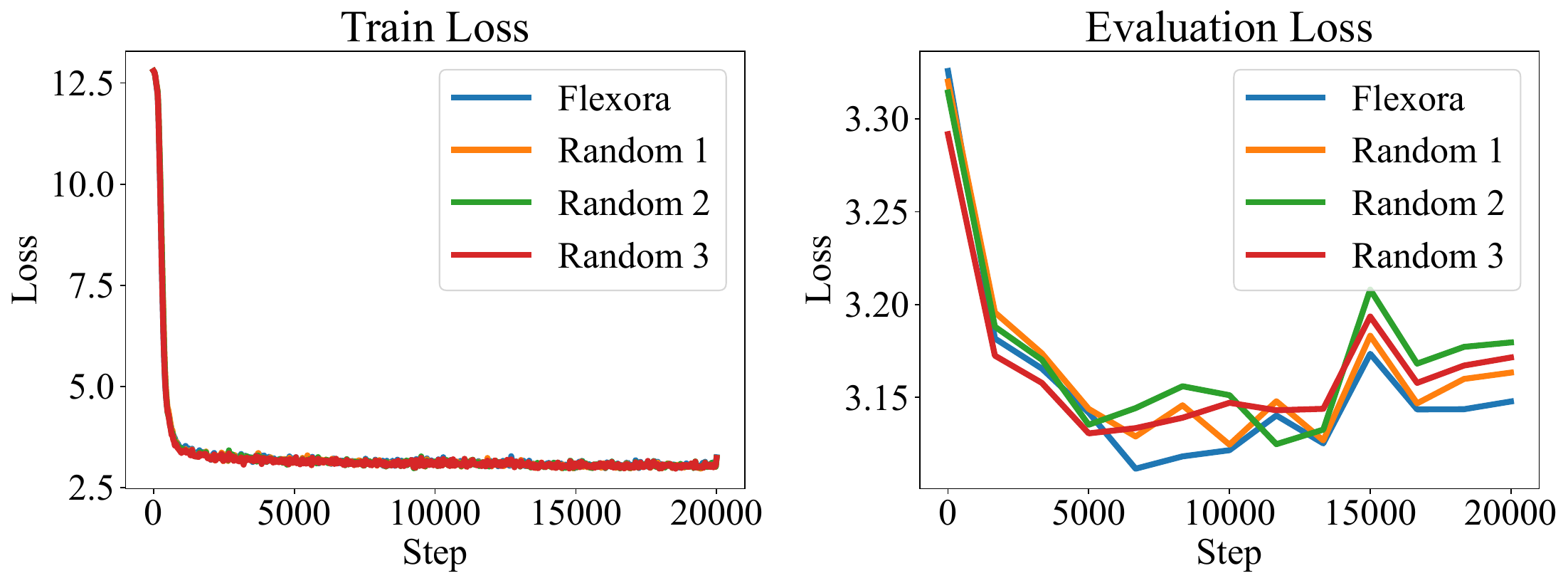} 
\caption{Comparison of train loss and evaluation loss in the Hellaswag dataset during the first ablation study. This figure presents the train loss (left) and evaluation loss (right) over 20,000 steps for the Hellaswag dataset, where the batch size at each step is 1. The performance of the \ours{} method is compared against three different random layer selection strategies (Random 1, Random 2, and Random 3). The train loss graph shows how the training performance of model evolves, while the evaluation loss graph highlights the generalization capability of model on the validation set. This detailed comparison provides insights into the effectiveness of \ours{} relative to random selection methods in terms of both training and evaluation metrics.
}
\label{fig6}
\end{figure}

\begin{figure}[t]
\centering
\includegraphics[width=0.47\textwidth]{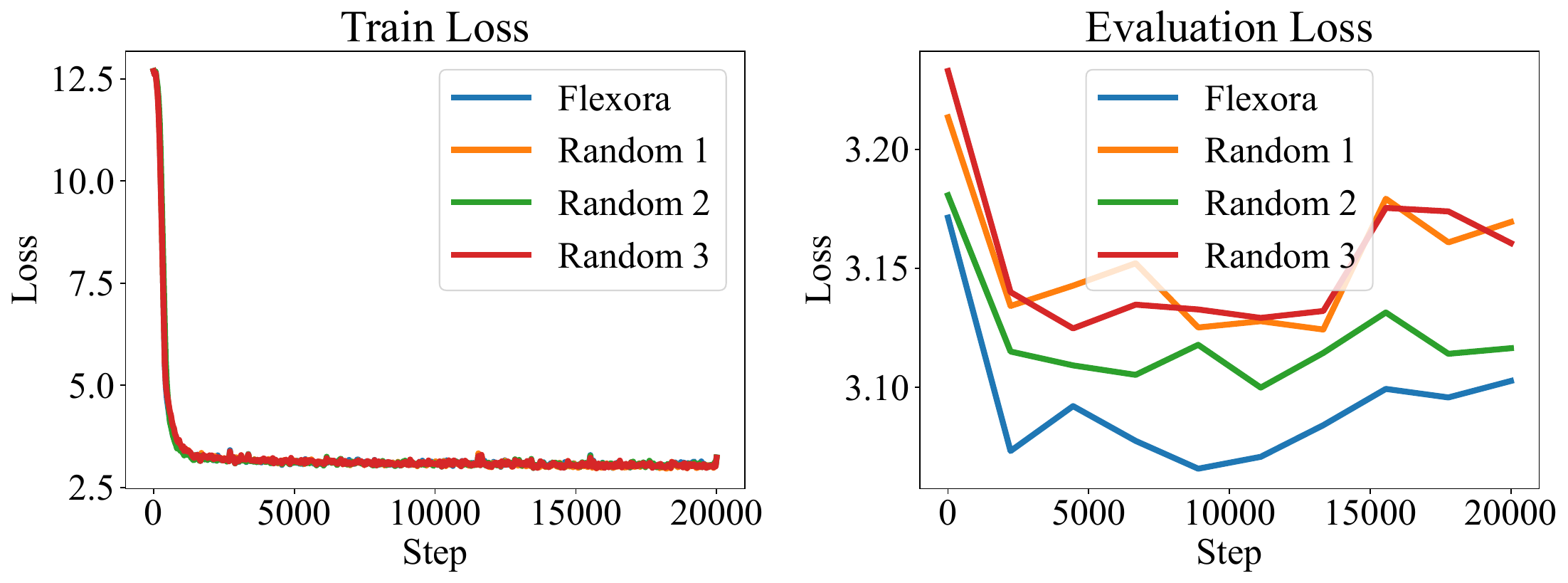} 
\caption{Comparison of train loss and evaluation loss in the PIQA dataset during the first ablation study. This figure presents the train loss (left) and evaluation loss (right) over 20,000 steps for the PIQA dataset, where the batch size at each step is 1. The performance of the \ours{} method is compared against three different random layer selection strategies (Random 1, Random 2, and Random 3). The train loss graph shows how the training performance of model evolves, while the evaluation loss graph highlights the generalization capability of model on the validation set. This detailed comparison provides insights into the effectiveness of \ours{} relative to random selection methods in terms of both training and evaluation metrics.}
\label{fig7}
\end{figure}

\begin{figure}[t]
\centering
\includegraphics[width=0.47\textwidth]{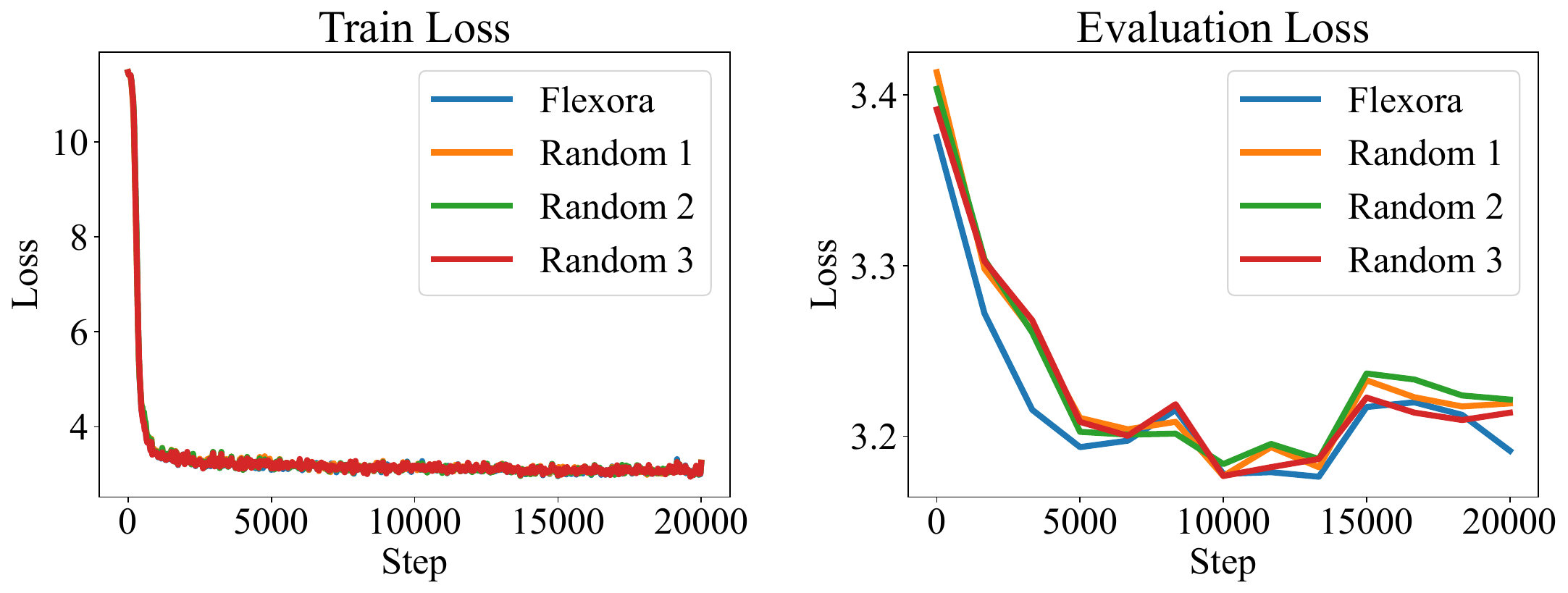} 
\caption{Comparison of train loss and evaluation loss in the RACE dataset during the first ablation study. This figure presents the train loss (left) and evaluation loss (right) over 20,000 steps for the RACE dataset, where the batch size at each step is 1. The performance of the \ours{} method is compared against three different random layer selection strategies (Random 1, Random 2, and Random 3). The train loss graph shows how the training performance of model evolves, while the evaluation loss graph highlights the generalization capability of model on the validation set. This detailed comparison provides insights into the effectiveness of \ours{} relative to random selection methods in terms of both training and evaluation metrics.}
\label{fig8}
\end{figure}

\begin{figure}[t]
\centering
\includegraphics[width=0.47\textwidth]{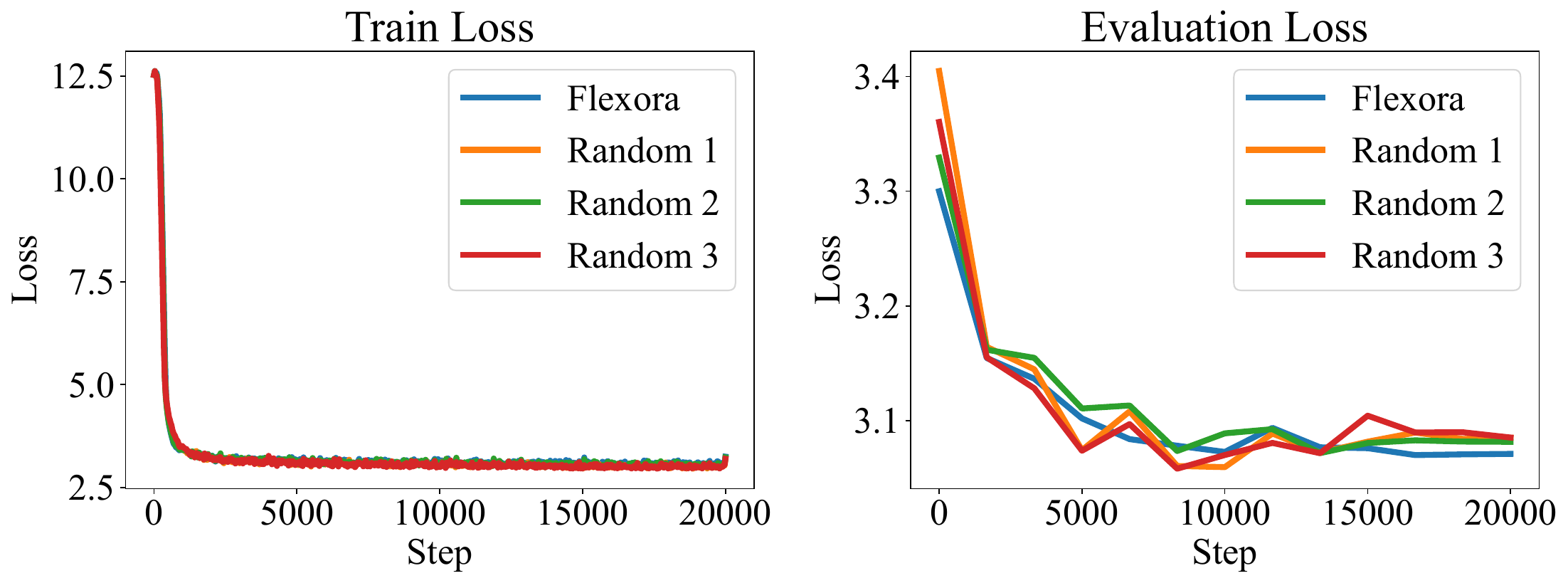} 
\caption{Comparison of train loss and evaluation loss in the Winogrande dataset during the first ablation study. This figure presents the train loss (left) and evaluation loss (right) over 20,000 steps for the Winogrande dataset, where the batch size at each step is 1. The performance of the \ours{} method is compared against three different random layer selection strategies (Random 1, Random 2, and Random 3). The train loss graph shows how the training performance of model evolves, while the evaluation loss graph highlights the generalization capability of model on the validation set. This detailed comparison provides insights into the effectiveness of \ours{} relative to random selection methods in terms of both training and evaluation metrics.}
\label{fig9}
\end{figure}

Figures~\ref{fig6},~\ref{fig7},~\ref{fig8}, and~\ref{fig9} depict the training and evaluation loss from the first ablation study. In all experiments, the training loss converges effectively, demonstrating robust training performance. However, variations in evaluation loss underscore the model's generalization capabilities. \ours{} generally surpasses methods that randomly select an equivalent number of layers, demonstrating its ability to accurately identify critical layers for more effective improvements.
\subsection{\ours{} can Reduce Overfitting.} \label{ablation study 2 loss}
\begin{figure}[t]
\centering
\includegraphics[width=0.47\textwidth]{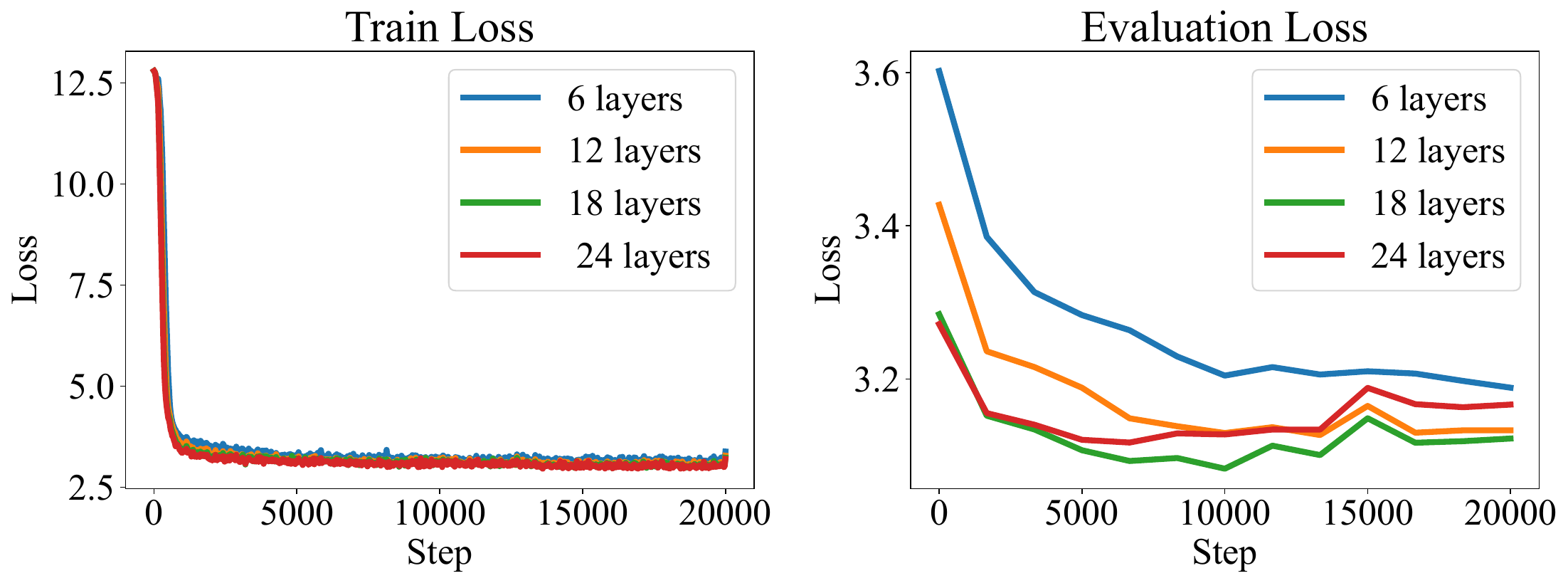} 
\caption{Training loss and evaluation loss during fine-tuning of different numbers of layers in the \ours{} on the Hellaswag dataset. This figure presents the training loss (left) and evaluation loss (right) over 20,000 steps for the Hellaswag dataset. The performance is compared across four different configurations where the first 6, 12, 18, and 24 layers of the \ours{} model are fine-tuned. The training loss graph shows that the model with 24 layers (red) achieves the lowest training loss, indicating it fits the training data very well. However, the evaluation loss graph reveals that the model with 18 layers (green) achieves the lowest evaluation loss, suggesting better generalization to unseen data. This discrepancy highlights the overfitting issue, where the model with 24 layers performs well on the training data but does not generalize as effectively as the model with 18 layers.}
\label{fig10}
\end{figure}

\begin{figure}[t]
\centering
\includegraphics[width=0.47\textwidth]{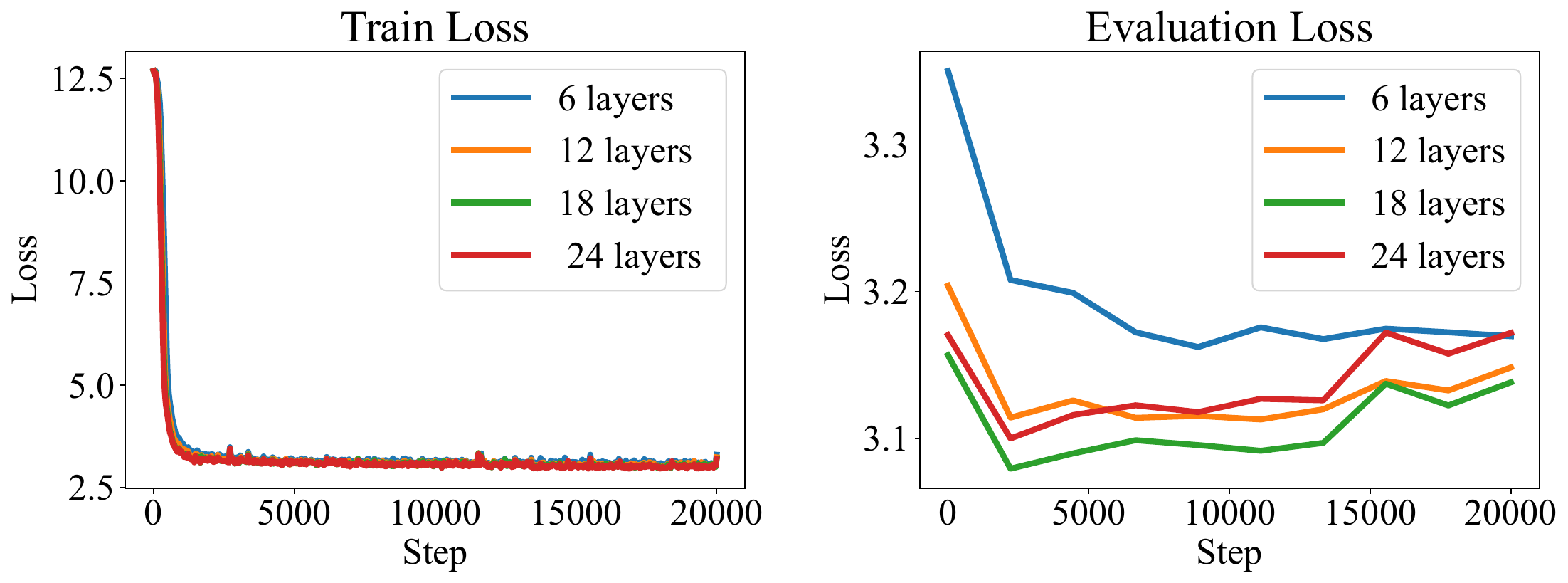} 
\caption{Training loss and evaluation loss during fine-tuning of different numbers of layers in the \ours{} on the PIQA dataset. This figure presents the training loss (left) and evaluation loss (right) over 20,000 steps for the PIQA dataset. The performance is compared across four different configurations where the first 6, 12, 18, and 24 layers of the \ours{} model are fine-tuned. The training loss graph shows that the model with 24 layers (red) achieves the lowest training loss, indicating it fits the training data very well. However, the evaluation loss graph reveals that the model with 18 layers (green) achieves the lowest evaluation loss, suggesting better generalization to unseen data. This discrepancy highlights the overfitting issue, where the model with 24 layers performs well on the training data but does not generalize as effectively as the model with 18 layers.}
\label{fig11}
\end{figure}

\begin{figure}[t]
\centering
\includegraphics[width=0.4\textwidth]{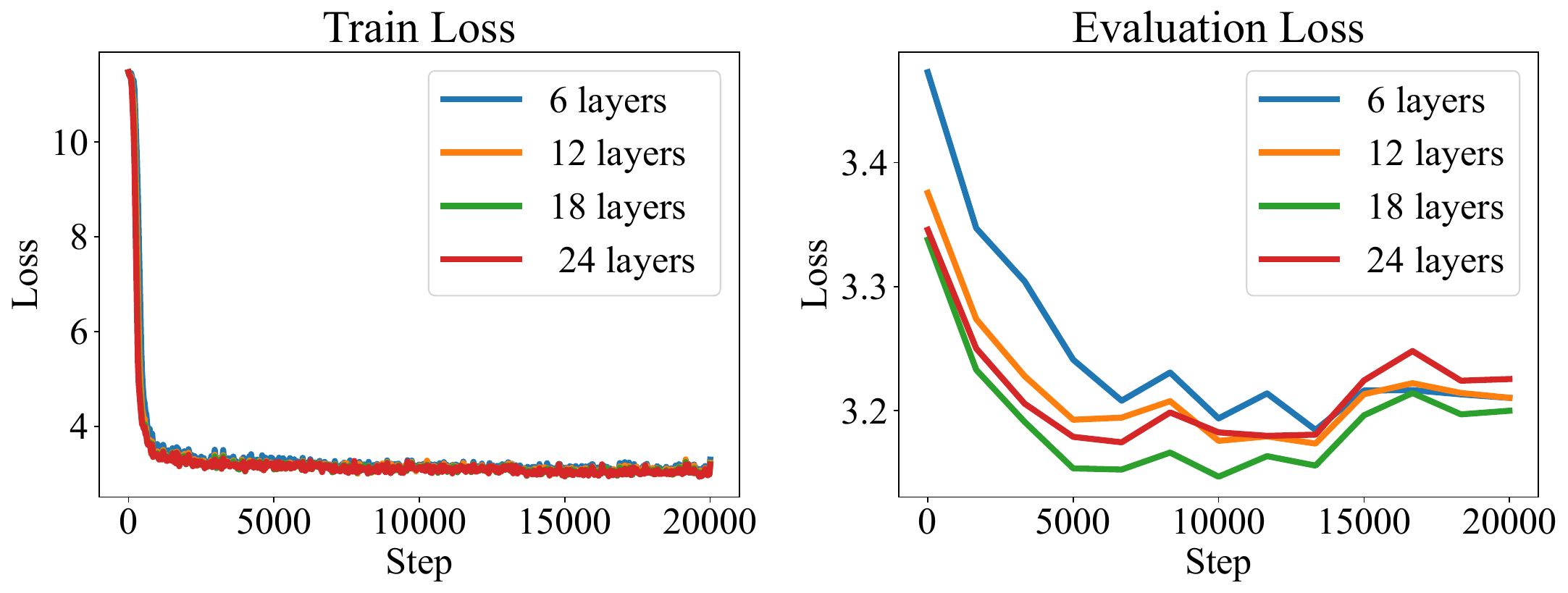} 
\caption{Training loss and evaluation loss during fine-tuning of different numbers of layers in the \ours{} on the RACE dataset. This figure presents the training loss (left) and evaluation loss (right) over 20,000 steps for the RACE dataset. The performance is compared across four different configurations where the first 6, 12, 18, and 24 layers of the \ours{} model are fine-tuned. The training loss graph shows that the model with 24 layers (red) achieves the lowest training loss, indicating it fits the training data very well. However, the evaluation loss graph reveals that the model with 18 layers (green) achieves the lowest evaluation loss, suggesting better generalization to unseen data. This discrepancy highlights the overfitting issue, where the model with 24 layers performs well on the training data but does not generalize as effectively as the model with 18 layers.}
\label{fig12}
\end{figure}

\begin{figure}[t]
\centering
\includegraphics[width=0.47\textwidth]{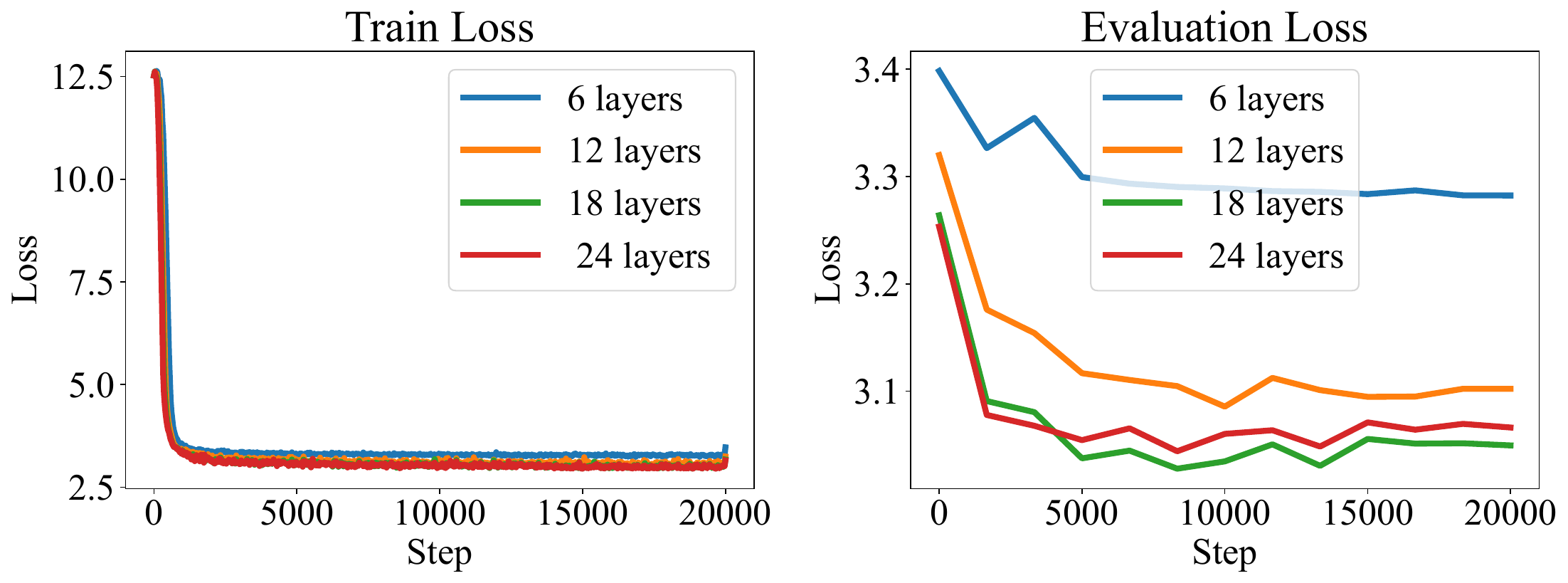} 
\caption{Training loss and evaluation loss during fine-tuning of different numbers of layers in the \ours{} on the Winogrande dataset. This figure presents the training loss (left) and evaluation loss (right) over 20,000 steps for the Winogrande dataset. The performance is compared across four different configurations where the first 6, 12, 18, and 24 layers of the \ours{} model are fine-tuned. The training loss graph shows that the model with 24 layers (red) achieves the lowest training loss, indicating it fits the training data very well. However, the evaluation loss graph reveals that the model with 18 layers (green) achieves the lowest evaluation loss, suggesting better generalization to unseen data. This discrepancy highlights the overfitting issue, where the model with 24 layers performs well on the training data but does not generalize as effectively as the model with 18 layers.}
\label{fig13}
\end{figure}

Figures~\ref{fig10},~\ref{fig11},~\ref{fig12}, and~\ref{fig13} present the training and evaluation loss from the sencond ablation study. Consistent with previous experiments, the training loss converges, indicating a strong training effect on the training set. Notably, the 24-layer (red) model consistently shows the lowest training loss, suggesting optimal learning, whereas the 6-layer (blue) model consistently records the highest, indicating poorer training performance. However, differences in evaluation loss reveal variations in model generalization across different layers. The 18-layer (green) model consistently exhibits the lowest evaluation loss, indicating superior generalization and downstream task performance, corroborated by actual results. The 24-layer (red) model's evaluation loss consistently exceeds that of the 18-layer (green) model, suggesting significant overfitting. Similarly, the 6-layer (blue) model consistently records the highest evaluation loss, indicative of underfitting.

In summary, too few training layers can lead to underfitting and poor performance, as seen in the 6-layer (blue) model. Conversely, too many layers can also result in overfitting, as evidenced by the 24-layer (red) model's performance. However following the selection strategy of \ours{}, choosing the right number of layers can minimize overfitting and improve performance
\twocolumn

\onecolumn
\section{Theoretical insights and numerical experiments on the smoothness constant of Llama3-8B } \label{sec:numerical experiments}
\subsection{Theoretical insights} \label{lora_therom}
Consider a neural network with weight matrices decomposed as:
\[
W^{(i)} = W_1^{(i)} (\text{frozen}) + W_2^{(i)} (\text{trainable}),
\]
similar to LoRA's low-rank adaptation. We analyze the gradient difference for the trainable component \( W_2^{(i)} \).

The network output is given by:
\[
y^{(N)} = \left( \prod_{j=1}^N (W_1^{(j)} + W_2^{(j)}) \right)x,
\]

Let \( \ell \) be the MSE loss:
\[
\ell = \frac{1}{2} \left( y^{(N)} - y \right)^2,
\]
where \( y^{(N)} \) is the network output. The gradient of \( \ell \) w.r.t. \( W_2^{(i)} \) is:
\begin{equation}
\frac{\partial \ell}{\partial W_2^{(i)}} = \left( \prod_{j=i+1}^N (W_1^{(j)} + W_2^{(j)}) \right) \cdot \frac{\partial \ell}{\partial y^{(i)}} \cdot x \left( \prod_{j=1}^{i-1} (W_1^{(j)} + W_2^{(j)}) \right).
\end{equation}

For two networks differing only in \( W_2^{(i)} \), the gradient difference is:
\begin{align}
\norm{\frac{\partial \ell}{\partial W_{2,1}^{(i)}} - \frac{\partial \ell}{\partial W_{2,2}^{(i)}}} &= \norm{ \left( \prod_{j=i+1}^N (W_1^{(j)} + W_2^{(j)}) \right) \left( \frac{\partial \ell}{\partial y_1^{(i)}} - \frac{\partial \ell}{\partial y_2^{(i)}} \right) x \left( \prod_{j=1}^{i-1} (W_1^{(j)} + W_2^{(j)}) \right) }.
\end{align}

For MSE loss, the gradient at layer \( i \) is:
\begin{equation}
\frac{\partial \ell}{\partial y^{(i)}} = (y^{(N)} - y) \prod_{j=i+1}^N (W_1^{(j)} + W_2^{(j)}).
\end{equation}
The difference between the two network outputs is:
\begin{align}
& y_1^{(N)} - y_2^{(N)} = \left(\left( \prod_{j=1}^i (W_{1,1}^{(j)} + W_{1,2}^{(j)}) \right) \ - \left( \prod_{j=1}^i (W_{2,1}^{(j)} + W_{2,2}^{(j)}) \right) \right) x \\
& = \left( \prod_{\substack{j=1}}^{i- 1} (W_1^{(j)} + W_2^{(j)}) \right)(W_{1,1}^{(i)} + W_{2,1}^{(i)} - W_{1,2}^{(i)} - W_{2,2}^{(i)})x\\
& = \left( \prod_{\substack{j=1}}^{i- 1} (W_1^{(j)} + W_2^{(j)}) \right)(W_{2,1}^{(i)} - W_{2,2}^{(i)})x.
\end{align}
The difference between the differentials of the two network loss functions with respect to the output is:
\begin{align}
& \frac{\partial \ell}{\partial y_1^{(i)}} - \frac{\partial \ell}{\partial y_2^{(i)}} = \left( y_1^{(N)} - y_2^{(N)}\right)\prod_{j=i+1}^N (W_1^{(j)} + W_2^{(j)}) \\
& = \left( \prod_{\substack{j=1}}^{i- 1} (W_1^{(j)} + W_2^{(j)}) \right)(W_{2,1}^{(i)} - W_{2,2}^{(i)})x \prod_{j=i+1}^N (W_1^{(j)} + W_2^{(j)})\\
& = \left( \prod_{\substack{j=1 \\ j \neq i}}^N (W_1^{(j)} + W_2^{(j)}) \right)(W_{2,1}^{(i)} - W_{2,2}^{(i)})x.
\end{align}

Let \( \lambda^{(j)} = \norm{W_1^{(j)} + W_2^{(j)}} \) be the spectral norm.
\begin{align}
\left\|{\frac{\partial \ell}{\partial W_{2,1}^{(i)}} - \frac{\partial \ell}{\partial W_{2,2}^{(i)}}} \right\|&= \left\|{ \left( \prod_{j=i+1}^N (W_1^{(j)} + W_2^{(j)}) \right) \left( \frac{\partial \ell}{\partial y_1^{(i)}} - \frac{\partial \ell}{\partial y_2^{(i)}} \right) x \left( \prod_{j=1}^{i-1} (W_1^{(j)} + W_2^{(j)}) \right) }\right\|\\
&=\left\|{ \left( \prod_{\substack{j=1 \\ j \neq i}}^N (W_1^{(j)} + W_2^{(j)}) \right)^2 (W_{2,1}^{(i)} - W_{2,2}^{(i)})x^2  }\right\|\\
& \leq  \left( \prod_{j=1, j\neq i}^{N} \lambda^{(j)} \right)^2  \left\|{\boldsymbol{x}}\right\|^2 \left\| (W_{2,1}^{(i)} - W_{2,2}^{(i)})  \right\| .
\end{align}

Therefore the block-wise smoothness $\beta_i^{(N)}$ on layer $i \in [N]$ of an \(N\)-th layer MLP can be bounded by:
$\beta_i^{(N)} \ \leq \left( \prod_{j=1, j\neq i}^{N} \lambda^{(j)} \right)^2\left\|\boldsymbol{x}\right\|^2 $.

Let the weights of the \( i \)-th layer be decomposed as \( W^{(i)} = W_1^{(i)} + W_2^{(i)} \), with the spectral norm given by:
\[
\lambda^{(i)} = \left\|{W_1^{(i)} + W_2^{(i)}}\right\|,
\]
where \( W_1^{(i)} \) is frozen, and \( W_2^{(i)} \) is trainable. We aim to decompose the following expression:
\[
\left( \prod_{j=1, j \neq i}^N \lambda^{(j)} \right)^2.
\]

Using the triangle inequality:
\[
\lambda^{(j)} = \left\|{W_1^{(j)} + W_2^{(j)}}\right\| \leq \left\|{W_1^{(j)}}\right\| + \left\|{W_2^{(j)}} \right\|\triangleq \lambda_1^{(j)} + \lambda_2^{(j)},
\]
where \(\lambda_1^{(j)} = \left\|{W_1^{(j)}} \right\|\) and \(\lambda_2^{(j)} = \left\|{W_2^{(j)}} \right\|\). Extract $\lambda_1^{(j)}$ to facilitate product operations:
\[
\lambda^{(j)} \leq \lambda_1^{(j)} \left( 1 + \frac{\lambda_2^{(j)}}{\lambda_1^{(j)}} \right).
\]

Substituting the spectral norms of each layer into the product and expanding:
\[
\prod_{j=1, j \neq i}^N \lambda^{(j)} \leq \prod_{j=1, j \neq i}^N \lambda_1^{(j)} \cdot \prod_{j=1, j \neq i}^N \left( 1 + \frac{\lambda_2^{(j)}}{\lambda_1^{(j)}} \right).
\]
Squaring the product and expanding:
\[
\left( \prod_{j=1, j \neq i}^N \lambda^{(j)} \right)^2 \leq \left( \prod_{j=1, j \neq i}^N \lambda_1^{(j)} \cdot \prod_{j=1, j \neq i}^N  1 + \frac{\lambda_2^{(j)}}{\lambda_1^{(j)}} \right)^2.
\]
The final decomposition result is:
\[
\left( \prod_{j=1, j \neq i}^N \lambda^{(j)} \right)^2 \leq \left( \prod_{j=1, j \neq i}^N \lambda_1^{(j)} \right)^2 \cdot \prod_{j=1, j \neq i}^N \left( 1 + \frac{\lambda_2^{(j)}}{\lambda_1^{(j)}} \right)^2
\]

Finally, we can get the upper bound of the block-wise smoothness $\beta_i^{(N)}$ as:
\begin{equation}\label{equ: lora_smooth}
\beta_i^{(N)}  \leq \left( \prod_{j=1, j\neq i}^{N} \lambda^{(j)} \right)^2\left\|\boldsymbol{x}\right\|^2 \leq \left( \prod_{j=1, j \neq i}^N \lambda_1^{(j)} \right)^2 \cdot \prod_{j=1, j \neq i}^N \left( 1 + \frac{\lambda_2^{(j)}}{\lambda_1^{(j)}} \right)^2 \left\|\boldsymbol{x}\right\|^2
\end{equation}
Therefore, for the same LLM backbone, $N$ and $\lambda_1$ are the same. If we do not add a LoRA adapter to a certain layer, then $\lambda_2$ of that layer is 0. The fewer LoRA adapters we add to the LLM backbone, the smaller the second term, and the lower the upper bound of the block-wise smoothness $\beta_i^{(N)}$.

\subsection{numerical experiments}
We introduced $\beta$-smoothness in definition~\ref{definition1}, which refers to the Lipschitz continuity of the gradient of the loss function. As shown in Appendix~\ref{lora_therom}, Proposition~\ref{proposition2} is a general proposition that can be well extended to the LoRA method. In this section, we use the results of numerical experiments to prove that our theory is reasonable in the LoRA method when discussing the relationship between the number of network layers and $\beta$-smoothness.

Assuming that the function we are discussing is continuous and differentiable, we introduce a very small perturbation $\epsilon = 1e-5$. Then, Definition~\ref{definition1} can be simplified to:
\begin{equation}\label{equ: smooth-episilon}
\|\nabla f(w; z) - \nabla f(w + \epsilon; z )\| \leq \beta \| \epsilon \|.
\end{equation}
Therefore, the estimation formula for $\beta$-smoothness can be obtained:
\begin{equation}\label{equ: smooth-estimation}
\frac{\|\nabla f(w; z) - \nabla f(w + \epsilon; z )\| }{\|\epsilon\|}\leq \beta.
\end{equation}

\begin{figure}[H]
\centering
\includegraphics[width=0.9\textwidth]{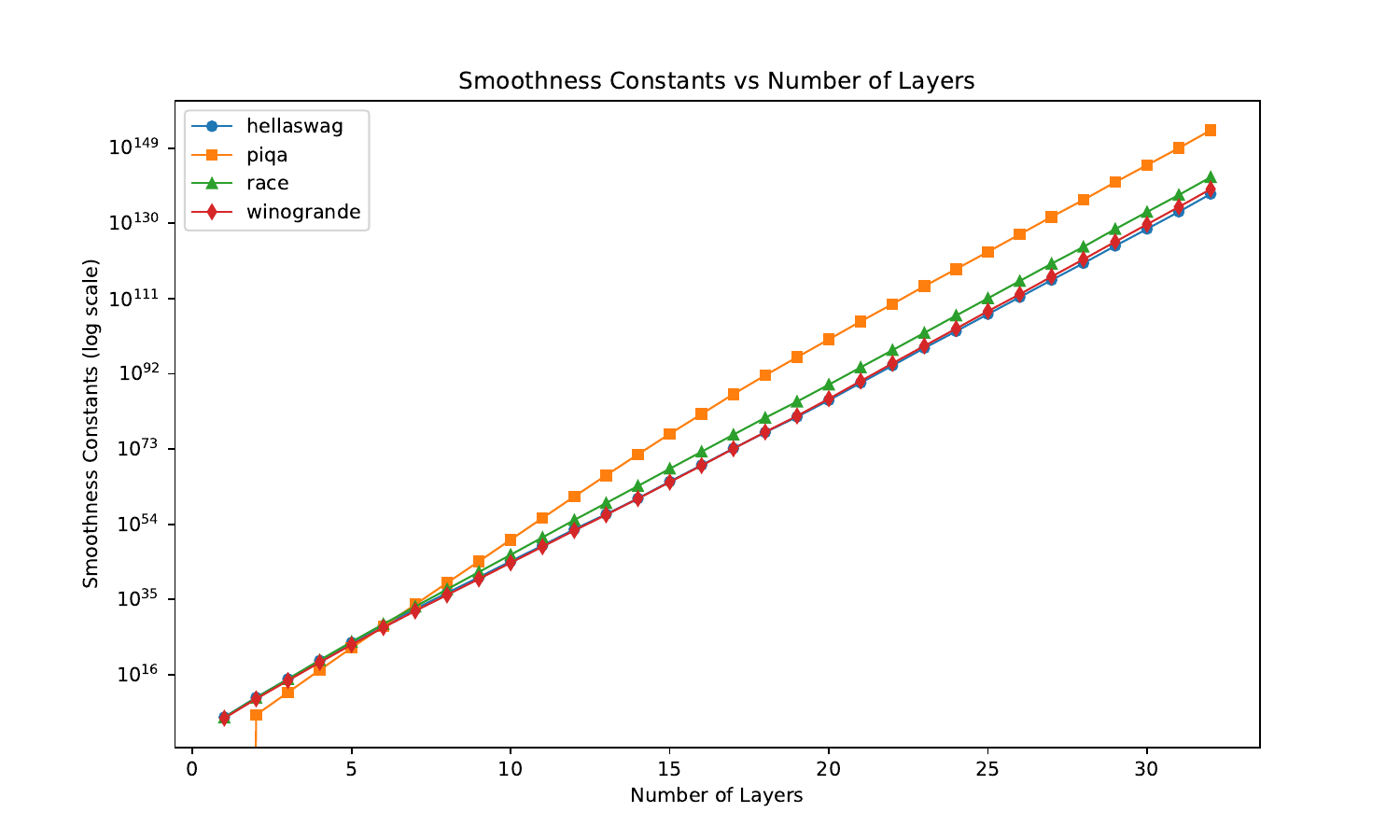} 
\caption{The $\beta$-smoothness constants of the Llama3-8B model across different datasets (Hellaswag, PIQA, RACE, Winogrande) as a function of the number of layers. The vertical axis is on a logarithmic scale, demonstrating that the $\beta$-smoothness increases exponentially with the number of layers for all datasets.}
\label{lips}
\end{figure}

According to Equation \ref{equ: smooth-estimation}, we use Llama3-8B as an example to calculate the $\beta$-smoothness of different layers of the model across various datasets (Hellaswag, PIQA, RACE, Winogrande). This is done to verify the relationship between the $\beta$-smoothness of model and the number of layers. The specific experimental steps are as follows: we selected a model fine-tuned with LoRA for each dataset, perturbed its trainable LoRA parameters, randomly sampled 10 data points from the corresponding dataset as input, and calculated the average $\beta$-smoothness of these ten data points. We use this average $\beta$-smoothness to represent the $\beta$-smoothness of network. The experimental results are shown in Figure~\ref{lips}. The vertical axis uses a logarithmic scale, and it can be seen that across different datasets, the $\beta$-smoothness of the Llama3-8B network (i.e., the Smoothness constant in Figure~\ref{lips}) increases exponentially with the number of network layers. In summary, large language models represented by Llama3-8B exhibit properties similar to those of MLP networks as described in Proposition~\ref{proposition2}, where $\beta$-smoothness increases exponentially with the number of network layers.

\section{\ours{} can identify the best single solution} \label{sec:local}
The literature extensively documents the prevalence of multiple local optima in hyperparameter optimization (HPO) problems~\cite{c:69,c:116,wei2025tmpo}. This phenomenon is particularly relevant in layer selection for LoRA fine-tuning, where varying layer combinations yield divergent performance outcomes due to intricate inter-layer interactions and task-specific characteristics. Here, we systematically analyze the emergence of multiple solutions and demonstrate how \ours{} effectively identifies the optimal configuration.
\subsection{Why Multiple Solutions Emerge?} \label{why}
The multiplicity of solutions arises from two primary factors.
\paragraph{Local vs. Global Optima:} In layer selection problems, multiple local optima typically exist due to varying performance of different layer combinations across tasks. Certain combinations may excel on training data but underperform on validation sets (indicating overfitting), while others demonstrate superior validation performance despite marginally weaker training results.

\paragraph{Optimization Trajectory:} As shown in Table~\ref{tab:different sample}, the evolutionary path of model during optimization leads to distinct local optima at different stages. Initial phases often favor simpler layer configurations, while subsequent optimization may uncover more sophisticated combinations that better capture task-specific characteristics.

\subsection{How \ours{} Finds the Optimal Single Solution} \label{how}
\begin{figure*}[t]
\centering
\includegraphics[width=0.8\textwidth]{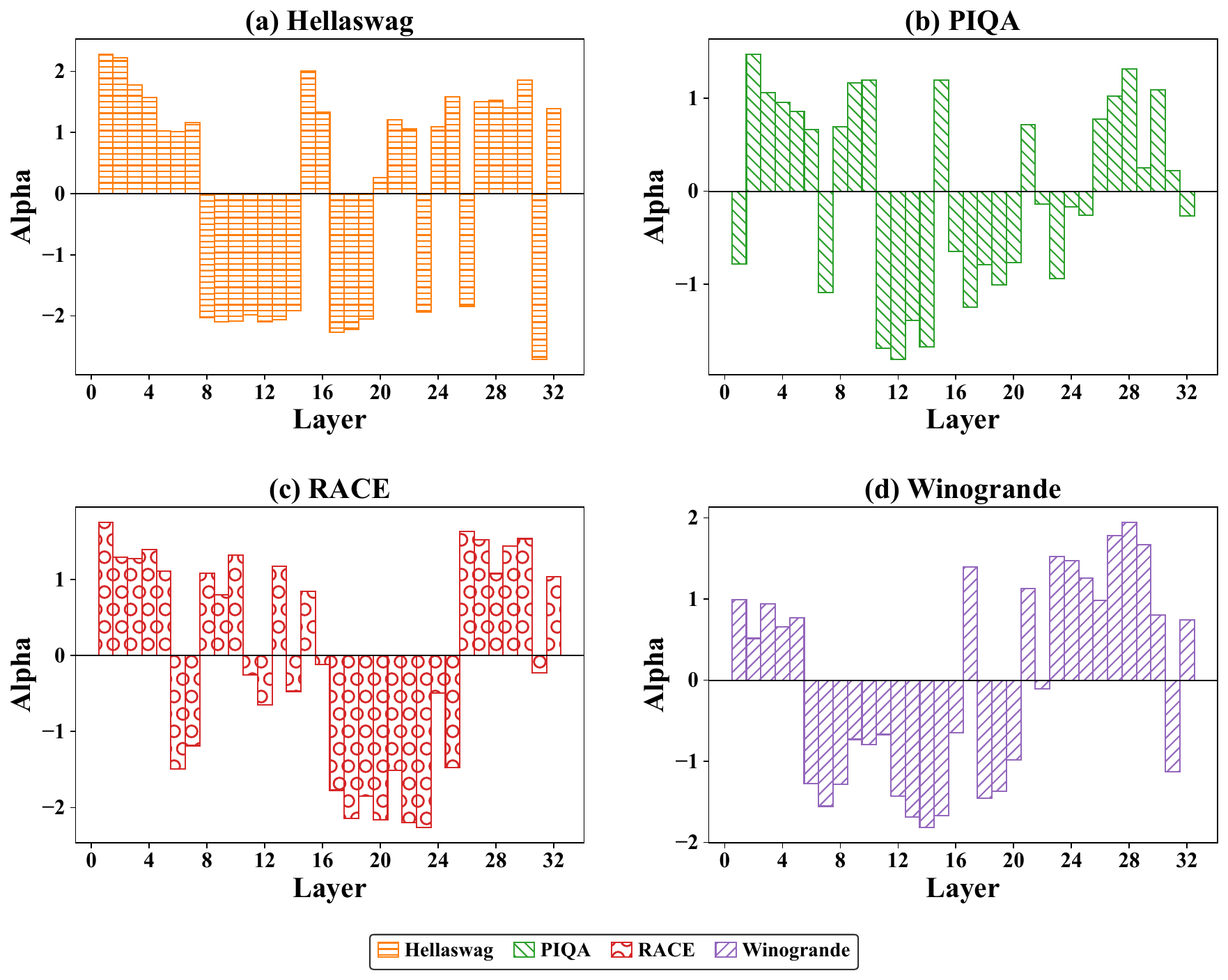} 
\caption{Distribution of $\alpha$ values across different layers for the Llama3-8B model using LoRA with a rank of 8. The panels (a) through (d) correspond to the Hellaswag, PIQA, RACE, and Winogrande datasets, respectively. Each bar represents the $\alpha$ value for a specific layer, with the x-axis indicating the layer number and the y-axis showing the $\alpha$ value.}
\label{fig:14}
\vspace{-3mm}
\end{figure*}

To improve the performance of \ours{} and prevent convergence to suboptimal local solutions, we implemented a dual-strategy approach. On the training strategy, we established validation set performance as our primary optimization criterion, systematically evaluating various layer combinations to determine the most general configuration. In addition, we adopted an early stopping mechanism that monitors validation performance and terminates optimization when a steady state is reached, thereby preventing overfitting and selecting the best layer combination. On the optimization target preconditioning, we apply continuous relaxation to the target hyperparameters using formula~\ref{eq:cshf} in Section~\ref{sec:initial}, which significantly improves optimization stability and efficiency. As shown in Figure ~\ref{fig:14}, our strategy produces well-converged hyperparameters with significant absolute value differences ($\alpha > 0$ vs. $\alpha \leq 0$), reflecting significant changes in layer importance. This differential importance ultimately enables \ours{} to consistently identify a single optimal solution.

\newpage
\section{Special cases} \label{sec:Special cases}
This section details the performance of \ours{} and LoRA across four distinct datasets. The results indicate that \ours{} demonstrates superior comprehension and judgment on more challenging questions within the test dataset, compared to LoRA. In certain instances, \ours{} successfully explains problems not previously encountered during training, showcasing its robust learning and generalization capabilities.
\begin{tcolorbox}[title={\textbf{\small Special cases of Hellaswag}},
colback=whitesmoke, colframe=darksalmon,  boxrule=2pt, arc=0mm]
{\scriptsize
\begin{lstlisting}[style=mystyle]
dataset: Hellaswag
    "1": {
        "origin_prompt": "A lady walks to a barbell. She bends down and grabs 
        the pole. The lady\n
        Question: Which ending makes the most sense?\n
        A. swings and lands in her arms.\n
        B. pulls the barbell forward.\n
        C. pulls a rope attached to the barbell.\n
        D. stands and lifts the weight over her head.\n
        You may choose from 'A', 'B', 'C', 'D'.\n
        Answer:",
        "Flexora prediciton": " D",
        "LoRA prediciton" : "B",
        "gold": "D"
    },
    "2": {
        "origin_prompt": "Two women in a child are shown in a canoe while a man 
        pulls the canoe while standing in the water, with other individuals 
        visible in the background. The child and a different man\n
        Question: Which ending makes the most sense?\n
        A. are then shown paddling down a river in a boat while a woman talks.\n
        B. are driving the canoe, they go down the river flowing side to side.\n
        C. sit in a canoe while the man paddles.\n
        D. walking go down the rapids, while the man in his helicopter almost 
        falls and goes out of canoehood.\n
        You may choose from 'A', 'B', 'C', 'D'.\n
        Answer:",
        "Flexora prediciton": " C",
        "LoRA prediciton" : "B",
        "gold": "C"
    },
    "3": {
        "origin_prompt": "The boy lifts his body above the height of a pole. 
        The boy lands on his back on to a red mat. The boy\n
        Question: Which ending makes the most sense?\n
        A. turns his body around on the mat.\n
        B. gets up from the mat.\n
        C. continues to lift his body over the pole.\n
        D. wiggles out of the mat.\n
        You may choose from 'A', 'B', 'C', 'D'.\n
        Answer:",
        "Flexora prediciton": " B",
        "LoRA prediciton" : "B",
        "gold": "B"
    }
    "4": {
        "origin_prompt": "We see a person holding face wash then putting it on
        their face. They rinse the face and add the face wash with a brush. We\n
        Question: Which ending makes the most sense?\n
        A. see a closing title screen.\n
        B. see a black screen with the credits.\n
        C. see an illustration on how to add the wash using a brush.\n
        D. then see a replay then the person putting the face wash on.\n
        You may choose from 'A', 'B', 'C', 'D'.\n
        Answer:",
        "Flexora prediction": " C",
        "LoRA prediciton" : "A",
        "gold": "C"
    },
    
    
\end{lstlisting}
}
\end{tcolorbox}

\begin{tcolorbox}[title={\textbf{\small Special cases of PIQA}},
colback=whitesmoke, colframe=darksalmon,  boxrule=2pt, arc=0mm]
{\scriptsize
\begin{lstlisting}[style=mystyle]
dataset: PIQA
    "1": {
        "origin_prompt": "ice box\n
        A. will turn into a cooler if you add water to it\n
        B. will turn into a cooler if you add soda to it\n
        Answer:",
        "Flexora prediciton": "A",
        "LoRA prediciton" : "A",
        "gold": "A"
    },
    "2": {
        "origin_prompt": "how do you put eyelashes on?\n
        A. glue them on with mascara.\n
        B. put eyelash glue on the fake eyelashes and then let it get tacky. 
        then, place it on top of your actual eyelashes and let it dry on.\n
        Answer:",
        "Flexora prediciton": "A",
        "LoRA prediciton" : "B",,
        "gold": "B"
    },
    "3": {
        "origin_prompt": "How do I fill holes and tiny gaps in the concrete when 
        making a concrete countertop?\n
        A. Use  a concrete slurry\n
        B. Use  a concrete brush\n
        Answer:",
        "Flexora prediciton": "A",
        "LoRA prediciton" : "B",
        "gold": "A"
    }
    "4": {
        "origin_prompt": "When I'm deep frying a turkey, how defrosted 
        should it be?\n
        A. It should be completely defrosted, otherwise the oil may 
        boil over and start a fire.\n
        B. It should be completely frozen, otherwise the oil may boil 
        over and start a fire.\n
        Answer:",
        "\ours{} prediction": " A\nExplanation: A turkey should be completely
        defrosted before deep frying. If it is not, the oil may boil over
        and start a fire.",
        "Flexora prediciton" : "A",
        "LoRA prediciton" : "B",
        "gold": "A"
    },
    "5": {
        "origin_prompt": "How do you properly prepare a steak.\n
        A. Take the steak out of warm storage and let come to room temperature,
        generously add salt and pepper to both sides and let sit 
        for 10 minutes.\n
        B. Take the steak out of cold storage and let come to room temperature,
        generously add salt and pepper to both sides and let sit 
        for 10 minutes.\n
        Answer:",
        "Flexora prediction": " B\nExplanation: B. Taking the steak 
        out of cold storage and letting it come to room temperature is
        the correct answer. The steak should be at room temperature before
        cooking. The steak should be generously salted and peppered on both
        sides and let sit for 10 minutes.",
        "LoRA prediciton" : "B",
        "gold": "B"
    },
    "6": {
        "origin_prompt": "To cream butter and sugar together, you can\n
        A. Place it in a bowl and use a hand warmer\n
        B. Place in a bowl and use a hand mixer\n
        Answer:",
        "Flexora prediction": " B\nExplanation: B. Place in a bowl and 
        use a hand mixer\nExplanation: To cream butter and sugar together,
        you can place it in a bowl and use a hand mixer.",
        "LoRA prediciton" : "B",
        "gold": "B"
    },
\end{lstlisting}
}
\end{tcolorbox}
\begin{tcolorbox}[title={\textbf{\small Special cases of RACE}},
colback=whitesmoke, colframe=darksalmon,  boxrule=2pt, arc=0mm]
{\scriptsize
\begin{lstlisting}[style=mystyle]
dataset: RACE
    "1": {
        "origin_prompt": "Read the article, and answer the question by replying A,
        B, C or D.\n\n
        Article:\nThe rain had continued for a week and the flood
        had created a big river which were running by Nancy Brown's
        farm. As she tried to gather her cows to a higher ground,
        she slipped and hit her head on a fallen tree trunk.
        The fall made her unconscious for a moment or two. When she came to,
        Lizzie, one of her oldest and favorite cows, was licking her face. \n
        At that time, the water level on the farm was still rising.
        Nancy gathered all her strength to get up and began walking
        slowly with Lizzie. The rain had become much heavier,
        and the water in the field was now waist high. Nancy's pace
        got slower and slower because she felt a great pain in her head.
        Finally, all she could do was to throw her arm around Lizzie's
        neck and try to hang on. About 20 minutes later, Lizzie managed
        to pull herself and Nancy out of the rising water and onto
        a bit of high land, which seemed like a small island in
        the middle of a lake of white water. \n
        Even though it was about noon, the sky was so dark and the rain
        and lightning was so bad that it took rescuers more than
        two hours to discover Nancy. A man from a helicopter 
        lowered a rope, but Nancy couldn't catch it. A moment later,
        two men landed on the small island from a ladder in the helicopter.
        They raised her into the helicopter and took her to the school gym,
        where the Red Cross had set up an emergency shelter.
        \nWhen the flood disappeared two days later, Nancy immediately
        went back to the \"island.\" Lizzie was gone. She was one of
        19 cows that Nancy had lost in the flood. \"I owe my life to
        her,\" said Nancy with tears.\n\n
        Q: What did Nancy try to do before she fell over?\n\n
        A. Measure the depth of the river\n
        B. Look for a fallen tree trunk\n
        C. Protect her cows from being drowned\n
        D. Run away from the flooded farm\n",
",
        "Flexora prediciton": "D",
        "LoRA prediciton" : "B",
        "gold": "D"
    }
\end{lstlisting}
}
\end{tcolorbox}
\begin{tcolorbox}[title={\textbf{\small Special cases of Winogrande}},
colback=whitesmoke, colframe=darksalmon,  boxrule=2pt, arc=0mm]
{\scriptsize
\begin{lstlisting}[style=mystyle]
dataset: Winogrande
     "1": {
        "origin_prompt": "Question: Sarah was a much better surgeon 
        than Maria so _ always got the easier cases.\n
        A. Sarah\n
        B. Maria\n
        Answer:",
        "Flexora prediciton": "B",
        "LoRA prediciton" : "B",
        "gold": "B"
    },
    "2": {
        "origin_prompt": "Question: Sarah was a much better surgeon 
        than Maria so _ always got the harder cases.\n
        A. Sarah\n
        B. Maria\n
        Answer:",
        "Flexora prediciton": "B",
        "LoRA prediciton" : "B",
        "gold": "A"
    },
    "3": {
        "origin_prompt": "Question: They were worried the wine would ruin 
        the bed and the blanket, but the _ was't ruined.\n
        A. blanket\n
        B. bed\n
        Answer:",
        "Flexora prediciton": "B",
        "LoRA prediciton" : "A",
        "gold": "B"
    },
    "4": {
        "origin_prompt": "Question: Terry tried to bake the eggplant 
        in the toaster oven but the _ was too big.\n
        A. eggplant\n
        B. toaster\nAnswer:",
        "Flexora prediction": " A\nExplanation: The eggplant was 
        too big to fit in the toaster oven.",
        "LoRA prediciton" : "B",
        "gold": "A"
    },
    "5": {
        "origin_prompt": "Question: At night, Jeffrey always stays up 
        later than Hunter to watch TV because _ wakes up late.\n
        A. Jeffrey\n
        B. Hunter\n
        Answer:",
        "Flexora prediciton": "A",
        "LoRA prediciton" : "B",
        "gold": "A"
    },
    "6": {
        "origin_prompt": "Question: The cat of Sarah has some mouth problems,
        so she takes it to see Maria. _ is a responsible cat owner.\n
        A. Sarah\n
        B. Maria\n
        Answer:",
        "Flexora prediction": " A\nExplanation: A is a responsible cat 
        owner because she takes her cat to see a veterinarian.",
        "LoRA prediciton" : "B",
        "gold": "A"
    },
    "7": {
        "origin_prompt": "Question: Benjamin was chosen instead of Brett to
        be the makeup artist for the play because _ was less experienced.\n
        A. Benjamin\n
        B. Brett\n
        Answer:",
        "Flexora prediction": " B",
        "LoRA prediciton" : "A",
        "gold": "B"
    },
\end{lstlisting}
}
\end{tcolorbox}
\end{appendices}

\end{document}